\documentclass{article} 
\usepackage[final]{colm2026_conference}

\usepackage{amsmath,amsfonts,bm}

\def\eqref#1{equation~\ref{#1}}

\def\1{\bm{1}}

\DeclareMathAlphabet{\mathsfit}{\encodingdefault}{\sfdefault}{m}{sl}
\SetMathAlphabet{\mathsfit}{bold}{\encodingdefault}{\sfdefault}{bx}{n}

\newcommand{\softmax}{\mathrm{softmax}}

\usepackage{lineno}
\usepackage{enumitem}
\usepackage{hyperref}
\usepackage{url}
\usepackage{graphicx}
\usepackage{color}
\usepackage{booktabs} 
\usepackage{multirow}
\usepackage{amsmath,amssymb,amsthm}
\usepackage{algorithm}
\usepackage{algpseudocode}
\usepackage{subcaption}
\usepackage{makecell}

\definecolor{darkblue}{rgb}{0, 0, 0.5}
\hypersetup{colorlinks=true, citecolor=darkblue, linkcolor=darkblue, urlcolor=darkblue}

\title{Beyond Post-Hoc Temperature Scaling: Bilevel Optimization for LLM Calibration}

\author{Ruochen Jin$^{1,}$\thanks{Equal contribution.\quad $^{\dagger}$Corresponding author.}\ , Zhanliang Wang$^{2,*}$, Zongyu Dai$^2$, Jiancong Xiao$^{3,\dagger}$, and Bojian Hou$^{2,\dagger}$\\
$^1$Dartmouth College, Hanover, NH; $^2$University of Pennsylvania, Philadelphia, PA; \\
$^3$National University of Singapore, Singapore\\
\texttt{ruochen.jin.gr@dartmouth.edu, aaronwzl@upenn.edu, daizy@sas.upenn.edu}\\
\texttt{jiancongxiao@nus.edu.sg, bojianh@upenn.edu}
}

\begin{document}

\ifcolmsubmission
\linenumbers
\fi

\maketitle

\begin{abstract}
Preference alignment often makes large language models (LLMs) overconfident and poorly calibrated. Traditional post-hoc temperature scaling is inherently domain-dependent: a temperature fitted on one domain does not generalize across domains. This motivates us to modify model parameters during training to improve calibration. We propose maximizing the entropy of predictive distributions as the calibration objective, which directly targets overconfidence by discouraging overly concentrated predictions. Inspired by temperature scaling, we realize this through a bilevel optimization formulation, where the lower level trains the model under a parametric loss and the upper level selects loss hyperparameters to maximize entropy. To make the framework practical at LLM scale, we adopt an efficient first-order approximation that avoids explicit second-order computation. Across both multiple-choice and open-ended generative question answering, experiments demonstrate that our method yields well-calibrated LLMs with particular advantages in out-of-domain generalization\footnote{Code is available at \url{https://github.com/BojianHou/calm}.}.
\end{abstract}

\section{Introduction}
Large Language Models (LLMs) \citep{openai2023gpt4,anthropic2024claude} have become a foundation of modern Natural Language Processing (NLP), achieving strong performance across a wide range of tasks~\citep{bubeck2023sparks,chowdhery2023palm,touvron2023llama,team2023gemini}. Beyond task performance, practical success also requires reliability. \emph{Preference alignment}, notably through Reinforcement Learning from Human Feedback (RLHF) \citep{ouyang2022training} and Direct Preference Optimization (DPO) \citep{rafailov2024direct}, steers models toward human-preferred behaviors but degrades calibration as a side effect. While pre-trained LLMs are often reasonably calibrated, aligned models frequently become overconfident \citep{openai2023gpt4,he2023investigating}. This matters because confidence is widely used as a proxy for reliability; in high-stakes settings such as healthcare or legal analysis, miscalibrated confidence can lead to poor decisions and misplaced trust \citep{savage2025large}.

Most existing calibration methods are post-hoc, with Temperature Scaling (TS) as the canonical example. TS is two-stage and \emph{test-time}: the model is first trained, then a temperature is selected on a validation set to improve the calibration of predicted probabilities. Although simple and effective, TS is inherently dataset-dependent---a temperature fitted on one dataset often does not transfer to another domain---so it is difficult to find a single ``general temperature'' that calibrates well across diverse datasets.

This motivates a training-time view of calibration. Instead of searching for a dataset-specific temperature only after training, we fine-tune the model itself so that a temperature-like adjustment generalizes across datasets, treating the model parameters, rather than a purely post-hoc correction, as the target. This requires a training objective that explicitly encourages calibrated predictions during optimization.

A second challenge lies in selecting an appropriate calibration-oriented objective. Our empirical observations in Figure~\ref{fig:cwece}, together with prior findings \citep{xiao2025restoring, leng2025taming}, suggest that overconfidence is the predominant calibration failure mode in aligned LLMs. This observation motivates the use of entropy maximization, which provides a controllable direction for improving calibration without prescribing a specific target solution. The maximum-entropy solution represents the opposite extreme of underconfidence. Increasing the predictive entropy therefore moves the model continuously between these two regimes, potentially allowing it to reach a point at which its confidence better matches its predictive accuracy.

Inspired by the two-stage nature of temperature scaling, we formulate calibration-oriented training as a bilevel optimization problem. Temperature scaling can be viewed as a special two-stage procedure: the lower level learns the model parameters, and the upper level selects a temperature to improve calibration. We generalize this idea from post-hoc test-time adjustment to training-time loss design. In our formulation, the lower level fine-tunes the LLM under a parametric training loss, while the upper level optimizes temperature-like loss hyperparameters using the entropy-based calibration objective on held-out inputs. From this perspective, post-hoc TS becomes a special case of a broader training-time framework for calibration-oriented optimization.

A key practical challenge, however, is that bilevel optimization is computationally expensive for LLMs. Exact hypergradient computation typically involves differentiating through the lower-level optimization, which requires second-order derivatives and, in implicit formulations, may further involve Hessian-inverse terms~\citep{tarzanagh2024online,nazari2025stochastic}. These operations are prohibitively expensive for modern LLM fine-tuning. To address this issue, we develop an efficient first-order bilevel optimization framework inspired by the algorithm introduced in Bilevel Optimization Made Easy (BOME; \citealt{liu2022bome}) and propose Calibration for Large Models via Bilevel Optimization (CALM). Our method avoids explicit hypergradient computation and instead uses first-order updates to approximate the upper-level optimization, making calibration-oriented bilevel training practical at scale.

Our contributions are as follows:

\begin{itemize}
    \item We identify the dataset dependence of post-hoc temperature scaling and propose a training-time perspective that aims to learn models admitting a more general calibration adjustment across datasets.
    \item We formulate calibration-oriented loss design as a bilevel optimization problem, where the lower level trains the model and the upper level optimizes temperature-like loss hyperparameters using an entropy-based objective.
    \item We propose an entropy-maximization upper-level objective to mitigate overconfidence in aligned LLMs without requiring additional hard labels.
    \item We empirically demonstrate that our method provides strong calibration performance across both multiple-choice and open-ended generative question answering, with especially clear advantages in out-of-domain settings, against strong training-time baselines including calibration-aware fine-tuning.
\end{itemize}

\section{Related Work}

\textbf{Calibration for LLMs.}
Recent work shows that calibration remains a central reliability issue for large language models, especially after instruction tuning \citep{jiang2021can,xiao2022uncertainty,chen2022close} or preference alignment \citep{ouyang2022training,xiao2024algorithmic,liu2025statistical}. Prior studies on LLM calibration have mainly followed three directions. First, several works adapt \emph{post-hoc} calibration ideas to LLMs, with temperature scaling often emerging as a strong baseline \citep{jiang2021can,chen2022close,shen2024thermometer}. Second, some methods introduce auxiliary models, additional calibration heads, or external confidence predictors to recalibrate model outputs \citep{zhang2021knowing,kadavath2022language,shen2024thermometer}, including lightweight calibration layers trained on top of a frozen LM \citep{liu2023litcab} and probes that read a model's hidden states to recover latent correctness signals \citep{burns2023discovering}. Third, another line of work teaches LLMs to explicitly express their uncertainty, either through prompting or through dedicated supervision on confidence estimation \citep{lin2022teaching,tian2023just,xu2024sayself}. Although related, work on debiasing in-context learning \citep{abbas24a,han2023prototypical,jiang2023generative} addresses prompt-induced prediction bias rather than statistical calibration in the strict sense. In addition, \citet{wang2026semantic} proposed a semantic-sampling framework for evaluating calibration in open-ended question answering.

\textbf{Training-time optimization for LLM calibration.}
More recent studies move beyond post-hoc correction and directly optimize calibration during fine-tuning or alignment. Prior work has proposed uncertainty-aware or calibration-aware objectives that modify the training loss so as to improve calibration while retaining task performance \citep{han2024enhancing,xiao2025restoring}. Related work also studies how the interaction between fine-tuning data and a model's prior knowledge can systematically induce overconfidence, and proposes cognition-aware training strategies to mitigate this effect \citep{wang2025objective}. A complementary line intervenes during alignment itself, calibrating the RLHF reward model to curb the overconfidence introduced by preference optimization \citep{leng2025taming}. In addition, recent evidence suggests that label smoothing can be an effective training-time regularizer for preserving calibration during LLM fine-tuning, although its effectiveness may depend on model scale and vocabulary size \citep{huang2025calibrated}.

\textbf{Positioning of our work.} Our work belongs to this emerging line of \emph{training-time optimization for calibration}: rather than post-hoc scaling or auxiliary confidence estimators, we directly optimize calibration during supervised fine-tuning, motivated by evidence that miscalibration in aligned LLMs is induced by training itself. Unlike auxiliary-head calibrators \citep{liu2023litcab}, hidden-state probes \citep{burns2023discovering}, or RLHF-time reward calibration \citep{leng2025taming} which add a module, read activations, or reshape rewards, CALM optimizes the base model's own predictive distribution during fine-tuning and is thus complementary to them.

\section{Problem Setup}

Let $\mathcal{S}_{\mathcal{T}}=\{(\mathbf{x}_i,y_i)\}_{i=1}^n$ be a training dataset of size $n$, and let $\mathcal{S}_{\mathcal{V}}=\{(\mathbf{x}_i^{v},y_i^{v})\}_{i=1}^m$ be a validation dataset of size $m$, where each sample $(\mathbf{x},y)$ is drawn i.i.d.\ from a distribution $\mathcal{D}$ over $\mathcal{X}\times[K]$, with input space $\mathcal{X}$ and label space $[K]:=\{1,\dots,K\}$. We use $(X,Y)\sim\mathcal{D}$ to denote a generic test sample drawn from the same distribution.

Let $z_\theta:\mathcal{X}\to\mathbb{R}^K$ denote a model parameterized by $\theta$, where the corresponding logit vector is given by $z_\theta(\mathbf{x}) = (z_{\theta,1}(\mathbf{x}), \dots, z_{\theta,K}(\mathbf{x}))$. Its predictive distribution is $f_\theta(\cdot\mid\mathbf{x})=\softmax(z_\theta(\mathbf{x}))\in\Delta_K$, where $\Delta_K:=\{\mathbf{p}\in[0,1]^K:\sum_{i=1}^K p_i=1\}$ is the $K$-dimensional probability simplex. For each class $i\in[K]$, the predicted probability is
\[
f_\theta(i\mid\mathbf{x})
=
\frac{\exp(z_{\theta,i}(\mathbf{x}))}{\sum_{j=1}^K \exp(z_{\theta,j}(\mathbf{x}))},
\]
and the predicted label is $\hat{y}_\theta(\mathbf{x})=\arg\max_{i\in[K]} z_{\theta,i}(\mathbf{x})$.

\paragraph{Multiple-choice LLM setting.}
To extend the classification formulation to a multiple-choice setting with four answer candidates $\{A,B,C,D\}$\footnote{We adopt the four-option format $\{A,B,C,D\}$ for readability; the formulation naturally generalizes to a $K$-choice setting.} in LLMs, we extract the corresponding logits $z_{\theta,A}(\mathbf{x}), z_{\theta,B}(\mathbf{x}), z_{\theta,C}(\mathbf{x})$, and $z_{\theta,D}(\mathbf{x})$. The definitions of predicted probabilities and labels then follow analogously from the classification setting.

\paragraph{Open-ended generation setting.}
The multiple-choice view above makes calibration easy to define over a fixed option set, but CALM itself is not tied to it: the entropy-based upper-level objective operates on the model's next-token predictive distribution and requires no fixed label set (Section~\ref{sec:method}). We therefore also evaluate CALM on free-form generation, where the answer is an arbitrary token sequence rather than one of $K$ options. In this setting a token-level confidence no longer corresponds to a single categorical prediction, so at evaluation time we follow the semantic-sampling protocol of \citet{wang2026semantic}: for a query we draw multiple samples, cluster semantically equivalent generations, and treat each cluster's empirical frequency as the model's confidence in that answer, which yields a well-defined calibration target (Sem-ECE) for open-ended outputs.

\section{Method}
\label{sec:method}
Let us start from the temperature scaling approach. Given a fixed model, temperature scaling is a post-processing (or test-time) calibration technique that rescales logits by a positive scalar temperature parameter $\tau>0$. Specifically, given logits $z_\theta(\mathbf{x})\in\mathbb{R}^K$, the temperature-scaled predictive distribution is defined as
\[
p_i^{(\tau)}(\mathbf{x})
=
\frac{\exp\big(z_{\theta,i}(\mathbf{x}) / \tau\big)}
{\sum_{j=1}^K \exp\big(z_{\theta,j}(\mathbf{x}) / \tau\big)},
\qquad i\in[K].
\]

The temperature $\tau$ controls the confidence of the prediction: $\tau<1$ sharpens the distribution, while $\tau>1$ produces a more uniform (less confident) distribution. The temperature $\tau$ is estimated by minimizing the empirical cross-entropy on a dataset:
\begin{equation}
\label{eq:ts_loss}
\min_{\tau>0}
\;
-\frac{1}{n}
\sum_{i=1}^n
\log p^{(\tau)}_{y_i}(\mathbf{x}_i).
\end{equation}
The optimized temperature often improves the agreement between predictive confidence and empirical accuracy on the calibration dataset. A significant drawback of temperature scaling is that the learned temperature $\tau$ is highly specific to the current dataset and generalizes poorly to others. Consequently, achieving proper model calibration requires identifying a distinct temperature for each unique dataset \citep{yu2022robust}. 

In the era of LLMs, this has become a major drawback compared to the previous era of specialized classification models. A truly robust LLM should be well-calibrated across a diverse range of tasks without requiring task-specific tuning.

To address this, we identify three potential causes for poor generalization and tackle them simultaneously. First, one possible source of the lack of generalizability stems from the scalar nature of $\tau$. A single-dimension value may be too restrictive to capture complex data distributions. Consequently, we consider a higher-dimensional temperature parameter:
\[
\boldsymbol{\alpha} = (\mathbf{l}, \boldsymbol{\tau}), \quad \mathbf{l} \in \mathbb{R}^K, \; \boldsymbol{\tau} \in \mathbb{R}_{>0}^K.
\]
Here, $\mathbf{l}$ introduces additive logit shifts, while $\boldsymbol{\tau}$ serves as a vector of class-dependent temperatures. In the experiments and appendices we equivalently write the multiplicative factor as $\mathbf{d}_y := \tau_y^{-1}$, so $(\mathbf{d}_y, \mathbf{l}_y)$ are the per-vocabulary adjustments; a single global temperature $T$ is the tied special case ($\mathbf{d}_y$ all equal, $\mathbf{l}=\mathbf{0}$). Given an input-label pair $(\mathbf{x}, y)$ and the model's logits $z_\theta(\mathbf{x}) \in \mathbb{R}^K$, the predicted probabilities are defined as:
\begin{equation}
f_{\boldsymbol{\alpha}, \theta}(y\mid \mathbf{x})=\frac{\exp(\tau_y^{-1} z_{\theta,y}(\mathbf{x}) + l_y)}
{\sum_{i=1}^K \exp(\tau_i^{-1} z_{\theta,i}(\mathbf{x}) + l_i)}.
\label{eq:train_loss}
\end{equation}
Notably, when $l_i = 0$ and $\tau_i = \tau$ for all $i \in [K]$, Equation (\ref{eq:train_loss}) reduces to standard temperature scaling. By applying the higher-dimensional temperature to Equation~(\ref{eq:ts_loss}), the loss becomes:
\begin{equation}
\label{eq:ts_loss2}
\min_{\boldsymbol{\alpha}}
\;
-\frac{1}{n}
\sum_{i=1}^n
\log f_{\boldsymbol{\alpha}, \theta}(y_i\mid \mathbf{x}_i).
\end{equation}
Second, we posit that the lack of generalizability may be an inherent limitation of post-processing, where the model parameters remain fixed. While a universal temperature may not exist for a specific given model, there likely exists a model with comparable performance that is more amenable to a universal temperature. Consequently, we extend temperature scaling to training time, aiming to optimize both the temperature and the model parameters jointly. Under this framework, Equation~(\ref{eq:ts_loss2}) is reformulated as:
\begin{equation}
\label{eq:ts_loss3}
\min_{\boldsymbol{\alpha}, \theta}
\;
-\frac{1}{n}
\sum_{i=1}^n
\log f_{\boldsymbol{\alpha}, \theta}(y_i\mid \mathbf{x}_i).
\end{equation}
This step moves beyond post-hoc calibration and should be viewed as a training-time generalization of temperature scaling rather than standard temperature scaling itself.

Third, the poor generalization may stem from overfitting to the specific characteristics of the validation dataset used for calibration. To address this, we reformulate Equation~(\ref{eq:ts_loss3}) into a more generalized loss function that reduces dependence on ground-truth labels. A common failure mode of aligned LLMs is \emph{overconfidence}: the predictive distribution is often excessively concentrated even when the model is uncertain. We therefore employ entropy maximization as a direct mechanism to discourage overly sharp predictions.

Given the predictive distribution $f_\theta(\cdot\mid\mathbf{x})\in\Delta_K$, its Shannon entropy is defined as:
\begin{equation}
H\!\left(f_{\boldsymbol{\alpha}, \theta}(\cdot\mid\mathbf{x})\right)
=
-\sum_{c=1}^K
f_{\boldsymbol{\alpha}, \theta}(c\mid\mathbf{x})\log f_{\boldsymbol{\alpha}, \theta}(c\mid\mathbf{x}).
\label{eq:entropy}
\end{equation}
By replacing the cross-entropy in Equation~(\ref{eq:ts_loss3}) with the entropy term, we obtain the following objective:
\begin{equation}
\label{eq:ts_loss4}
\min_{\boldsymbol{\alpha}, \theta}
\;
-\frac{1}{n}
\sum_{i=1}^n
H\!\left(f_{\boldsymbol{\alpha}, \theta}(\cdot\mid\mathbf{x})\right).
\end{equation}

This objective is appealing because entropy maximization provides a controllable direction for calibration rather than prescribing the final solution. Modern neural networks are typically overconfident, whereas the maximum-entropy solution represents extreme underconfidence. As entropy increases, the model moves continuously between these regimes and may pass through a point where its confidence better matches its predictive accuracy. 

Entropy maximization alone could overshoot this point and impair discrimination. Without further constraints, optimizing solely for entropy could lead to a degradation in task performance. To address this, we reintroduce the cross-entropy loss on the training set as a constraint, requiring the model parameters to be optimal with respect to $\boldsymbol{\alpha}$:
\begin{equation}
\label{eq:constraint}
\theta \in \arg\min_{\theta'}
\;
-\frac{1}{n}
\sum_{i=1}^n
\log f_{\boldsymbol{\alpha}, \theta'}(y_i\mid \mathbf{x}_i).
\end{equation}
Finally, by applying the training set $\mathcal{S}_{\mathcal{T}}$ and the calibration set $\mathcal{S}_{\mathcal{V}}$ to the objectives in Equation~(\ref{eq:ts_loss4}) and Equation~(\ref{eq:constraint}), we arrive at the bilevel optimization formulation. For notational convenience, we denote the upper-level entropy objective as $\mathcal{L}_{\mathrm{ent}}$ and the lower-level training loss as $\mathcal{L}_{\mathrm{tr}}$:
\begin{equation}
\label{eq:bilevel}
\begin{aligned}
\min_{\boldsymbol{\alpha}, \theta} & \quad
\underbrace{-\frac{1}{|\mathcal{S}_{\mathcal{V}}|}
\sum_{\mathbf{x}_i \in \mathcal{S}_{\mathcal{V}}}
H\!\left(f_{\boldsymbol{\alpha}, \theta}(\cdot\mid\mathbf{x}_i)\right)}_{\displaystyle \mathcal{L}_{\mathrm{ent}}^{\mathcal{S}_{\mathcal{V}}}(\boldsymbol{\alpha},\theta)} 
\text{s.t.} & \quad \theta \in \arg\min_{\theta'}
\;
\underbrace{-\frac{1}{|\mathcal{S}_{\mathcal{T}}|}
\sum_{(\mathbf{x}_i, y_i) \in \mathcal{S}_{\mathcal{T}}}
\log f_{\boldsymbol{\alpha}, \theta'}(y_i \mid \mathbf{x}_i)}_{\displaystyle \mathcal{L}_{\mathrm{tr}}^{\mathcal{S}_{\mathcal{T}}}(\theta';\boldsymbol{\alpha})}.
\end{aligned}
\end{equation}

The upper level minimizes $\mathcal{L}_{\mathrm{ent}}$ to discourage overconfident predictions, while the lower level ensures that $\theta$ remains optimal for $\mathcal{L}_{\mathrm{tr}}$ under the given $\boldsymbol{\alpha}$, preserving discriminative ability. An equivalent formulation that makes the implicit dependence $\theta(\boldsymbol{\alpha})$ explicit is:
\begin{equation}
\label{eq:bilevel_v2}
\begin{aligned}
\min_{\boldsymbol{\alpha}} & \quad
\mathcal{L}_{\mathrm{ent}}^{\mathcal{S}_{\mathcal{V}}}(\boldsymbol{\alpha}, \theta(\boldsymbol{\alpha})) \\
\text{s.t.} & \quad \theta(\boldsymbol{\alpha}) \in \arg\min_{\theta'}
\;
\mathcal{L}_{\mathrm{tr}}^{\mathcal{S}_{\mathcal{T}}}(\theta';\boldsymbol{\alpha}).
\end{aligned}
\end{equation}

\textbf{First-order bilevel optimization.}
To optimize $\boldsymbol{\alpha}$ in Equation~(\ref{eq:bilevel_v2}), one must compute the hypergradient $\nabla_{\boldsymbol{\alpha}} \mathcal{L}_{\mathrm{ent}}(\boldsymbol{\alpha}, \theta(\boldsymbol{\alpha}))$. By the chain rule, this requires the implicit derivative $\nabla_{\boldsymbol{\alpha}} \theta(\boldsymbol{\alpha})$, which standard approaches obtain via the implicit function theorem:
\[
\nabla_{\boldsymbol{\alpha}} \theta(\boldsymbol{\alpha})
= -\left[\nabla^2_\theta \mathcal{L}_{\mathrm{tr}}(\theta;\boldsymbol{\alpha})\right]^{-1} \nabla_{\boldsymbol{\alpha}} \nabla_\theta \mathcal{L}_{\mathrm{tr}}(\theta;\boldsymbol{\alpha}).
\]
This involves second-order derivatives and a Hessian inversion of $\nabla^2_\theta \mathcal{L}_{\mathrm{tr}}$, which is computationally infeasible for LLMs with billions of parameters.

To circumvent this, we adopt a first-order approximation inspired by BOME~\citep{liu2022bome}. The key idea is to reformulate the bilevel problem as a constrained optimization over the joint variable $u = (\boldsymbol{\alpha}, \theta)$, where the lower-level optimality is enforced through a residual constraint $q(\boldsymbol{\alpha}, \theta) := \mathcal{L}_{\mathrm{tr}}(\theta; \boldsymbol{\alpha}) - \mathcal{L}_{\mathrm{tr}}^*(\boldsymbol{\alpha}) \le 0$, with $\mathcal{L}_{\mathrm{tr}}^*(\boldsymbol{\alpha}) := \min_{\theta'} \mathcal{L}_{\mathrm{tr}}(\theta'; \boldsymbol{\alpha})$ denoting the value function. Since the exact value function is intractable, we approximate it by running $T$ inner gradient steps from the current $\theta$ to obtain a surrogate $\tilde{\theta}$, and construct a plug-in residual $\hat{q} = \mathcal{L}_{\mathrm{tr}}(\theta; \boldsymbol{\alpha}) - \mathcal{L}_{\mathrm{tr}}(\tilde{\theta}; \boldsymbol{\alpha})$. The joint update direction combines the upper-level entropy gradient $\nabla_u \mathcal{L}_{\mathrm{ent}}$ with the residual gradient $\nabla_u \hat{q}$, requiring only first-order derivatives throughout. The complete algorithm is given in Appendix~\ref{sec:alg_sum}.
\vspace{-5pt}
\section{Experiment}
\subsection{Experimental Setup}

\textbf{Models.} We evaluate our methods using four open-source large language models: Llama-3.1-Tulu-8B~\citep{lambert2024t}, Vicuna-7B-v1.5~\citep{chiang2023vicuna}, Olmo 2-7B~\citep{olmo20242}, Mistral-7B~\citep{jiang2023mistral}, with detailed information provided in Appendix~\ref{sec:descriptions}.
We deliberately focus on these four models because each has undergone alignment using either RLHF~\citep{ouyang2022training} or DPO~\citep{rafailov2024direct} and exhibits notably poor calibration. Specifically, Vicuna-7B is aligned via RLHF, while Mistral-7B, Olmo 2-7B, and Llama-3.1-Tulu-8B are aligned via DPO.

\textbf{Baselines and ablations.}
We compare CALM against four external calibration methods and two ablations of CALM itself. The external baselines are:
\begin{itemize}[nosep]
\item \textbf{DPO/RLHF baseline}: The original preference-aligned model without calibration.
\item \textbf{Temperature scaling (TS)}~\citep{guo2017calibration}: Post-hoc scaling of logits by a single temperature $T$ fit on a validation set; the post-hoc method CALM generalizes.
\item \textbf{Label smoothing (LS)}~\citep{muller2019does}: Soft labels assigning $1-\varepsilon$ to the correct class and spreading $\varepsilon$ over the rest.
\item \textbf{Calibration-aware fine-tuning (CFT)}~\citep{xiao2025restoring}: Fine-tunes with a calibration-aware objective to restore calibration lost during alignment; as the closest training-time baseline, it receives a per-model tuning budget comparable to CALM's (Section~\ref{sec:general_qa}).
\end{itemize}
Two ablations isolate the contribution of CALM's bilevel structure:
\begin{itemize}[nosep]
\item \textbf{Regularization}: The entropy objective applied as a flat single-level penalty with a grid-tuned fixed weight; an ablation of CALM that removes the bilevel constraint.
\item \textbf{Iterate}: Alternates task and calibration updates without the coupling (implicit/residual) term; an ablation of CALM that removes the coupling.
\end{itemize}

\textbf{Evaluation settings.}
We evaluate all methods across two task families, multiple-choice QA (MCQA) and general open-ended QA, each under a domain-shifted (OOD) and a matched in-domain (ID) condition. In both families the OOD setting is our \emph{primary} evaluation, as it reflects the more realistic and challenging scenario, while the ID setting serves as a complementary reference.

\emph{Multiple-choice QA (MCQA).}
\begin{itemize}[nosep]
\item \textbf{Out-of-domain (OOD)}: The model is trained on the Alpaca dataset~\citep{alpaca} (a general question-answering dataset) and tested on MCQA\footnote{\url{https://huggingface.co/datasets/aaronwzl/mcqa_calibration_dataset}.}, a multiple-choice benchmark consisting of MMLU, MedMCQA, OpenBookQA, and ARC-Challenge (3{,}000 test samples). This setting tests the model's ability to \emph{transfer} calibration from one domain to another, highlighting zero-shot generalization.
\item \textbf{In-domain (ID)}: Both training and evaluation use the MCQA dataset (2{,}000 training / 1{,}000 Calibration / 2{,}000 test samples). This setting represents the best-case scenario where calibration and evaluation distributions are matched.
\end{itemize}

\emph{General open-ended QA.}
\begin{itemize}[nosep]
\item \textbf{Out-of-domain (OOD)}: Each method is trained on one open-ended QA dataset and evaluated on the other, using PopQA~\citep{mallen2023not} and TriviaQA~\citep{joshi-etal-2017-triviaqa}. The generative answer format is held fixed across training and evaluation, so the only source of shift is the underlying question distribution rather than the output format. This is a direct test of cross-domain calibration transfer under free-form generation.
\end{itemize}

Note that Label Smoothing is evaluated only in the ID setting of the MCQA family, since MCQA-OOD trains on Alpaca, which lacks the ground-truth MCQA labels that LS requires. Both open-ended QA benchmarks provide answer supervision, so LS is evaluated in the general-QA family as well.

\textbf{Evaluation metrics.}
For the MCQA family we report two calibration metrics: confidence Expected Calibration Error (conf-ECE)~\citep{guo2017calibration} and class-wise Expected Calibration Error (cw-ECE). For the general open-ended QA family, where answers are free-form and the label set is unbounded so the ECE variants above no longer apply, we report Sem-ECE~\citep{wang2026semantic}, which draws multiple samples from the model, clusters semantically equivalent generations, and treats each cluster's empirical frequency as the model's confidence, thereby overcoming the strict-format limitations of standard ECE in open-ended settings. All calibration metrics are lower-is-better. Across both families we additionally report Accuracy (higher is better) to confirm that calibration gains do not come at the expense of language understanding. Formal definitions are provided in Appendix~\ref{sec:descriptions}.

\textbf{Implementation details.}
We employ the Quantized Low Rank (QLoRA) technique~\citep{dettmers2024qlora} to fine-tune all models with rank = 64, QLoRA scaling parameter $\alpha = 32$, and \texttt{float16} precision on NVIDIA A100 (40G) GPUs. Training is conducted for 5 epochs with a batch size of 16 and a weight decay of 0.02. Hyperparameters for each method are selected via grid search based on the lowest ECE while maintaining reasonable accuracy ($>80\%$ of baseline); detailed configurations, grids, and training-cost comparisons are provided in Appendix~\ref{sec:hyperparams}.

\subsection{Multiple-Choice QA Results}
\label{sec:numerical_results}

\paragraph{OOD calibration (Primary Evaluation).}
Table~\ref{tab:calib_main} summarizes calibration quality and task accuracy across four aligned models in the OOD setting.

\textbf{CALM achieves the best conf-ECE on three of four models while preserving accuracy.}
On Llama-3.1 it reduces conf-ECE by 41.1\% ($0.1784\!\rightarrow\!0.1050$, $-2.35$ pp accuracy); on Vicuna-7B it attains the lowest conf-ECE ($0.0380$) and cw-ECE ($0.0576$) while marginally \emph{improving} accuracy; on OLMo-2-7B the best cw-ECE ($0.0912$, $-0.75$ pp); and on Mistral-7B the best conf-ECE ($0.0822$) with accuracy ($51.60\%$) above Regularization ($44.20\%$) and Iterate ($25.80\%$).

\textbf{Competing methods exhibit severe accuracy--calibration trade-offs.}
Iterate occasionally achieves strong ECE (e.g., Mistral cw-ECE 0.0518) but suffers catastrophic accuracy collapse (59.80\%$\rightarrow$25.80\%). Regularization preserves accuracy on some models but severely degrades it on others (Mistral: 44.20\%). CFT restores calibration relative to the base model on some backbones but leaves conf-ECE elevated ($\geq 0.17$ on all four). Temperature Scaling preserves accuracy by construction but its OOD calibration is inconsistent---it \emph{worsens} conf-ECE on LLaMA (0.1784$\rightarrow$0.2513). Notably, Regularization is exactly CALM's entropy objective as a grid-tuned single-level penalty, so its failure to match CALM at any grid point shows that no fixed penalty weight recovers what the bilevel constraint finds automatically---the coupling, not entropy alone, is necessary.

\textbf{CALM provides the strongest accuracy--calibration trade-off.}
Unlike methods that achieve low ECE only via utility collapse (Iterate) or fail to generalize across domains (TS-OOD), CALM reduces calibration error while retaining the highest accuracy among training-based methods. Two design choices explain this: the bilevel constraint prevents calibration from degrading task performance, and the entropy objective does not fit a particular test distribution, suiting the OOD setting.

\begin{table*}[t]
\centering
\caption{Performance comparison across four aligned models in the OOD MCQA setting. Best results (per model, per metric) are in \textbf{bold}. Lower is better for conf-ECE/cw-ECE ($\downarrow$), and higher is better for Accuracy ($\uparrow$).}
\label{tab:calib_main}
\small
\setlength{\tabcolsep}{2pt}
\resizebox{\textwidth}{!}{
\begin{tabular}{l|ccc|ccc|ccc|ccc}
\toprule
& \multicolumn{3}{c|}{Llama-3.1}
& \multicolumn{3}{c|}{Vicuna-7B}
& \multicolumn{3}{c|}{OLMo-2-7B}
& \multicolumn{3}{c}{Mistral-7B} \\
\cmidrule(lr){2-4}\cmidrule(lr){5-7}\cmidrule(lr){8-10}\cmidrule(lr){11-13}
\makecell{Method\\\\}
& \makecell{conf-\\ECE}$\downarrow$ & \makecell{cw-\\ECE}$\downarrow$ & Acc.$\uparrow$
& \makecell{conf-\\ECE}$\downarrow$ & \makecell{cw-\\ECE}$\downarrow$ & Acc.$\uparrow$
& \makecell{conf-\\ECE}$\downarrow$ & \makecell{cw-\\ECE}$\downarrow$ & Acc.$\uparrow$
& \makecell{conf-\\ECE}$\downarrow$ & \makecell{cw-\\ECE}$\downarrow$ & Acc.$\uparrow$ \\
\midrule
DPO / RLHF
& 0.1784 & 0.0931 & \textbf{66.80\%}
& 0.0581 & 0.0646 & 44.70\%
& 0.1115 & 0.0901 & \textbf{59.70\%}
& 0.3351 & 0.1717 & \textbf{59.80\%} \\

Temp.\ Scale.
& 0.2513 & 0.1291 & \textbf{66.80\%}
& 0.0953 & 0.0765 & 44.70\%
& \textbf{0.0608} & 0.0812 & \textbf{59.70\%}
& 0.2735 & 0.1418 & \textbf{59.80\%} \\

Regularization
& 0.1431 & 0.0900 & 66.50\%
& 0.0477 & 0.0591 & \textbf{45.70\%}
& 0.1394 & 0.1036 & 58.00\%
& 0.2100 & 0.1247 & 44.20\% \\

Iterate
& 0.2438 & 0.1317 & 39.55\%
& 0.3270 & 0.1754 & 31.50\%
& 0.2612 & 0.1420 & 58.90\%
& 0.0923 & \textbf{0.0518} & 25.80\% \\

CFT
& 0.1707 & 0.0925 & 64.05\%
& 0.1750 & 0.1130 & 42.15\%
& 0.1813 & 0.1033 & 62.50\%
& 0.2043 & 0.1372 & 49.30\% \\

CALM (Ours)
& \textbf{0.1050} & \textbf{0.0621} & 64.45\%
& \textbf{0.0380} & \textbf{0.0576} & 44.95\%
& 0.0878 & \textbf{0.0912} & 58.95\%
& \textbf{0.0822} & 0.0881 & 51.60\% \\
\bottomrule
\end{tabular}
}
\end{table*}

\paragraph{Robustness across seeds.}
A three-seed study (Appendix~\ref{sec:multiseed}, Table~\ref{tab:calib_seeds}) confirms that CALM's mean conf-ECE improves over the uncalibrated baseline on all four models. Vicuna-7B and OLMo-2-7B are highly stable (conf-ECE std $\le 0.002$); Llama-3.1 and Mistral-7B show larger run-to-run variance, so their single-run margins in Table~\ref{tab:calib_main} should be read with more caution.

\textbf{ID calibration (Complementary Reference).}
Table~\ref{tab:calib_id} presents the in-domain results, where both training and evaluation use MCQA data. Temperature Scaling achieves the best ECE on most models in the ID setting, as expected since it directly optimizes calibration on the matched distribution, but it cannot improve accuracy. Label Smoothing provides strong calibration on Olmo (ECE 0.0510, cw-ECE 0.0517) but degrades both calibration and accuracy on Mistral. CFT attains high accuracy on Vicuna (56.40\%) and Mistral (67.15\%) but its conf-ECE stays near $0.24$ on all four models. CALM (ID) is the only method that \emph{consistently improves accuracy} over the baseline across all four models (+1.3, +1.0, +2.5, +1.1 percentage points), while simultaneously improving calibration. Regularization (ID) achieves the highest accuracy gains (+5 to +10 percentage points) but at the cost of worsened calibration.

\begin{table*}[t]
\centering
\caption{Performance comparison across four aligned models in the ID MCQA setting. Best results (per model, per metric) are in \textbf{bold}.}
\label{tab:calib_id}
\small
\setlength{\tabcolsep}{2pt}
\resizebox{\textwidth}{!}{
\begin{tabular}{l|ccc|ccc|ccc|ccc}
\toprule
& \multicolumn{3}{c|}{Llama-3.1}
& \multicolumn{3}{c|}{Vicuna-7B}
& \multicolumn{3}{c|}{OLMo-2-7B}
& \multicolumn{3}{c}{Mistral-7B} \\
\cmidrule(lr){2-4}\cmidrule(lr){5-7}\cmidrule(lr){8-10}\cmidrule(lr){11-13}
\makecell{Method\\\\}
& \makecell{conf-\\ECE}$\downarrow$ & \makecell{cw-\\ECE}$\downarrow$ & Acc.$\uparrow$
& \makecell{conf-\\ECE}$\downarrow$ & \makecell{cw-\\ECE}$\downarrow$ & Acc.$\uparrow$
& \makecell{conf-\\ECE}$\downarrow$ & \makecell{cw-\\ECE}$\downarrow$ & Acc.$\uparrow$
& \makecell{conf-\\ECE}$\downarrow$ & \makecell{cw-\\ECE}$\downarrow$ & Acc.$\uparrow$ \\
\midrule
DPO / RLHF
& 0.1784 & 0.0931 & 66.80\%
& 0.0581 & 0.0646 & 44.70\%
& 0.1115 & 0.0901 & 59.70\%
& 0.3351 & 0.1717 & 59.80\% \\

Temp.\ Scale.
& \textbf{0.0224} & \textbf{0.0389} & 66.80\%
& \textbf{0.0294} & \textbf{0.0559} & 44.70\%
& 0.0587 & 0.0878 & 59.70\%
& \textbf{0.0560} & \textbf{0.0474} & 59.80\% \\

Label Smooth.
& 0.0495 & 0.0550 & 65.90\%
& 0.0649 & 0.0646 & 42.65\%
& \textbf{0.0510} & \textbf{0.0517} & 62.10\%
& 0.1898 & 0.1198 & 53.45\% \\

Regularization
& 0.2132 & 0.1095 & \textbf{74.20\%}
& 0.2470 & 0.1378 & 53.35\%
& 0.2355 & 0.1212 & \textbf{69.90\%}
& 0.2730 & 0.1393 & \textbf{67.00\%} \\

Iterate
& 0.2681 & 0.1348 & 67.50\%
& 0.4041 & 0.2110 & 48.95\%
& 0.2970 & 0.1494 & 66.15\%
& 0.3129 & 0.1627 & 58.85\% \\

CFT
& 0.2408 & 0.1213 & 71.70\%
& 0.2389 & 0.1351 & \textbf{56.40\%}
& 0.2374 & 0.1225 & 69.35\%
& 0.2450 & 0.1272 & \textbf{67.15\%} \\

CALM (Ours)
& 0.1752 & 0.0910 & \textbf{68.10\%}
& 0.0376 & 0.0536 & 45.65\%
& 0.0862 & 0.0765 & 62.20\%
& 0.2525 & 0.1328 & 60.85\% \\
\bottomrule
\end{tabular}
}
\end{table*}

\paragraph{Scalar vs.\ vector parameterization.} An ablation on Mistral-7B (OOD) shows CALM's per-vocabulary $\mathbf{d}_y,\mathbf{l}_y$ are far better calibrated than a collapsed scalar (learned-TS) variant (conf-ECE $0.0822$ vs.\ $0.3140$); details in Appendix~\ref{sec:ablation}.

\subsection{General Open-Ended QA Results}
\label{sec:general_qa}

We now move beyond the multiple-choice format to free-form generation, where LLM calibration matters most in practice. Table~\ref{tab:ood-calibration} reports the cross-domain generative test: each method is trained on one QA dataset and evaluated on the other (train PopQA, test TriviaQA), with the generative answer format held constant so that question-distribution shift is the only variable. Calibration is measured with Sem-ECE~\citep{wang2026semantic} (defined in the metrics paragraph above and Appendix~\ref{sec:descriptions}).

\textbf{CALM transfers best under generative domain shift.} CALM achieves the lowest mean Sem-ECE, $0.0671$, improving on DPO/RLHF ($0.0846$) and every training-time baseline. It is best on Llama-3.1 and Mistral-7B and within $0.001$ of the best on OLMo-2-7B; no baseline stays competitive across all four backbones. Label smoothing collapses under the shift (Sem-ECE up to $1.0$) and is reported only as a cautionary baseline. These gains preserve answer quality: CALM matches or exceeds DPO/RLHF accuracy on TriviaQA, so lower Sem-ECE reflects better-calibrated confidence over predictions of equal or higher quality.

\textbf{A tuning-budget-matched comparison with CFT.} In the generative setting the most relevant training-time baseline, CFT, must be adapted from its native MCQA design; to avoid under-serving it, we give CFT a per-model sweep over its calibration weight $\alpha\in\{0.1,0.5,0.9\}$, comparable in size to CALM's learning-rate grid (Appendix~\ref{sec:hyperparams}). Tuning helps: the best-$\alpha$ CFT improves mean Sem-ECE from $0.1330$ (fixed $\alpha=0.5$) to $0.1109$. However, even with a matched tuning budget, best-$\alpha$ CFT still trails both CALM ($0.0671$) and the uncalibrated baseline ($0.0846$) in this setting. The comparison is therefore fair and the conclusion is unchanged: CALM is the strongest method under generative domain shift.

\begin{table}[t]
\centering
\caption{Cross-domain calibration under generative distribution shift (train PopQA, test TriviaQA). Values are Sem-ECE (lower is better); best per column in \textbf{bold}. CFT is shown at its original fixed $\alpha=0.5$ and at the best per-model $\alpha\in\{0.1,0.5,0.9\}$ (Llama-3.1 $0.9$, Vicuna-7B $0.1$, OLMo-2-7B $0.1$, Mistral-7B $0.5$).}
\label{tab:ood-calibration}
\small
\setlength{\tabcolsep}{6pt}
\begin{tabular}{lccccc}
\toprule
Method & Llama-3.1 & Vicuna-7B & OLMo-2-7B & Mistral-7B & Mean \\
\midrule
DPO / RLHF             & 0.0512 & 0.1421 & \textbf{0.0716} & 0.0736 & 0.0846 \\
Reg                    & 0.0745 & \textbf{0.0762} & 0.1444 & 0.1110 & 0.1015 \\
CFT ($\alpha=0.5$)     & 0.1117 & 0.1532 & 0.1800 & 0.0870 & 0.1330 \\
CFT (best-$\alpha$)    & 0.0754 & 0.1215 & 0.1597 & 0.0870 & 0.1109 \\
LS (collapsed)         & 0.9167 & 1.0000 & 0.3049 & 0.3633 & 0.6462 \\
CALM (Ours)            & \textbf{0.0490} & 0.1105 & 0.0724 & \textbf{0.0366} & \textbf{0.0671} \\
\bottomrule
\end{tabular}
\end{table}

\subsection{Plot Analysis}
\begin{figure}[t]
    \centering
    \begin{subfigure}[b]{0.24\textwidth}
        \centering
        \includegraphics[width=\linewidth]{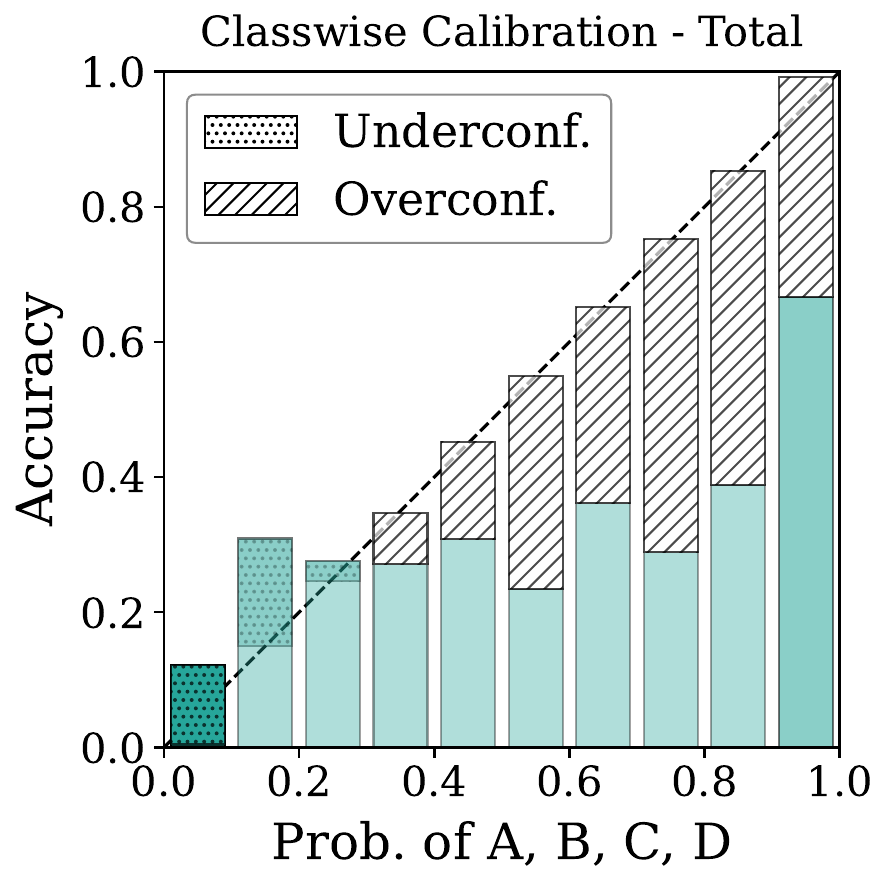}
        \caption{DPO baseline}
    \end{subfigure}
    \begin{subfigure}[b]{0.24\textwidth}
        \centering
        \includegraphics[width=\linewidth]{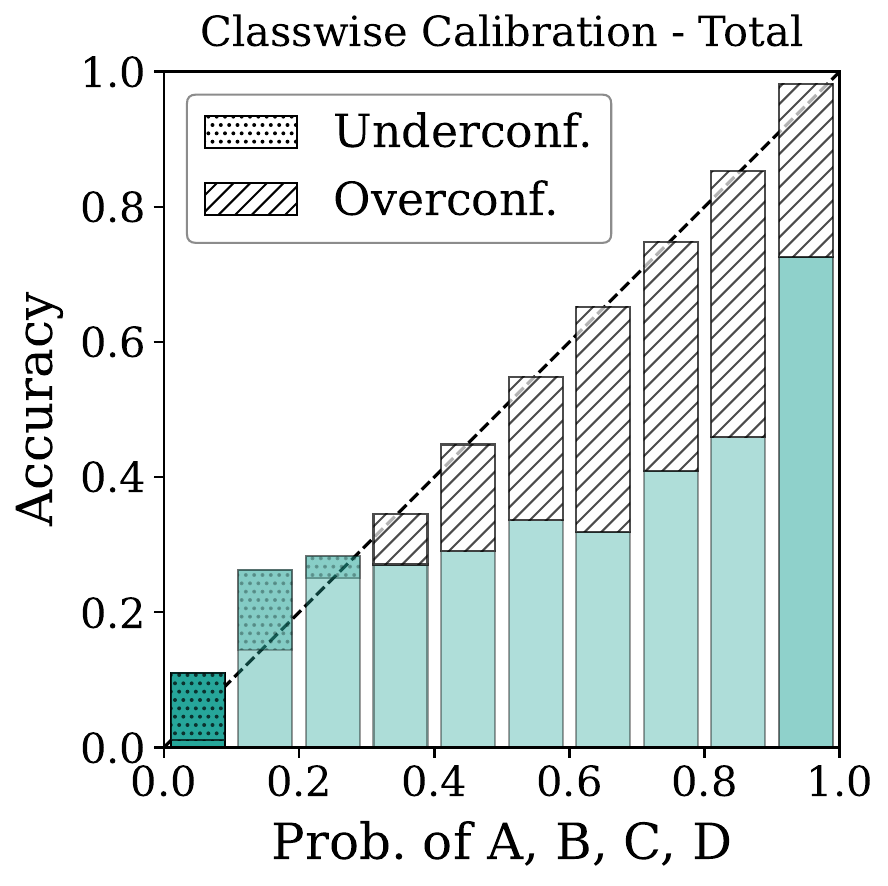}
        \caption{Temp.\ Scale.}
    \end{subfigure}
    \begin{subfigure}[b]{0.24\textwidth}
        \centering
        \includegraphics[width=\linewidth]{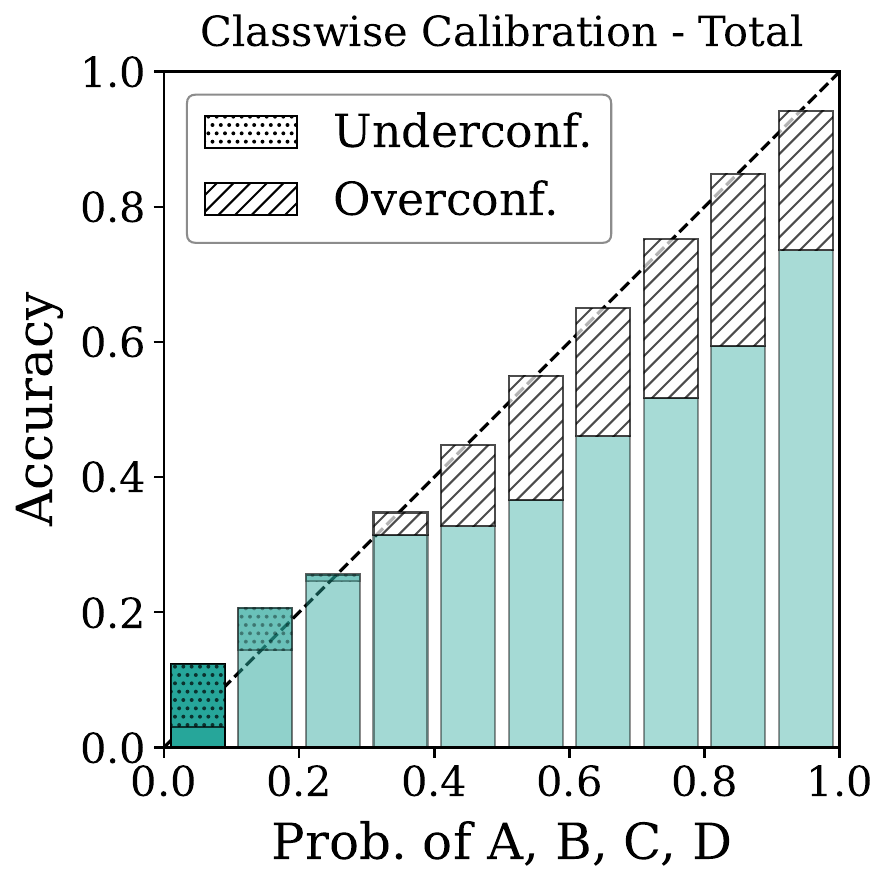}
        \caption{CFT}
    \end{subfigure}
    \begin{subfigure}[b]{0.24\textwidth}
        \centering
        \includegraphics[width=\linewidth]{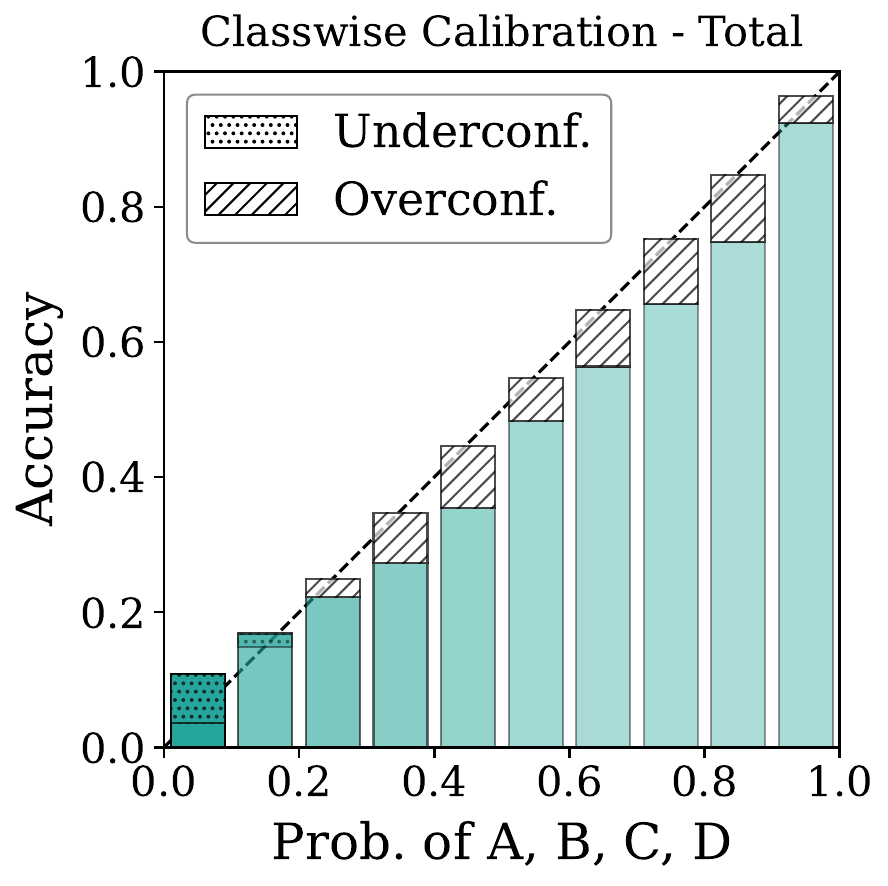}
        \caption{CALM (Ours)}
    \end{subfigure}
    \caption{Classwise reliability diagrams on Mistral-7B (OOD): (a) DPO baseline, (b) Temperature Scaling, (c) CFT, (d) CALM. Each panel plots predicted probability ($x$-axis) against observed accuracy ($y$-axis) over ten bins; the diagonal is perfect calibration and bars below it indicate overconfidence. Diagrams for all methods and all four models (OOD and ID) are in Appendix~\ref{sec:additional_results}.}
    \label{fig:cwece}
\end{figure}

Figure~\ref{fig:cwece} shows classwise reliability diagrams on Mistral-7B (OOD) for the DPO baseline, Temperature Scaling, CFT, and CALM. The DPO baseline is clearly overconfident in the medium- and high-confidence regions, where empirical correctness falls well below the predicted probability; Temperature Scaling and CFT reduce this only partially, with high-confidence bins still below the diagonal; and CALM aligns the bins closely with the diagonal without distorting the overall confidence profile. This confirms that overconfidence is the dominant failure mode and that CALM provides the most robust correction under distribution shift.

\subsection{Language Ability Retention}
\label{sec:language_retention}

A practical calibration method must also preserve language understanding. We evaluate all calibrated models on five benchmarks---HellaSwag~\citep{zellers2019hellaswag}, ARC-Easy, ARC-Challenge~\citep{clark2018think}, WinoGrande~\citep{sakaguchi2020winogrande}, and PIQA~\citep{bisk2020piqa}---with the LM Evaluation Harness~\citep{eval-harness} (zero-shot, log-likelihood scoring); Table~\ref{tab:language_retention} reports average accuracy under both settings. In the ID setting, CALM ($+0.2$ avg.), Iterate ($+0.2$), and Regularization ($+0.9$) all maintain or slightly improve over the baseline, confirming that when calibration and evaluation distributions are aligned, calibration training does not degrade general capabilities. Under the harder OOD setting, methods diverge: CALM is the most stable training-based method (near-zero change on OLMo-2-7B and Vicuna-7B, minor drops on Llama-3.1 and Mistral-7B), whereas Iterate regresses catastrophically ($-21.3$ avg.) and Label Smoothing degrades severely. Consequently, CALM has the smallest OOD-to-ID gap ($2.3$ on average, vs.\ Iterate $21.5$ and Regularization $4.0$), reflecting the bilevel framework's robustness to distribution shift.

\begin{table}[t]
\centering
\caption{Language ability retention: average accuracy (\%) over five standard benchmarks. Values in parentheses denote absolute changes relative to the baseline. Best accuracy-preserving result per model is \textbf{bold}.}
\label{tab:language_retention}
\small
\setlength{\tabcolsep}{7pt}
\begin{tabular}{llcccc}
\toprule
Setting & Method & Llama-3.1 & Vicuna-7B & OLMo-2-7B & Mistral-7B \\
\midrule
--- & Baseline & 72.82 & 64.87 & 71.33 & 69.02 \\
\midrule
\multirow{4}{*}{ID}
& CALM (Ours) & \textbf{72.97} \small{(+0.2)} & 64.92 \small{(+0.1)} & 71.34 \small{(+0.0)} & \textbf{69.60} \small{(+0.6)} \\
& Iterate & 72.96 \small{(+0.1)} & 64.39 \small{(-0.5)} & \textbf{71.76} \small{(+0.4)} & 69.79 \small{(+0.8)} \\
& Regularization & 73.03 \small{(+0.2)} & \textbf{64.98} \small{(+0.1)} & \textbf{71.76} \small{(+0.4)} & 71.72 \small{(+2.7)} \\
& Label Smooth. & 58.72 \small{(-14.1)} & 51.30 \small{(-13.6)} & 55.83 \small{(-15.5)} & 55.40 \small{(-13.6)} \\
\midrule
\multirow{4}{*}{OOD}
& CALM (Ours) & \textbf{66.99} \small{(-5.8)} & \textbf{64.59} \small{(-0.3)} & \textbf{71.29} \small{(-0.0)} & \textbf{66.94} \small{(-2.1)} \\
& Iterate & 56.44 \small{(-16.4)} & 44.31 \small{(-20.6)} & 57.65 \small{(-13.7)} & 34.64 \small{(-34.4)} \\
& Regularization & 70.13 \small{(-2.7)} & 65.27 \small{(+0.4)} & 69.33 \small{(-2.0)} & 60.61 \small{(-8.4)} \\
& Label Smooth. & 67.13 \small{(-5.7)} & 62.94 \small{(-1.9)} & 61.12 \small{(-10.2)} & 69.43 \small{(+0.4)} \\
\bottomrule
\end{tabular}
\end{table}

\section{Conclusion}

In this work, we proposed a bilevel framework for calibration-oriented loss design in aligned LLMs. Unlike post-hoc temperature scaling, which is dataset-dependent, our method improves calibration at training time by optimizing temperature-like loss hyperparameters with an entropy-based upper-level objective, together with a BOME-inspired first-order optimization scheme that makes the framework practical for LLM fine-tuning. Empirically, across both multiple-choice and open-ended generative question answering, our method achieves strong calibration gains across multiple aligned models while maintaining a better calibration--utility trade-off than competing baselines with especially clear advantages in out-of-domain settings where alternative methods often become unstable or suffer substantial utility loss.

\section*{Acknowledgments}
We thank the area chair and the anonymous reviewers for their comments and suggestions. This work was supported in part by the NUS School of Computing Start-up Grant to Jiancong Xiao.

\bibliography{reference}
\bibliographystyle{colm2026_conference}
\newpage

\appendix
\section{Algorithm Details}
\label{sec:alg_sum}

This section provides the detailed derivation and the complete algorithm for the first-order bilevel optimization scheme introduced in Section~\ref{sec:method}. We use the notation $\mathcal{L}_{\mathrm{tr}}^{\mathcal{S}_{\mathcal{T}}}(\theta;\boldsymbol{\alpha})$ and $\mathcal{L}_{\mathrm{ent}}^{\mathcal{S}_{\mathcal{V}}}(\boldsymbol{\alpha},\theta)$ as defined in Equation~(\ref{eq:bilevel}).

\subsection{Value Function Reformulation}

As discussed in the main text, the standard implicit-differentiation approach to bilevel optimization requires computing $\nabla_{\boldsymbol{\alpha}} \theta(\boldsymbol{\alpha})$ via the implicit function theorem, which involves the Hessian $\nabla^2_\theta \mathcal{L}_{\mathrm{tr}}$ and its inverse. To avoid this, we adopt a value-function-based reformulation~\citep{liu2022bome}.

We define the lower-level value function
\[
\mathcal{L}_{\mathrm{tr}}^\star(\boldsymbol{\alpha})
:=
\min_{\theta}
\mathcal{L}_{\mathrm{tr}}^{\mathcal{S}_{\mathcal{T}}}(\theta;\boldsymbol{\alpha}),
\]
and the associated optimality residual
\begin{equation}
q(\boldsymbol{\alpha},\theta)
:=
\mathcal{L}_{\mathrm{tr}}^{\mathcal{S}_{\mathcal{T}}}(\theta;\boldsymbol{\alpha})
-
\mathcal{L}_{\mathrm{tr}}^\star(\boldsymbol{\alpha}).
\label{eq:residual}
\end{equation}
Note that $q(\boldsymbol{\alpha},\theta) \ge 0$ by construction, and $q(\boldsymbol{\alpha},\theta) = 0$ if and only if $\theta$ is a minimizer of $\mathcal{L}_{\mathrm{tr}}$ for the given $\boldsymbol{\alpha}$. Therefore, the bilevel problem in Equation~(\ref{eq:bilevel}) can be equivalently rewritten as the following constrained optimization over the joint variable $u = (\boldsymbol{\alpha}, \theta)$:
\begin{equation}
\min_{u}
\;
\mathcal{L}_{\mathrm{ent}}^{\mathcal{S}_{\mathcal{V}}}(u)
\qquad
\text{subject to}
\qquad
q(\boldsymbol{\alpha},\theta)\le 0.
\label{eq:joint_problem}
\end{equation}
This constrained form makes explicit that each update should both improve the upper-level entropy objective and move the current iterate toward lower-level optimality. Crucially, by Danskin's theorem, $\nabla_{\boldsymbol{\alpha}} \mathcal{L}_{\mathrm{tr}}^\star(\boldsymbol{\alpha}) = \nabla_{\boldsymbol{\alpha}} \mathcal{L}_{\mathrm{tr}}(\theta^*(\boldsymbol{\alpha}); \boldsymbol{\alpha})$, which does \emph{not} require computing $\nabla_{\boldsymbol{\alpha}} \theta^*(\boldsymbol{\alpha})$. Hence, the gradient of the residual $q$ with respect to $u$ involves only first-order derivatives.

\subsection{Plug-in Residual Approximation}

In practice, the exact value function $\mathcal{L}_{\mathrm{tr}}^\star(\boldsymbol{\alpha})$ is intractable because it requires solving the lower-level problem to optimality. At outer iteration $k$, we instead approximate the lower-level optimum by running $T$ gradient descent steps on $\mathcal{L}_{\mathrm{tr}}$ starting from the current parameter $\theta_k$, while keeping $\boldsymbol{\alpha}_k$ fixed:
\[
\theta_k^{(t+1)}
=
\theta_k^{(t)}
-
\rho\,
\nabla_\theta
\mathcal{L}_{\mathrm{tr}}^{\mathcal{S}_{\mathcal{T}}}(\theta_k^{(t)};\boldsymbol{\alpha}_k),
\qquad
t=0,\dots,T-1,
\]
with $\theta_k^{(0)} = \theta_k$. We denote the resulting approximate optimum by $\tilde{\theta}_k := \theta_k^{(T)}$ and define the plug-in residual surrogate
\begin{equation}
\hat q_k(\boldsymbol{\alpha},\theta)
=
\mathcal{L}_{\mathrm{tr}}^{\mathcal{S}_{\mathcal{T}}}(\theta;\boldsymbol{\alpha})
-
\mathcal{L}_{\mathrm{tr}}^{\mathcal{S}_{\mathcal{T}}}(\tilde{\theta}_k;\boldsymbol{\alpha}_k),
\label{eq:plug_in}
\end{equation}
where $\tilde{\theta}_k$ is treated as a constant when differentiating. This surrogate satisfies $\hat q_k \ge q$ (since $\tilde{\theta}_k$ is not an exact minimizer), providing a relaxation of the exact constraint.

\begin{algorithm}[t]
\caption{CALM: First-Order Bilevel Optimization for LLM Calibration}
\label{alg:joint_bilevel}
\begin{algorithmic}[1]
\Require Training set $\mathcal{S}_{\mathcal{T}}$, validation inputs $\mathcal{S}_{\mathcal{V}}$, initial parameters $(\boldsymbol{\alpha}_0, \theta_0)$, outer iterations $K$, inner steps $T$, learning rates $\eta_{\boldsymbol{\alpha}}, \eta_\theta, \rho$, balancing coefficient $\gamma$
\For{$k = 0, 1, \dots, K-1$}
    \State \textbf{(Inner optimization)} Starting from $\theta^{(0)} \leftarrow \theta_k$, run $T$ steps:
    \[
    \theta^{(t+1)}
    =
    \theta^{(t)}
    -
    \rho\,
    \nabla_\theta
    \mathcal{L}_{\mathrm{tr}}^{\mathcal{S}_{\mathcal{T}}}(\theta^{(t)};\boldsymbol{\alpha}_k),
    \quad
    t = 0, \dots, T-1
    \]
    \State Set $\tilde{\theta}_k \leftarrow \theta^{(T)}$
    \State \textbf{(Residual surrogate)} $\hat q_k(\boldsymbol{\alpha}, \theta) = \mathcal{L}_{\mathrm{tr}}^{\mathcal{S}_{\mathcal{T}}}(\theta; \boldsymbol{\alpha}) - \mathcal{L}_{\mathrm{tr}}^{\mathcal{S}_{\mathcal{T}}}(\tilde{\theta}_k; \boldsymbol{\alpha}_k)$
    \State \textbf{(Gradients)} $g_{\mathrm{ent}} = \nabla_u \mathcal{L}_{\mathrm{ent}}^{\mathcal{S}_{\mathcal{V}}}(u_k)$, \quad $g_q = \nabla_u \hat q_k(u_k)$
    \State \textbf{(Adaptive weight)} $\lambda_k = \max\!\left(\frac{\gamma \|g_q\|^2 - \langle g_q, g_{\mathrm{ent}} \rangle}{\|g_q\|^2},\; 0\right)$
    \State \textbf{(Joint direction)} $d_k = g_{\mathrm{ent}} + \lambda_k\, g_q$; split $d_k = (d_k^{\boldsymbol{\alpha}}, d_k^\theta)$
    \State \textbf{(Update)} $\boldsymbol{\alpha}_{k+1} \leftarrow \boldsymbol{\alpha}_k - \eta_{\boldsymbol{\alpha}}\, d_k^{\boldsymbol{\alpha}}$, \quad $\theta_{k+1} \leftarrow \theta_k - \eta_\theta\, d_k^\theta$
\EndFor
\State \Return $(\boldsymbol{\alpha}_K, \theta_K)$
\end{algorithmic}
\end{algorithm}

\subsection{Joint Update Direction}

Using the plug-in residual, we compute two gradient directions:
\[
g_{\mathrm{ent}}
=
\nabla_u \mathcal{L}_{\mathrm{ent}}^{\mathcal{S}_{\mathcal{V}}}(u_k),
\qquad
g_q
=
\nabla_u \hat q_k(u_k).
\]
The first is the gradient of the upper-level entropy objective, and the second is the gradient of the approximate residual. Following the BOME framework, the joint update direction is constructed to simultaneously decrease $\mathcal{L}_{\mathrm{ent}}$ and the residual $\hat{q}$. Specifically, we set the dynamic barrier
\[
\phi_k = \gamma \|g_q\|^2,
\]
where $\gamma > 0$ is a balancing coefficient, and compute the adaptive weight
\[
\lambda_k
=
\max\!\left(
\frac{\phi_k - \langle g_q, g_{\mathrm{ent}} \rangle}{\|g_q\|^2},\;
0
\right),
\]
with $\lambda_k = 0$ if $\|g_q\| = 0$. The joint search direction is then
\[
d_k = g_{\mathrm{ent}} + \lambda_k\, g_q.
\]
Intuitively, when the entropy gradient $g_{\mathrm{ent}}$ already makes sufficient progress toward reducing the residual (i.e., $\langle g_q, g_{\mathrm{ent}} \rangle \ge \phi_k$), we have $\lambda_k = 0$ and the update follows the entropy gradient alone. Otherwise, $\lambda_k > 0$ adds a correction that steers the update toward lower-level feasibility.

Writing $d_k = (d_k^{\boldsymbol{\alpha}}, d_k^\theta)$, the parameters are updated as
\[
\boldsymbol{\alpha}_{k+1}
=
\boldsymbol{\alpha}_k - \eta_{\boldsymbol{\alpha}}\, d_k^{\boldsymbol{\alpha}},
\qquad
\theta_{k+1}
=
\theta_k - \eta_\theta\, d_k^\theta,
\]
where $\eta_{\boldsymbol{\alpha}}$ and $\eta_\theta$ are the learning rates for the calibration hyperparameters and model parameters, respectively.

\subsection{Complete Algorithm}

Algorithm~\ref{alg:joint_bilevel} summarizes the complete procedure. Each outer iteration consists of two phases: (i)~a short inner optimization that runs $T$ gradient steps on $\mathcal{L}_{\mathrm{tr}}$ to construct the plug-in residual surrogate $\hat{q}_k$, and (ii)~a joint update of $\boldsymbol{\alpha}$ and $\theta$ along the combined direction $d_k$ that balances entropy reduction with approximate lower-level optimality.

\section{Models, Datasets, and Metrics}
\label{sec:descriptions}

This section provides descriptions of the models, datasets, and evaluation metrics used in this paper. For training configurations and hyperparameter details, see Appendix~\ref{sec:hyperparams}.

\paragraph{Models.}
We evaluate on four open-source LLMs that have undergone preference alignment:
\begin{itemize}
    \item \textbf{LLaMA-3.1-Tulu-8B-DPO}~\citep{lambert2024t}\footnote{\url{https://huggingface.co/allenai/Llama-3.1-Tulu-3-8B}}: An instruction-following model from Allen Institute for AI, trained using supervised fine-tuning and DPO. Part of the Tulu 3 family, designed for diverse tasks.
    \item \textbf{Vicuna-7B-v1.5}~\citep{chiang2023vicuna}\footnote{\url{https://huggingface.co/lmsys/vicuna-7b-v1.5}}: A chat assistant from LMSYS, fine-tuned from LLaMA 2 on approximately 125,000 user-shared conversations via RLHF.
    \item \textbf{OLMo 2-7B-DPO}~\citep{olmo20242}\footnote{\url{https://huggingface.co/allenai/OLMo-2-1124-7B-DPO}}: A fully open-source model from Allen Institute for AI, trained on the Dolma dataset and fine-tuned using DPO.
    \item \textbf{Mistral 7B-DPO}~\citep{jiang2023mistral}\footnote{\url{https://huggingface.co/princeton-nlp/Mistral-7B-Base-SFT-DPO}}: A high-performance model fine-tuned using DPO, known for strong efficiency and benchmark results.
\end{itemize}
All models are aligned via either RLHF (Vicuna) or DPO (the others), and exhibit notable overconfidence after alignment, making them suitable targets for calibration evaluation.

\paragraph{Datasets.}
Our calibration experiments use two datasets:
\begin{itemize}
    \item \textbf{Alpaca}~\citep{alpaca}: A question-answering dataset of 52,002 instruction-response pairs. Used for out-of-domain (OOD) training.
    \item \textbf{MCQA}: A multiple-choice question-answering benchmark\footnote{\url{https://huggingface.co/datasets/aaronwzl/mcqa_calibration_dataset}} aggregating questions from four sources:
    \begin{itemize}
        \item \emph{MMLU}~\citep{hendrycks2021measuring}: 57-subject knowledge and reasoning benchmark covering STEM, humanities, and social sciences.
        \item \emph{MedMCQA}~\citep{pal2022medmcqa}: Medical multiple-choice questions from AIIMS and NEET PG entrance exams.
        \item \emph{OpenBookQA}~\citep{mihaylov2018can}: Elementary-level science questions requiring common knowledge beyond provided facts.
        \item \emph{ARC-Challenge}~\citep{clark2018think}: Advanced science questions from grades 3--9, requiring multi-step reasoning.
    \end{itemize}
    The MCQA dataset contains 3,000 training samples, from which we extract 1,000 validation samples, and 2,000 test samples. It is used for both in-domain (ID) training and for evaluating all methods.
    \item \textbf{PopQA}~\citep{mallen2023not}: An open-domain, entity-centric question-answering benchmark with short free-form answers. Used in the general open-ended QA family.
    \item \textbf{TriviaQA}~\citep{joshi-etal-2017-triviaqa}: A large-scale open-domain trivia question-answering benchmark with free-form answers. Used in the general open-ended QA family.
\end{itemize}
For the general open-ended QA family, each method is trained on one of \{PopQA, TriviaQA\} and evaluated on the other, holding the free-form answer format fixed so that only the question distribution shifts, yielding a pure cross-domain calibration-transfer test under generation.

\paragraph{Metrics.}
We use several evaluation metrics. Accuracy measures the fraction of correctly answered questions (higher is better). For multiple-choice QA we use the two calibration metrics defined below (conf-ECE and cw-ECE); for open-ended generation we use Sem-ECE. All calibration metrics are lower-is-better.

\textbf{Confidence expected calibration error (conf-ECE)}~\citep{guo2017calibration} measures the alignment between a model's maximum predicted probability and its empirical accuracy:
\begin{equation}
    \text{conf-ECE} = \sum_{m=1}^M \frac{|B_{m}|}{N}\left|\text{acc}(B_m) - \text{conf}(B_m)\right|,
\end{equation}
where $\{B_m\}_{m=1}^M$ partitions the predictions into $M$ bins based on the maximum predicted probability, $|B_m|$ is the number of samples in bin $m$, $\text{acc}(B_m)$ is the empirical accuracy in that bin, and $\text{conf}(B_m)$ is the average confidence (maximum predicted probability) in that bin.

\textbf{Class-wise expected calibration error (cw-ECE)} extends conf-ECE by evaluating calibration separately for each class:
\begin{equation}
    \text{cw-ECE} = \frac{1}{K}\sum_{j=1}^K\sum_{m=1}^M \frac{|B_{m,j}|}{N}\left|\text{acc}_j(B_{m,j}) - \text{conf}_j(B_{m,j})\right|,
\end{equation}
where $B_{m,j}$ is the $m$-th bin of the $j$-th class based on the predicted probability for class $j$, $\text{acc}_j(B_{m,j})$ is the fraction of samples in that bin whose true label is $j$, and $\text{conf}_j(B_{m,j})$ is the average predicted probability for class $j$ in that bin. While conf-ECE measures calibration only for the most confident class, cw-ECE provides a more comprehensive assessment by checking whether the predicted probability for \emph{every} class matches its empirical frequency.

\textbf{Semantic expected calibration error (Sem-ECE)}~\citep{wang2026semantic} extends ECE to open-ended generation, where the answer is free-form and the label set is unbounded so conf-ECE/cw-ECE no longer apply. For each query, the model generates multiple samples; semantically equivalent generations are clustered, and the empirical frequency of the selected answer's cluster is taken as the model's confidence. Sem-ECE then bins these cluster-frequency confidences against empirical correctness exactly as in conf-ECE:
\begin{equation}
    \text{Sem-ECE} = \sum_{m=1}^M \frac{|B_m|}{N}\left|\text{acc}(B_m) - \bar{p}(B_m)\right|,
\end{equation}
where $\bar{p}(B_m)$ is the average cluster-frequency confidence in bin $m$ and $\text{acc}(B_m)$ is the fraction of samples in that bin whose sampled answer is semantically correct. This overcomes the strict-format limitation of standard ECE, which cannot bin free-form outputs.

\subsection{Other Related Work}

\paragraph{Post-training and model confidence.}
Preference-based post-training methods, including PPO-based RLHF~\citep{schulman2017proximal}, DPO and its variants~\citep{azar2024general,chang2024dataset,gorbatovski2024learn,rafailov2024r,yang2024asymptotics}, and Nash learning from human feedback~\citep{munos2023nash,shi2025fundamental,wang2025magnetic,xiao2025theoretical}, can substantially alter the output distributions of language models and hence their confidence. Prior work has studied the limitations of PPO~\citep{li2023remax}, the gap between online and offline alignment algorithms~\citep{tang2024understanding}, and the effects of preference-model specification and distribution shift~\citep{li2023policy,ye2024theoretical}. Related studies have also examined diversity during supervised fine-tuning~\citep{li2025preserving} and model editing as an alternative mechanism for adapting model behavior~\citep{jin2025finetuning}. Our work is complementary to these approaches: rather than proposing a new alignment objective, we study how to calibrate models after their predictive distributions have been modified by post-training.

\paragraph{Calibration and recalibration.}
Temperature scaling is a widely used post-hoc calibration method and is often more effective than label smoothing and related training-time techniques~\citep{guo2017calibration,muller2019does}. For language models, \citet{zhao2021calibrate} introduced contextual calibration to mitigate biases in few-shot prediction. A broader line of work has investigated the interaction between predictive accuracy and calibration~\citep{kumar2018trainable,krishnan2020improving,karandikar2021soft,popordanoska2022consistent}, as well as proper calibration errors and minimum-risk recalibration~\citep{gruber2022better,sun2023minimum}. The reliability of calibration evaluation has also received theoretical attention, particularly regarding bias in ECE estimators based on uniform-mass binning~\citep{gupta2020distribution,gupta2021distribution}. In contrast to methods that optimize a single calibration loss or fit a fixed post-hoc transformation, our approach formulates calibration as a bilevel problem: an entropy-based objective suppresses overconfidence, while the task objective preserves predictive discrimination. This interaction allows the calibration parameters and model adaptation to be optimized jointly rather than treating calibration as an isolated post-processing step.
\section{Implementation Details}
\label{sec:hyperparams}

This section provides comprehensive implementation details for all calibration methods evaluated in this paper.

\subsection{Common Training Configuration}

All training-based methods (CALM, Iterate, Regularization, Label Smoothing, CFT) share the following base configuration:

\begin{itemize}[nosep]
    \item \textbf{Parameter-efficient fine-tuning}: QLoRA~\citep{dettmers2024qlora} with rank $r=64$, scaling factor $\alpha=32$, and dropout $0.01$.
    \item \textbf{QLoRA target modules}: All linear layers in the transformer (excluding the \texttt{lm\_head} output projection), identified automatically via module introspection.
    \item \textbf{Precision}: \texttt{float16} mixed precision.
    \item \textbf{Optimizer}: AdamW with $(\beta_1, \beta_2)=(0.9, 0.95)$.
    \item \textbf{Learning rate schedule}: Cosine annealing with linear warmup (2 epochs for bilevel methods, 1 epoch for label smoothing).
    \item \textbf{Hardware}: NVIDIA A100 (40G) GPUs.
\end{itemize}

\subsection{Method-Specific Configurations}

\paragraph{CALM (Ours).}
CALM uses the first-order BOME (Bilevel Optimization Made Easy)~\citep{liu2022bome} scheme described in Algorithm~\ref{alg:joint_bilevel}, which avoids explicit hypergradient, implicit-gradient, and Hessian-inverse computation. The key training parameters are:

\begin{itemize}[nosep]
    \item \textbf{Bilevel structure}: The lower-level optimizes model parameters (QLoRA weights) to minimize cross-entropy loss, while the upper-level optimizes the per-vocabulary calibration parameters of Section~\ref{sec:method}---the multiplicative logit adjustment $\mathbf{d}_y \in \mathbb{R}^{V}$ (equivalently $\tau_y^{-1}$) and the additive logit adjustment $\mathbf{l}_y \in \mathbb{R}^{V}$---to maximize predictive entropy on held-out inputs subject to a lower-level task-optimality constraint. Together these generalize the single global temperature $T$ of standard temperature scaling, which is recovered when the $\mathbf{d}_y$ are tied.
    \item \textbf{Inner optimization steps}: 20 batches per epoch for the lower-level surrogate $\hat{x}$.
    \item \textbf{BOME coefficient}: $\eta = 0.5$ for the bilevel descent direction $d = \nabla f + \text{ReLU}\!\left(\eta - \frac{\nabla f \cdot \nabla g}{\|\nabla g\|^2}\right) \nabla g$.
    \item \textbf{Epochs}: 5 (OOD) / 5 (ID).
    \item \textbf{Batch size}: 16.
    \item \textbf{Weight decay}: 0.02.
    \item \textbf{Parameter clipping}: All trainable parameters clipped to $[-10, 10]$.
\end{itemize}

\paragraph{Temperature scaling.}
Temperature scaling is a post-hoc method that does not modify any model weights. A single scalar temperature $T$ is optimized to minimize the negative log-likelihood on a validation set:
\begin{itemize}[nosep]
    \item \textbf{Optimization}: Bounded scalar minimization via \texttt{scipy.optimize.minimize\_scalar} with L-BFGS-B method.
    \item \textbf{Search bounds}: $T \in [0.1, 10.0]$.
    \item \textbf{Tolerance}: $10^{-4}$.
    \item \textbf{Data split}: 50\% of the calibration data for finding $T$, 50\% for evaluation.
\end{itemize}

\paragraph{Label smoothing.}
Label smoothing uses adaptive soft labels based on per-bin accuracy statistics:
\begin{itemize}[nosep]
    \item \textbf{Smoothing strategy}: For each sample, the smoothing parameter $\varepsilon$ is determined by the model's confidence bin: $\varepsilon = 1 - \text{bin\_accuracy}(\text{confidence})$. The correct class receives probability $\text{bin\_accuracy}$, and the remaining probability is distributed uniformly.
    \item \textbf{Learning rate}: $10^{-4}$ (fixed, no grid search).
    \item \textbf{Epochs}: 5.
    \item \textbf{Weight decay}: 0.02.
    \item \textbf{Note}: Only evaluated in the ID setting, since it requires ground-truth MCQA labels for computing bin accuracies. The OOD variant using synthetic MCQA from Alpaca is invalid, as the correct answer is always option A with trivial distractors, leading to accuracy collapse.
\end{itemize}

\paragraph{Calibration-aware fine-tuning (CFT).}
CFT~\citep{xiao2025restoring} augments the supervised loss with a calibration-aware term of the form $\mathcal{L}=(1-\alpha)\,\mathcal{L}_{\text{CE}}+\alpha\cdot c\cdot\mathcal{L}_{\text{cal}}$, where $\alpha$ is the calibration weight, $c$ a fixed scale, and $\mathcal{L}_{\text{cal}}$ a per-bin objective that drives predicted confidence toward empirical bin accuracy.
\begin{itemize}[nosep]
    \item \textbf{Calibration weight}: $\alpha$, swept in the general open-ended QA setting (Table~\ref{tab:hyperparam_grid}); fixed at $\alpha=0.5$ for the MCQA experiments.
    \item \textbf{Calibration scale}: $c=10^{-3}$ (fixed).
    \item \textbf{Learning rate}: $10^{-4}$. \textbf{Epochs}: 5. \textbf{Weight decay}: 0.02.
    \item \textbf{Architecture}: Standard QLoRA fine-tuning (no \texttt{Logit\_2level} wrapper).
\end{itemize}

\paragraph{Regularization.}
The regularization baseline augments the standard cross-entropy loss with a confidence penalty term:
\begin{itemize}[nosep]
    \item \textbf{Loss}: $\mathcal{L} = \mathcal{L}_{\text{CE}}(\mathbf{y}, \hat{\mathbf{y}}) - \alpha \cdot \mathcal{L}_{\text{CE}}(\hat{\mathbf{y}}, \hat{\mathbf{y}})$, where $\hat{\mathbf{y}}$ denotes the model's predicted labels. The second term encourages higher entropy (lower confidence) in predictions.
    \item \textbf{Epochs}: 30 (OOD) / 5 (ID).
    \item \textbf{Weight decay}: 0.02.
    \item \textbf{Architecture}: Standard QLoRA fine-tuning without the \texttt{Logit\_2level} wrapper (no multiplicative/additive logit adjustments).
\end{itemize}

\paragraph{Iterate.}
The Iterate method alternates between lower-level (task) and upper-level (calibration) optimization without computing implicit gradients:
\begin{itemize}[nosep]
    \item \textbf{Lower level}: 20 gradient steps per epoch on the cross-entropy loss with hyperparameters fixed.
    \item \textbf{Upper level}: 1 gradient step per epoch on the calibration loss ($0.9 \cdot \mathcal{L}_{\text{CE}}(\mathbf{y}, \hat{\mathbf{y}}) - \mathcal{L}_{\text{CE}}(\hat{\mathbf{y}}_{\text{pred}}, \hat{\mathbf{y}})$) with model parameters frozen.
    \item \textbf{Epochs}: 5 (OOD and ID).
    \item \textbf{Weight decay}: 0.02.
    \item \textbf{Key difference from CALM}: No implicit gradient computation; hyperparameters are updated using direct gradients only.
\end{itemize}

\subsection{Hyperparameter Search Strategy}

Table~\ref{tab:hyperparam_grid} summarizes the hyperparameter search grid for each method. The best configuration per model is selected based on the lowest conf-ECE while maintaining accuracy above 80\% of the baseline.

\begin{table}[h]
\centering
\caption{Hyperparameter search grids for each method.}
\label{tab:hyperparam_grid}
\small
\setlength{\tabcolsep}{4pt}
\begin{tabular}{lll}
\toprule
Method & Hyperparameters & Search Grid \\
\midrule
CALM (OOD) & Lower LR $\times$ Upper LR & $\{10^{-3}, 10^{-2}, 10^{-1}\} \times \{10^{-4}, 10^{-5}, 10^{-6}, 10^{-7}\}$ \\
CALM (ID) & Lower LR $\times$ Upper LR & $\{10^{-3}, 10^{-2}\} \times \{10^{-5}, 10^{-6}, 10^{-7}\}$ \\
Iterate & Lower LR $\times$ Upper LR & $\{10^{-3}, 10^{-2}\} \times \{10^{-4}, 10^{-5}, 10^{-6}, 10^{-7}\}$ \\
Regularization & LR $\times$ $\alpha$ & $\{10^{-3}, 10^{-4}\} \times \{10^{-5}, 10^{-6}\}$ \\
Temp. Scaling & Temperature $T$ & $T \in [0.1, 10.0]$ (continuous optimization) \\
Label Smoothing & Learning rate & $10^{-4}$ (fixed) \\
CFT (generative) & Calibration weight $\alpha$ & $\{0.1, 0.5, 0.9\}$ (lr $10^{-4}$, scale $10^{-3}$ fixed) \\
\bottomrule
\end{tabular}
\end{table}

\subsection{Best Hyperparameters Per Model}

Tables~\ref{tab:best_calm}--\ref{tab:best_temp} report the best hyperparameters selected for each model under each method.

\begin{table}[h]
\centering
\caption{Best CALM hyperparameters per model (selected by lowest conf-ECE with accuracy $>80\%$ of baseline).}
\label{tab:best_calm}
\small
\begin{tabular}{llcccc}
\toprule
Setting & Model & Lower LR & Upper LR & conf-ECE & Accuracy \\
\midrule
\multirow{4}{*}{OOD}
& Llama-3.1 & 0.01 & $10^{-4}$ & 0.1050 & 64.45\% \\
& Vicuna-7B & 0.001 & $10^{-5}$ & 0.0380 & 44.95\% \\
& OLMo-2-7B & 0.001 & $10^{-4}$ & 0.0878 & 58.95\% \\
& Mistral-7B & 0.01 & $10^{-6}$ & 0.0822 & 51.60\% \\
\midrule
\multirow{4}{*}{ID}
& Llama-3.1 & 0.01 & $10^{-5}$ & 0.1752 & 68.10\% \\
& Vicuna-7B & 0.01 & $10^{-5}$ & 0.0376 & 45.65\% \\
& OLMo-2-7B & 0.01 & $10^{-6}$ & 0.0862 & 62.20\% \\
& Mistral-7B & 0.01 & $10^{-6}$ & 0.2525 & 60.85\% \\
\bottomrule
\end{tabular}
\end{table}

\begin{table}[h]
\centering
\caption{Best Regularization hyperparameters per model.}
\label{tab:best_reg}
\small
\begin{tabular}{llcccc}
\toprule
Setting & Model & LR & $\alpha$ & conf-ECE & Accuracy \\
\midrule
\multirow{4}{*}{OOD}
& Llama-3.1 & $10^{-4}$ & $10^{-5}$ & 0.1431 & 66.50\% \\
& Vicuna-7B & $10^{-4}$ & $10^{-6}$ & 0.0477 & 45.70\% \\
& OLMo-2-7B & $10^{-4}$ & $10^{-6}$ & 0.1394 & 58.00\% \\
& Mistral-7B & $10^{-3}$ & $10^{-5}$ & 0.2100 & 44.20\% \\
\midrule
\multirow{4}{*}{ID}
& Llama-3.1 & $10^{-4}$ & $10^{-5}$ & 0.2132 & 74.20\% \\
& Vicuna-7B & $10^{-4}$ & $10^{-5}$ & 0.2470 & 53.35\% \\
& OLMo-2-7B & $10^{-4}$ & $10^{-5}$ & 0.2355 & 69.90\% \\
& Mistral-7B & $10^{-4}$ & $10^{-5}$ & 0.2730 & 67.00\% \\
\bottomrule
\end{tabular}
\end{table}

\begin{table}[h]
\centering
\caption{Optimal temperature values for Temperature Scaling.}
\label{tab:best_temp}
\small
\begin{tabular}{llccc}
\toprule
Setting & Model & Optimal $T$ & conf-ECE & Accuracy \\
\midrule
\multirow{4}{*}{ID}
& Llama-3.1 & 2.1828 & 0.0224 & 66.80\% \\
& Vicuna-7B & 1.4170 & 0.0294 & 44.70\% \\
& OLMo-2-7B & 1.9804 & 0.0587 & 59.70\% \\
& Mistral-7B & 6.4410 & 0.0560 & 59.80\% \\
\midrule
\multirow{4}{*}{OOD}
& Llama-3.1 & 0.5316 & 0.2513 & 66.80\% \\
& Vicuna-7B & 0.8468 & 0.0953 & 44.70\% \\
& OLMo-2-7B & 1.4835 & 0.0608 & 59.70\% \\
& Mistral-7B & 2.0007 & 0.2735 & 59.80\% \\
\bottomrule
\end{tabular}
\end{table}

\subsection{Robustness and Ablation Protocols}
\paragraph{Multi-seed runs.} For the robustness study (Table~\ref{tab:calib_seeds}), we retrain CALM in the OOD MCQA setting with three random seeds at each model's selected hyperparameters (Table~\ref{tab:best_calm}) and report the mean and standard deviation of conf-ECE, cw-ECE, and accuracy. All other settings are held fixed.
\paragraph{Scalar vs.\ vector ablation.} To isolate the effect of the per-vocabulary parameterization, we re-run CALM on Mistral (OOD) with $\mathbf{d}_y$ and $\mathbf{l}_y$ collapsed to scalars (size $1$), recovering a learned temperature-scaling-style adjustment; all other settings match the full CALM run.

\subsection{Language Ability Evaluation Details}

To evaluate language ability retention (Section~\ref{sec:language_retention}), we use the LM Evaluation Harness~\citep{eval-harness} (v0.4.12) with the following configuration:
\begin{itemize}[nosep]
    \item \textbf{Benchmarks}: HellaSwag~\citep{zellers2019hellaswag}, ARC-Easy, ARC-Challenge~\citep{clark2018think}, WinoGrande~\citep{sakaguchi2020winogrande}, PIQA~\citep{bisk2020piqa}.
    \item \textbf{Evaluation mode}: Zero-shot, log-likelihood scoring. For each example, the model scores candidate answers by computing the log-probability of the answer text conditioned on the context. The candidate with the highest log-likelihood is selected.
    \item \textbf{Metric}: Accuracy (fraction of correctly answered examples). For HellaSwag, ARC-Easy, ARC-Challenge, and PIQA, we use length-normalized accuracy (\texttt{acc\_norm}); for WinoGrande, we use standard accuracy (\texttt{acc}).
    \item \textbf{Precision}: \texttt{float16}.
    \item \textbf{Batch size}: Automatically determined based on available GPU memory.
\end{itemize}

\subsection{Dataset Details}

\paragraph{Training Data.}
\begin{itemize}[nosep]
    \item \textbf{OOD setting}: Alpaca~\citep{alpaca} dataset (52{,}002 instruction-response pairs). We use the first 48{,}002 samples for training, the next 2{,}000 for validation, and the last 2{,}000 for testing.
    \item \textbf{ID setting}: MCQA calibration dataset\footnote{\url{https://huggingface.co/datasets/aaronwzl/mcqa_calibration_dataset}} containing 4-way multiple choice questions drawn from MMLU, MedMCQA, OpenBookQA, and ARC-Challenge. We use the \texttt{calibration\_train} split (3{,}000 training / 1{,}000 validation) and the \texttt{test\_1} split (2{,}000 samples) for testing.
    \item \textbf{Open-ended QA}: PopQA~\citep{mallen2023not} and TriviaQA~\citep{joshi-etal-2017-triviaqa}. Each method is trained on one dataset and evaluated on the other under a fixed free-form answer format. For Sem-ECE we draw 20 samples per question over 300 evaluation questions, cluster semantically equivalent generations, and use cluster frequency as confidence, following \citet{wang2026semantic}.
\end{itemize}

\paragraph{Model Details.}
Table~\ref{tab:model_details} summarizes the four models used in our experiments.

\begin{table}[h]
\centering
\caption{Model specifications.}
\label{tab:model_details}
\small
\begin{tabular}{lccc}
\toprule
Model & Parameters & Vocabulary Size & Alignment \\
\midrule
Llama-3.1-Tulu-3-8B-DPO & 8B & 128{,}256 & DPO \\
Vicuna-7B-v1.5 & 7B & 32{,}000 & RLHF \\
OLMo-2-1124-7B-DPO & 7B & 100{,}352 & DPO \\
Mistral-7B-Instruct-DPO & 7B & 32{,}000 & DPO \\
\bottomrule
\end{tabular}
\end{table}

\subsection{Training Cost Comparison}

Table~\ref{tab:training_cost} reports the approximate training time and peak GPU memory for each calibration method on a 7--8B parameter model using a single NVIDIA A100 (40G) GPU. Temperature Scaling requires no training and has negligible cost. Among training-based methods, CALM is the most expensive due to the bilevel structure (inner optimization loop and model surrogate), but remains practical: a full training run completes within a few hours on a single GPU. Regularization and Label Smoothing use standard single-level fine-tuning and are correspondingly cheaper. The Iterate method is the fastest training-based method since it performs only a few alternating gradient steps per epoch without maintaining a model copy.

\begin{table}[h]
\centering
\caption{Approximate training cost per model on a single NVIDIA A100 (40G) GPU. Training time is for a complete run with the default number of epochs. Peak memory includes the quantized model, optimizer states, and any surrogate copies.}
\label{tab:training_cost}
\small
\begin{tabular}{lccc}
\toprule
Method & Epochs & Training Time & Peak GPU Memory \\
\midrule
Temperature Scaling & --- & $<$1 min & $\sim$15 GB \\
Iterate & 5 & $\sim$5 min & $\sim$20 GB \\
Label Smoothing & 5 & $\sim$10 min & $\sim$20 GB \\
Regularization & 5 & $\sim$35 min & $\sim$20 GB \\
CALM (Ours) & 5 & $\sim$2 hr & $\sim$38 GB \\
\bottomrule
\end{tabular}
\end{table}

The higher cost of CALM relative to single-level methods is attributable to two factors: (i)~at each outer iteration, $T=20$ inner gradient steps are performed on a surrogate model to approximate the lower-level solution, and (ii)~the surrogate model $\tilde{\theta}$ must be maintained in memory alongside the primary model $\theta$. We note that this overhead is a one-time training cost; at inference time, CALM introduces no additional latency compared to standard models.

\section{Additional Results}
\label{sec:additional_results}

\subsection{Multi-Seed Robustness (OOD MCQA)}
\label{sec:multiseed}
Because the single-run margins in Table~\ref{tab:calib_main} could in principle be affected by random variation, we repeat OOD-MCQA training for CALM with three random seeds at each model's selected configuration (Table~\ref{tab:best_calm}) and report the mean~$\pm$~standard deviation in Table~\ref{tab:calib_seeds}. Vicuna-7B and OLMo-2-7B are highly stable (conf-ECE std $\le 0.002$), confirming that their single-run margins are reliable. Llama-3.1 and Mistral-7B exhibit larger run-to-run variance; the single-run values in Table~\ref{tab:calib_main} fall within the observed seed range, but we report the full distribution for transparency and caution that these two margins are less tight than the point estimates suggest. Even under this conservative accounting, the mean CALM conf-ECE improves over the uncalibrated baseline on all four models (e.g., Mistral-7B $0.335\!\to\!0.149$, Llama-3.1 $0.178\!\to\!0.110$).

\begin{table}[h]
\centering
\caption{Multi-seed robustness of CALM in the OOD MCQA setting (mean~$\pm$~std over 3 seeds). Vicuna-7B/OLMo-2-7B are highly stable; Llama-3.1/Mistral-7B show larger variance. Baseline conf-ECE is shown for reference.}
\label{tab:calib_seeds}
\small
\setlength{\tabcolsep}{6pt}
\begin{tabular}{lcccc}
\toprule
Model & conf-ECE $\downarrow$ & cw-ECE $\downarrow$ & Accuracy (\%) $\uparrow$ & Baseline conf-ECE \\
\midrule
Llama-3.1  & $0.110 \pm 0.085$ & $0.089 \pm 0.023$ & $61.9 \pm 3.2$ & 0.1784 \\
Vicuna-7B  & $0.040 \pm 0.002$ & $0.058 \pm 0.001$ & $45.0 \pm 0.1$ & 0.0581 \\
OLMo-2-7B  & $0.089 \pm 0.002$ & $0.091 \pm 0.001$ & $59.0 \pm 0.1$ & 0.1115 \\
Mistral-7B & $0.149 \pm 0.121$ & $0.105 \pm 0.044$ & $51.1 \pm 0.7$ & 0.3351 \\
\bottomrule
\end{tabular}
\end{table}

\subsection{Scalar vs.\ Vector Calibration Parameters}
\label{sec:ablation}
CALM generalizes the single scalar temperature of TS to per-vocabulary multiplicative ($\mathbf{d}_y$) and additive ($\mathbf{l}_y$) logit adjustments. To isolate the contribution of this vector parameterization, we re-run CALM on Mistral-7B (OOD) with $\mathbf{d}_y,\mathbf{l}_y$ collapsed to scalars, which recovers a learned temperature-scaling-style adjustment. As shown in Table~\ref{tab:scalar_vector}, the vector form is dramatically better calibrated (conf-ECE $0.0822$ vs.\ $0.3140$; cw-ECE $0.0881$ vs.\ $0.1618$) at a small accuracy cost ($51.6\%$ vs.\ $59.9\%$). Miscalibration in aligned LLMs is asymmetric across answer options: a scalar can only rescale overall sharpness, whereas per-vocabulary parameters can redistribute confidence between options.

\begin{table}[h]
\centering
\caption{Scalar vs.\ vector calibration parameters for CALM on Mistral-7B (OOD). Lower is better for conf-ECE/cw-ECE.}
\label{tab:scalar_vector}
\small
\begin{tabular}{lccc}
\toprule
Parameterization & conf-ECE $\downarrow$ & cw-ECE $\downarrow$ & Accuracy (\%) $\uparrow$ \\
\midrule
Scalar (learned TS)             & 0.3140 & 0.1618 & 59.9 \\
Vector $\mathbf{d}_y,\mathbf{l}_y$ (CALM) & \textbf{0.0822} & \textbf{0.0881} & 51.6 \\
\bottomrule
\end{tabular}
\end{table}

\begin{figure}[htbp]
    \centering
    \textbf{DPO/RLHF}\\[1pt]
    \begin{subfigure}[b]{0.21\textwidth}\centering\includegraphics[width=\linewidth]{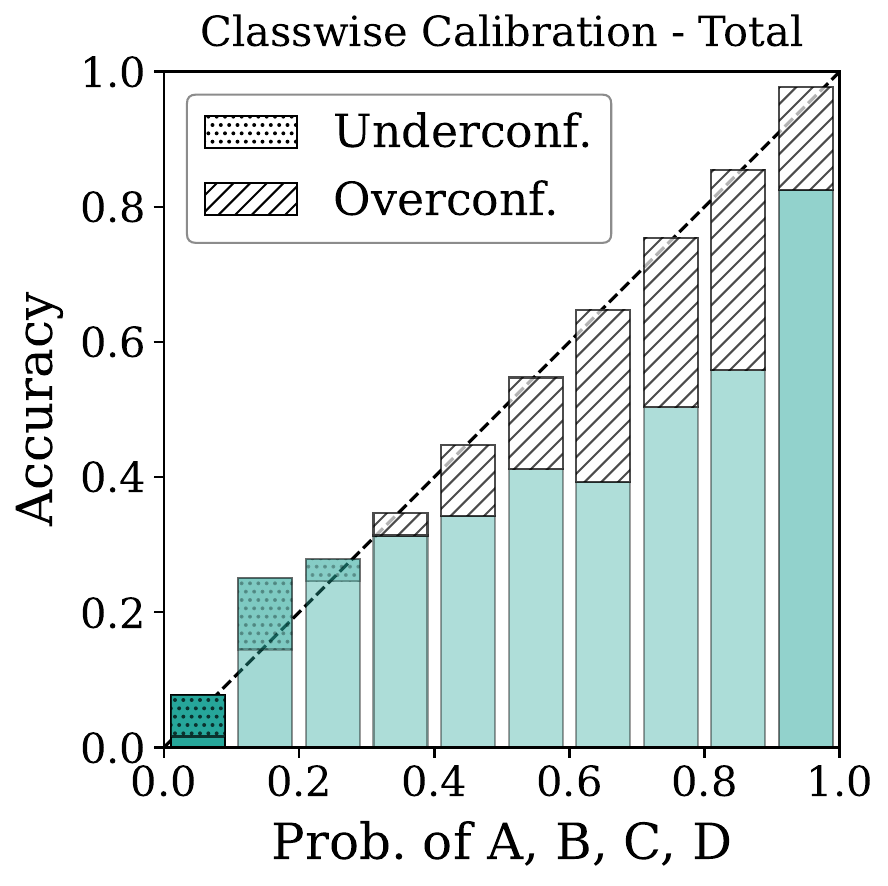}\caption{Llama-3.1}\end{subfigure}
    \begin{subfigure}[b]{0.21\textwidth}\centering\includegraphics[width=\linewidth]{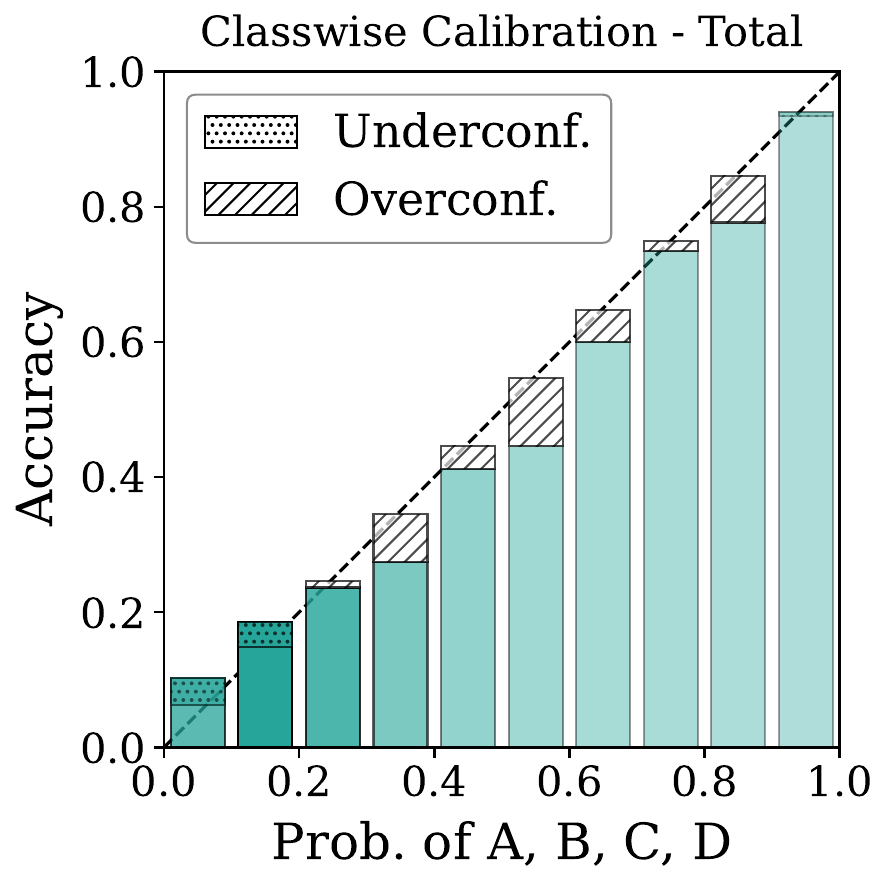}\caption{Vicuna-7B}\end{subfigure}
    \begin{subfigure}[b]{0.21\textwidth}\centering\includegraphics[width=\linewidth]{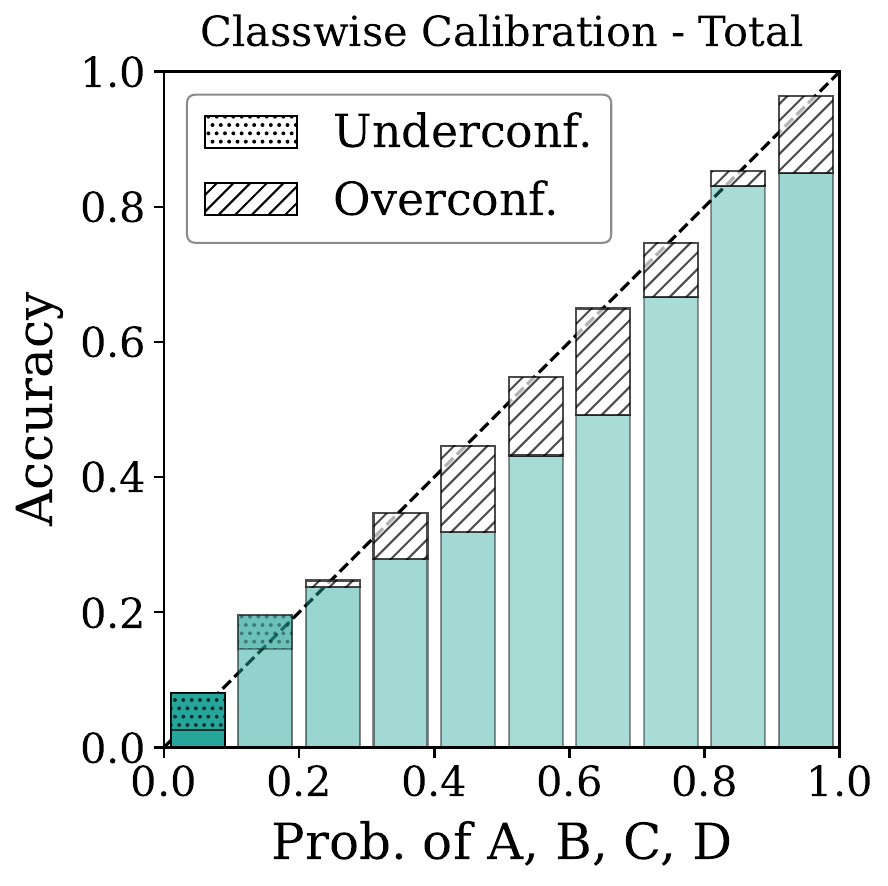}\caption{OLMo-2-7B}\end{subfigure}
    \begin{subfigure}[b]{0.21\textwidth}\centering\includegraphics[width=\linewidth]{figures/baseline_cwece_total.pdf}\caption{Mistral-7B}\end{subfigure}
    \\[3pt]
    \textbf{Temp.\ Scale.}\\[1pt]
    \begin{subfigure}[b]{0.21\textwidth}\centering\includegraphics[width=\linewidth]{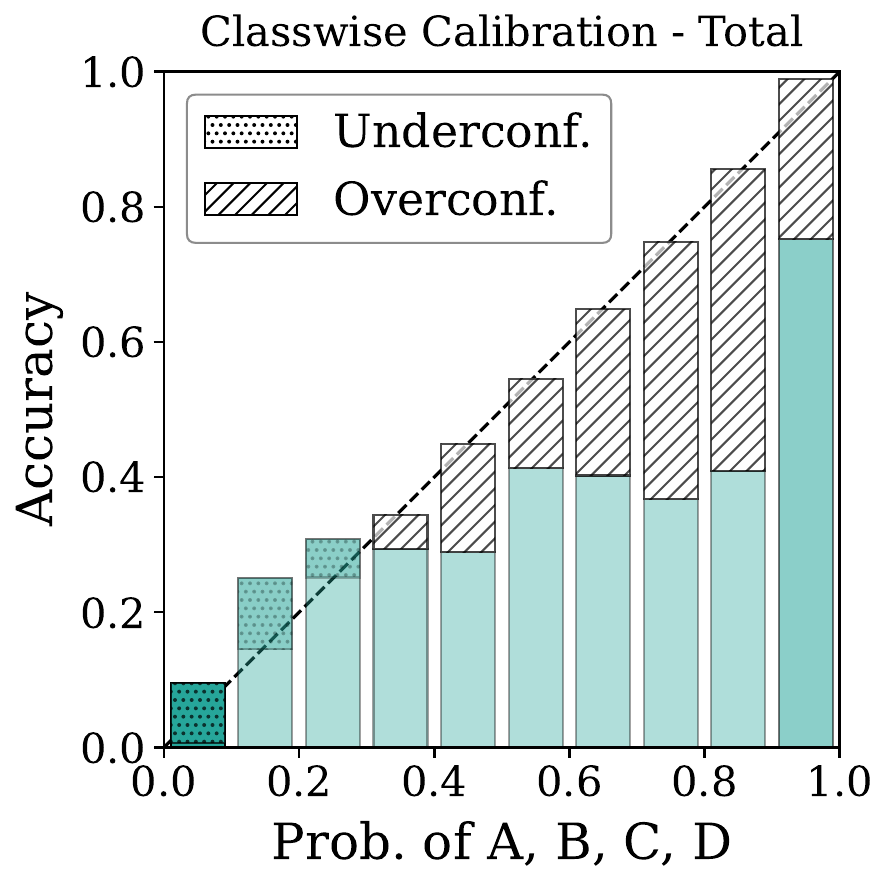}\end{subfigure}
    \begin{subfigure}[b]{0.21\textwidth}\centering\includegraphics[width=\linewidth]{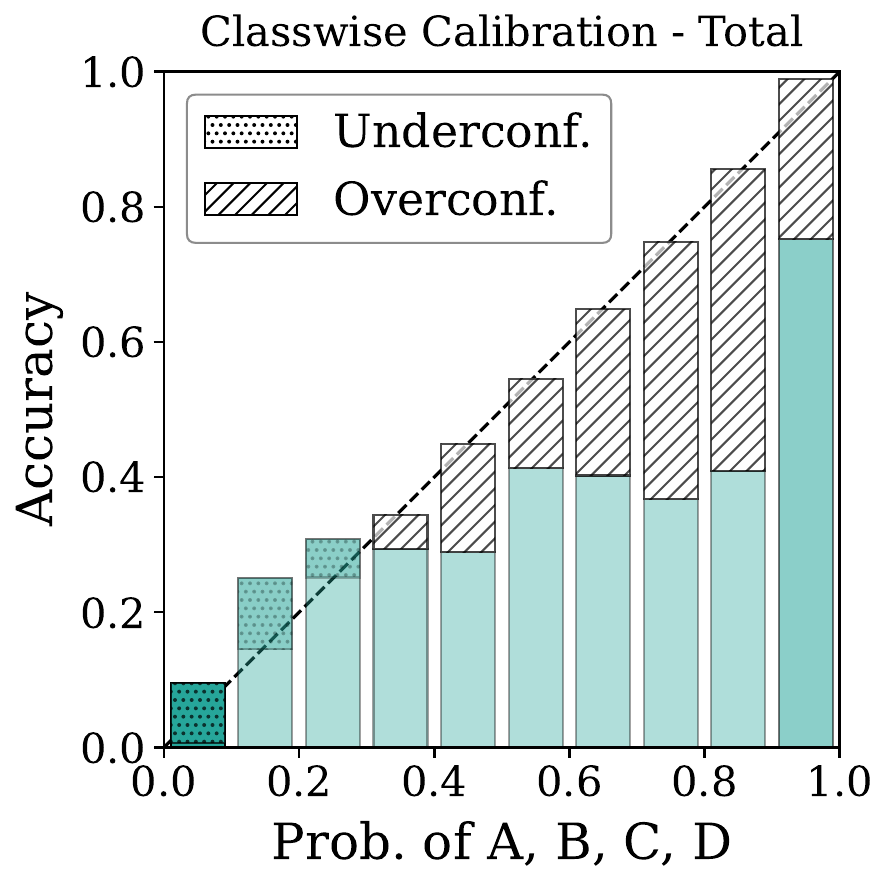}\end{subfigure}
    \begin{subfigure}[b]{0.21\textwidth}\centering\includegraphics[width=\linewidth]{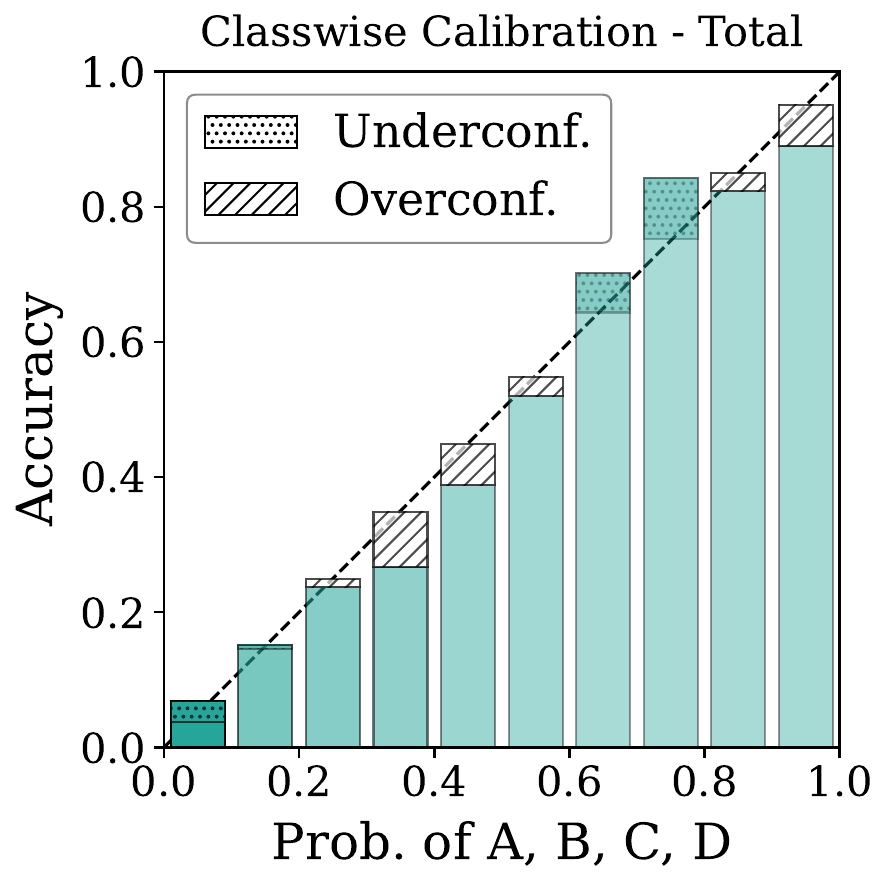}\end{subfigure}
    \begin{subfigure}[b]{0.21\textwidth}\centering\includegraphics[width=\linewidth]{figures/temperature_scaling_ood_cwece_total.pdf}\end{subfigure}
    \\[3pt]
    \textbf{Regularization}\\[1pt]
    \begin{subfigure}[b]{0.21\textwidth}\centering\includegraphics[width=\linewidth]{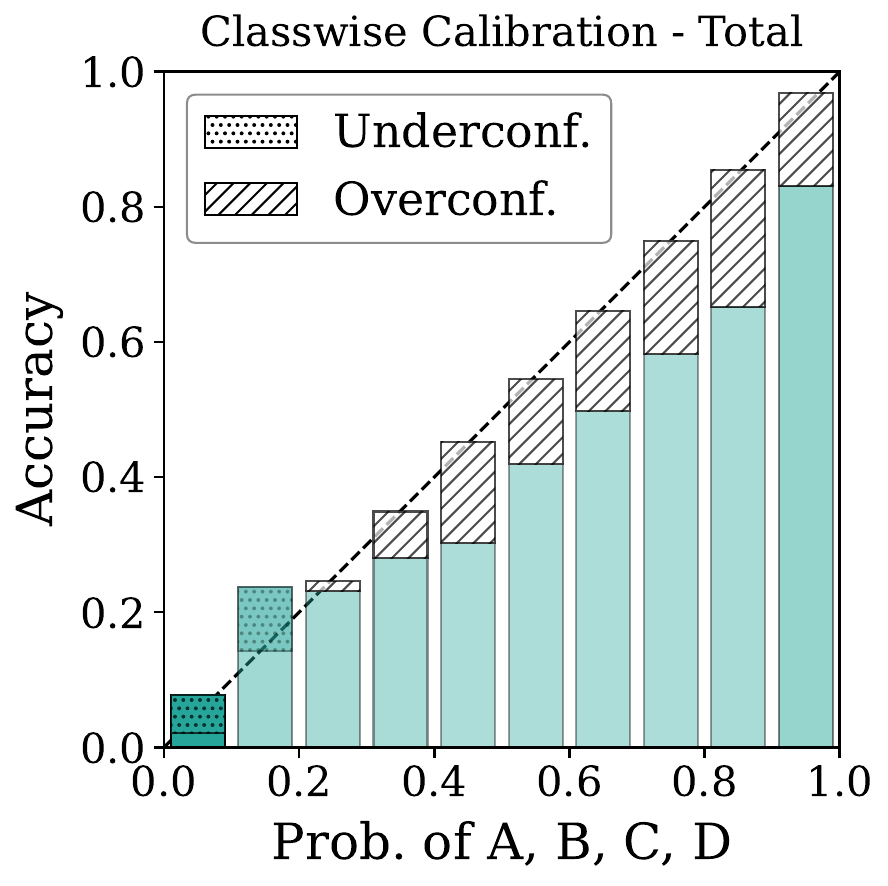}\end{subfigure}
    \begin{subfigure}[b]{0.21\textwidth}\centering\includegraphics[width=\linewidth]{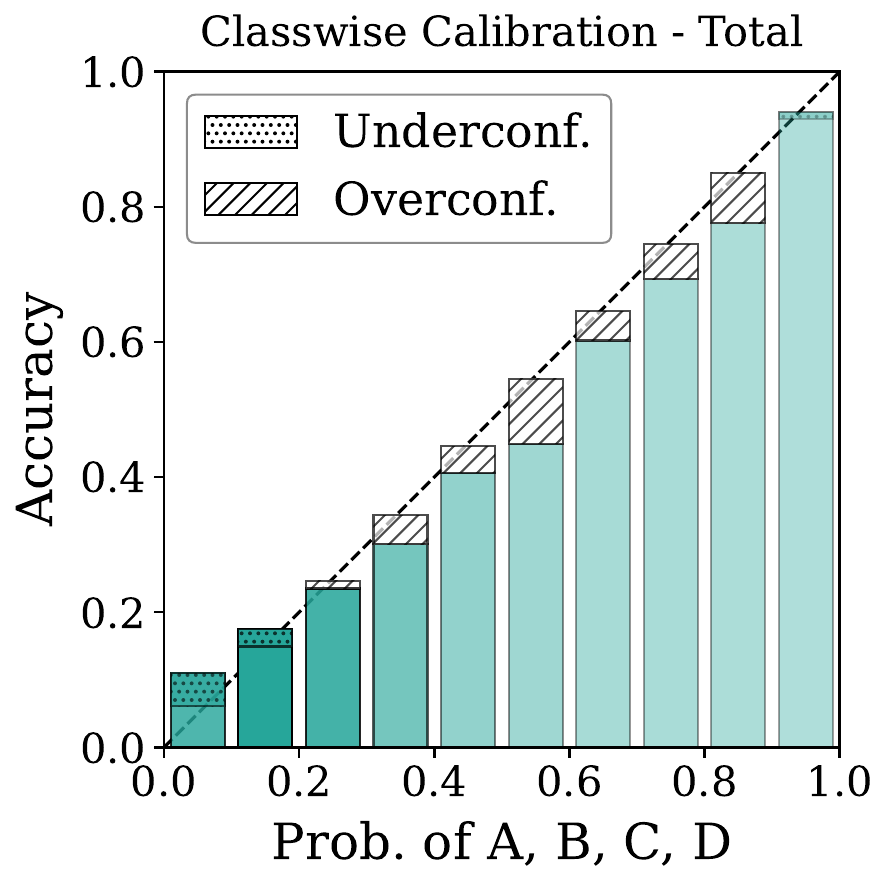}\end{subfigure}
    \begin{subfigure}[b]{0.21\textwidth}\centering\includegraphics[width=\linewidth]{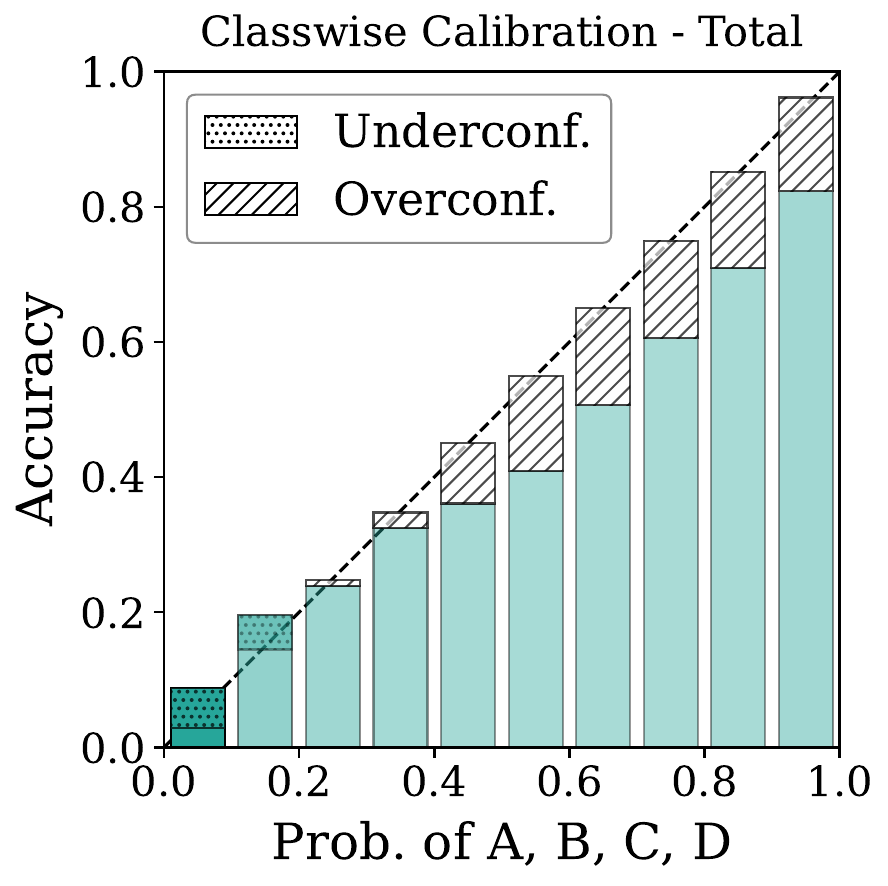}\end{subfigure}
    \begin{subfigure}[b]{0.21\textwidth}\centering\includegraphics[width=\linewidth]{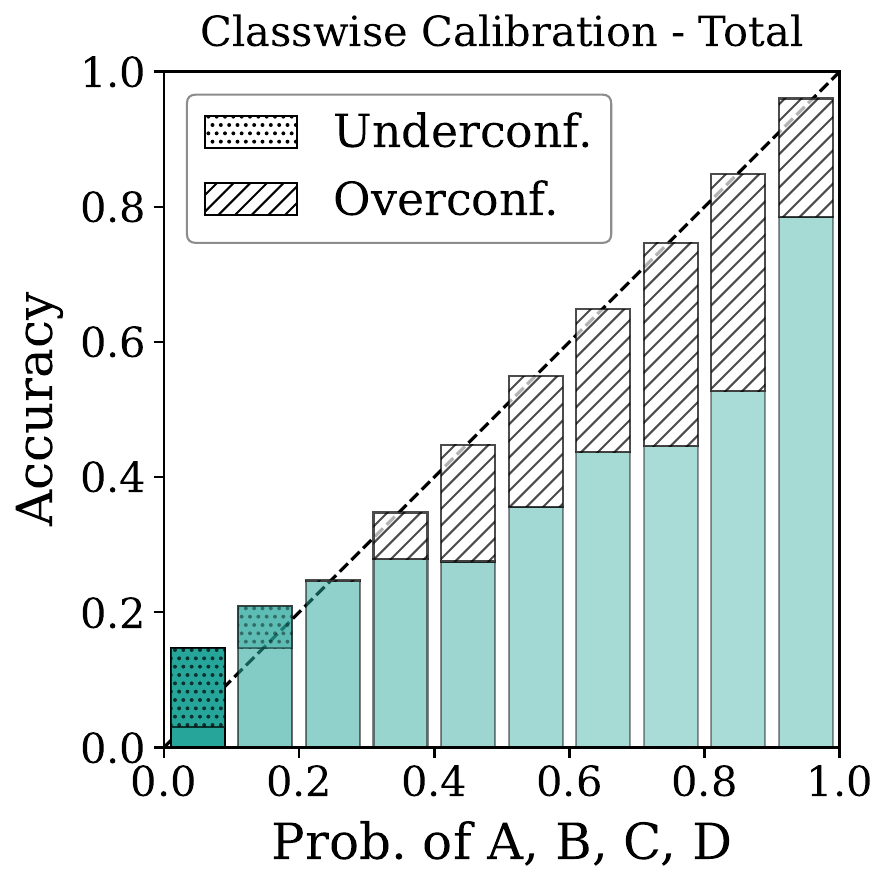}\end{subfigure}
    \\[3pt]
    \textbf{Iterate}\\[1pt]
    \begin{subfigure}[b]{0.21\textwidth}\centering\includegraphics[width=\linewidth]{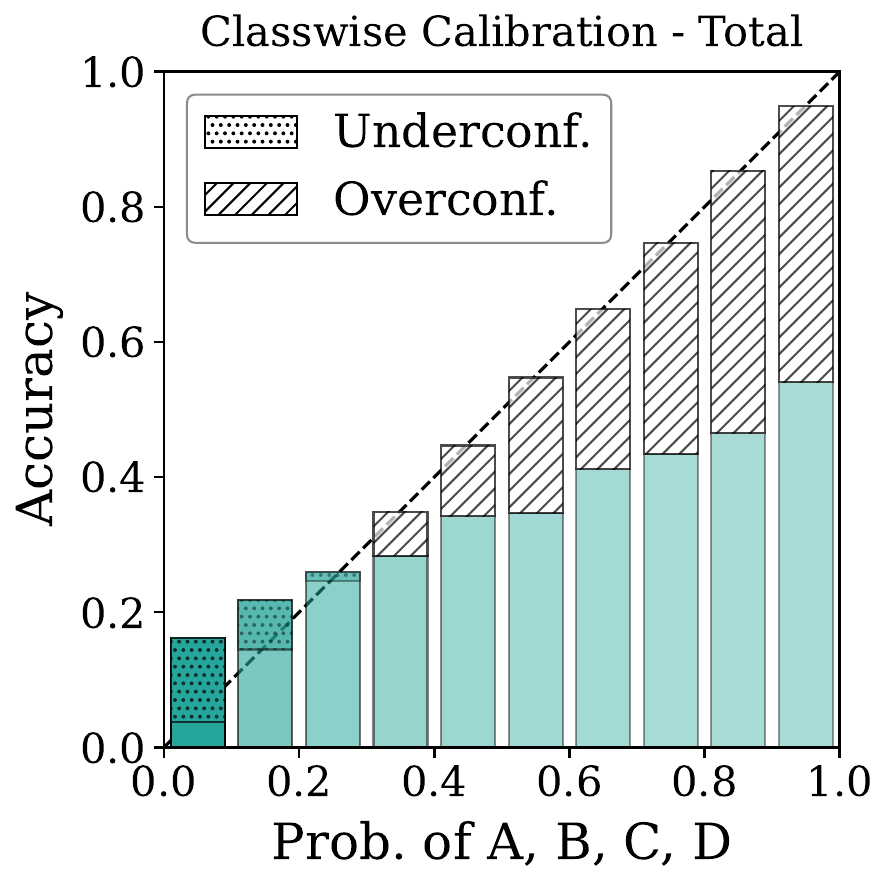}\end{subfigure}
    \begin{subfigure}[b]{0.21\textwidth}\centering\includegraphics[width=\linewidth]{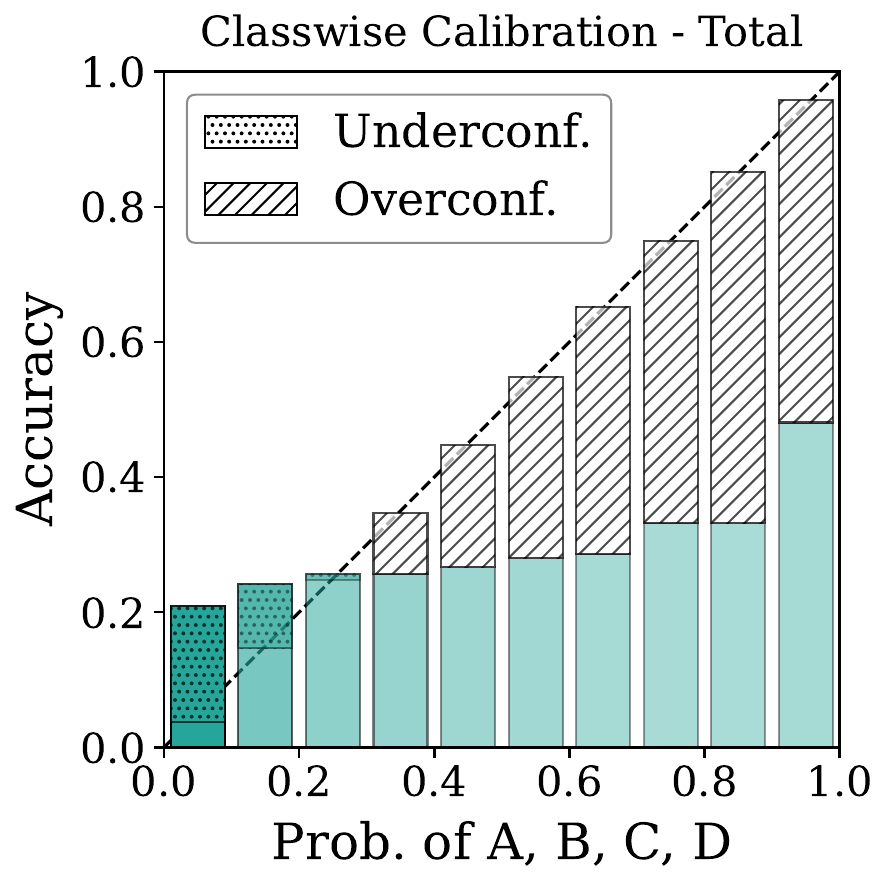}\end{subfigure}
    \begin{subfigure}[b]{0.21\textwidth}\centering\includegraphics[width=\linewidth]{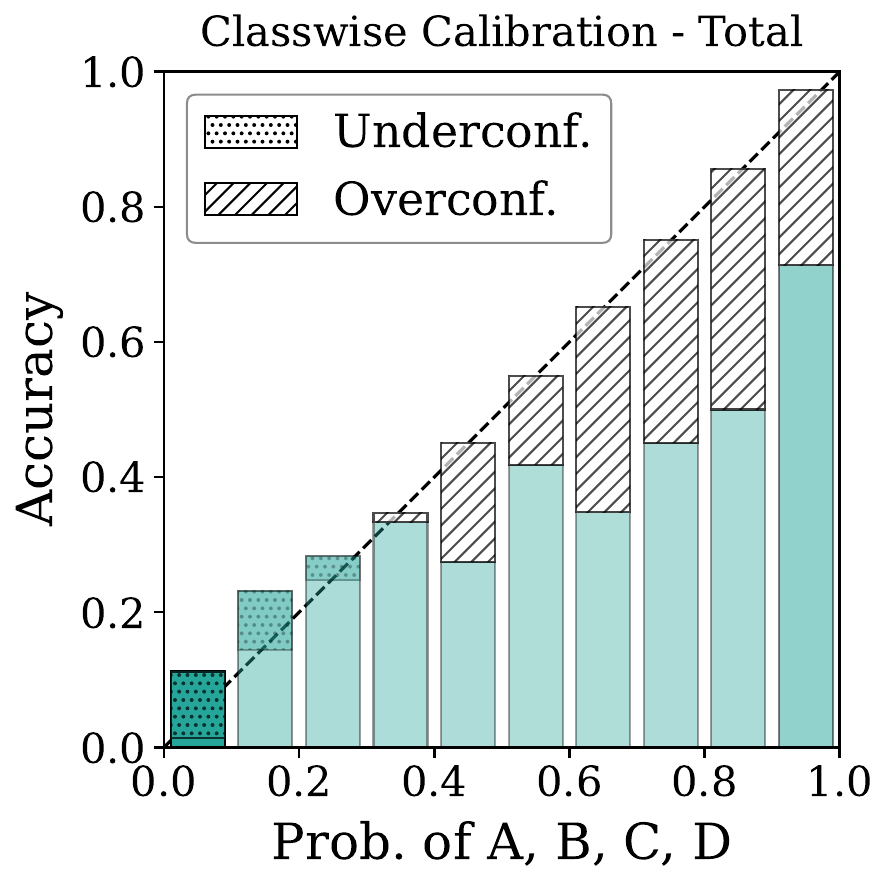}\end{subfigure}
    \begin{subfigure}[b]{0.21\textwidth}\centering\includegraphics[width=\linewidth]{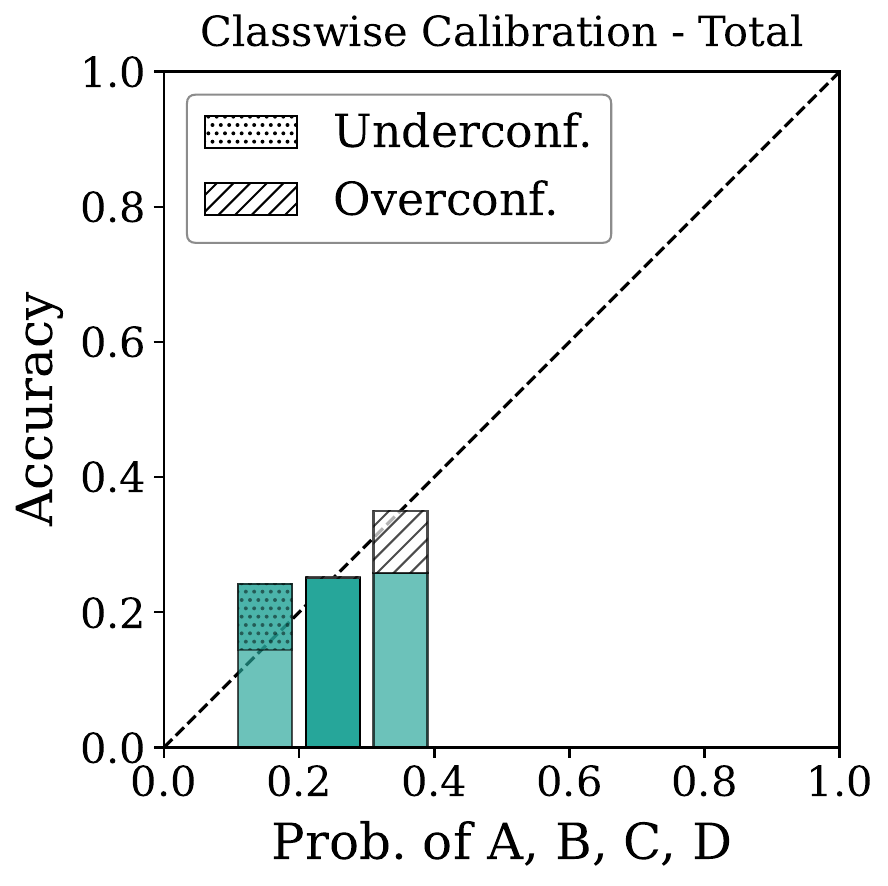}\end{subfigure}
    \\[3pt]
    \textbf{CFT}\\[1pt]
    \begin{subfigure}[b]{0.21\textwidth}\centering\includegraphics[width=\linewidth]{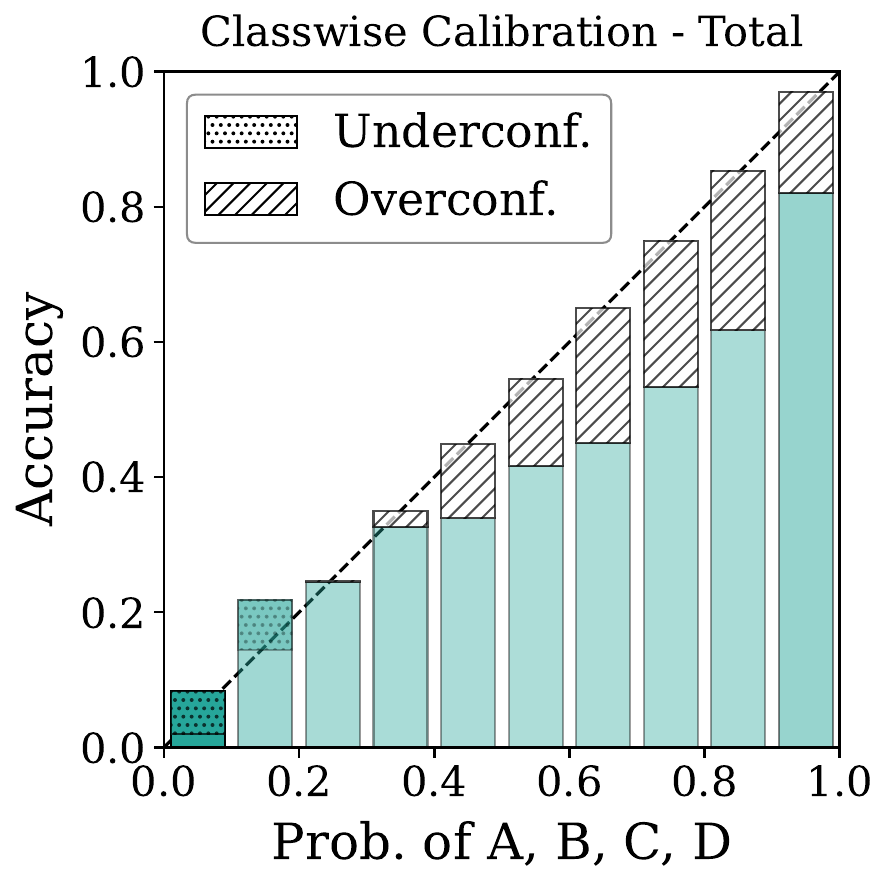}\end{subfigure}
    \begin{subfigure}[b]{0.21\textwidth}\centering\includegraphics[width=\linewidth]{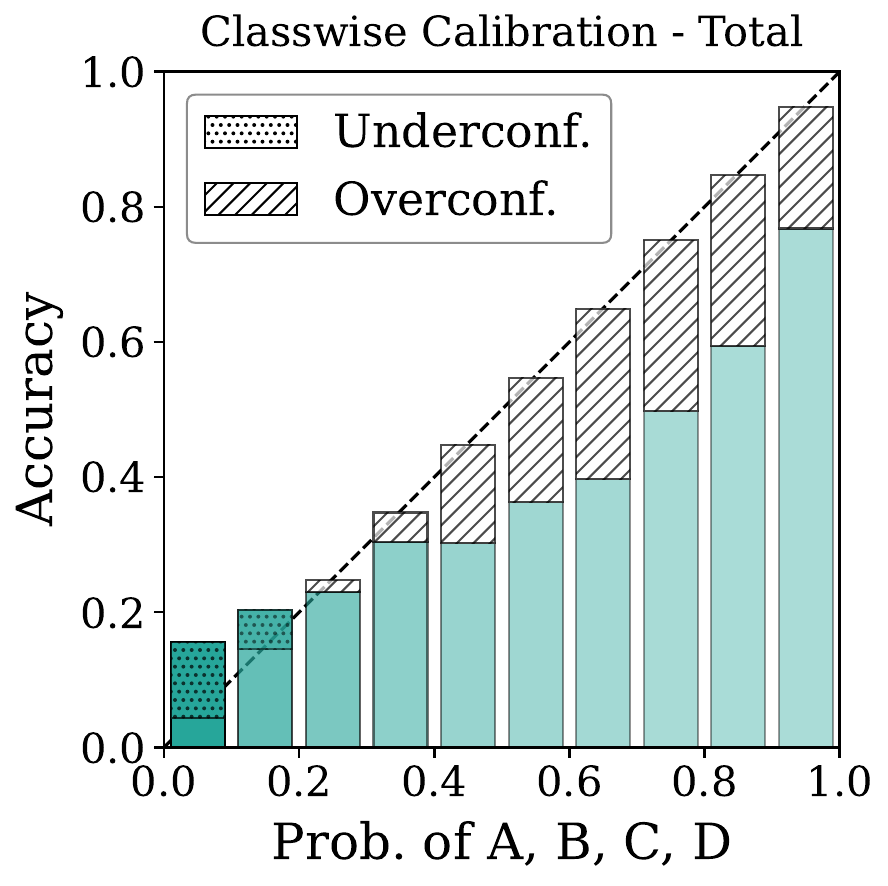}\end{subfigure}
    \begin{subfigure}[b]{0.21\textwidth}\centering\includegraphics[width=\linewidth]{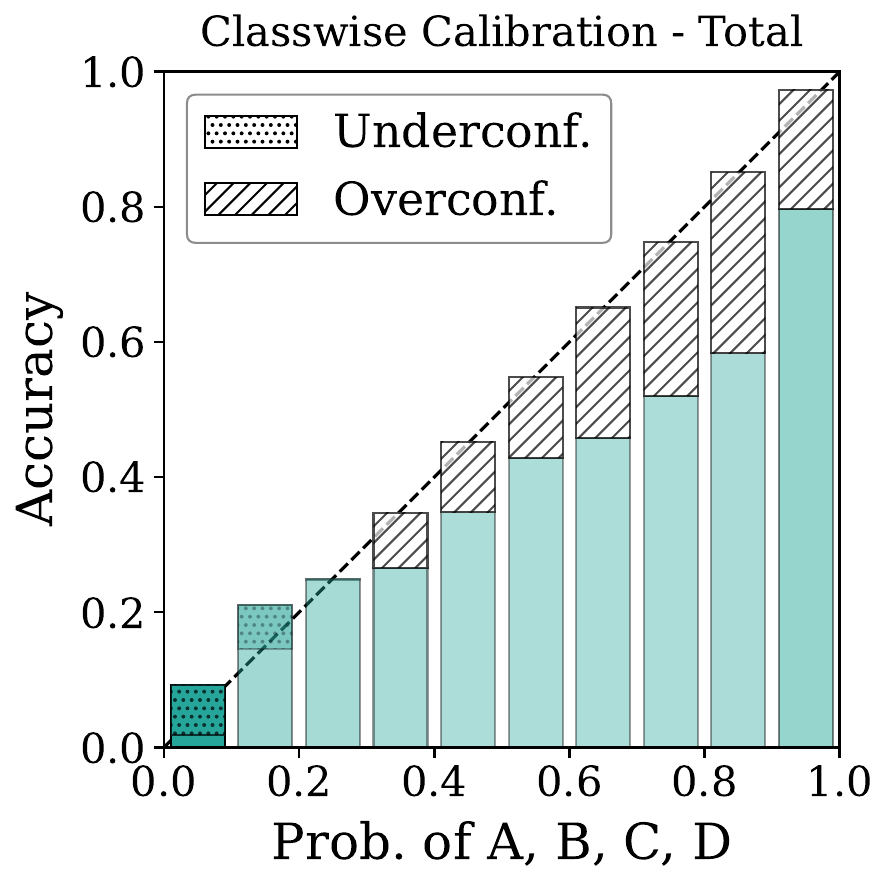}\end{subfigure}
    \begin{subfigure}[b]{0.21\textwidth}\centering\includegraphics[width=\linewidth]{figures/cft_mcqa/CFT_Mistral_ood_cwece_total.pdf}\end{subfigure}
    \\[3pt]
    \textbf{CALM}\\[1pt]
    \begin{subfigure}[b]{0.21\textwidth}\centering\includegraphics[width=\linewidth]{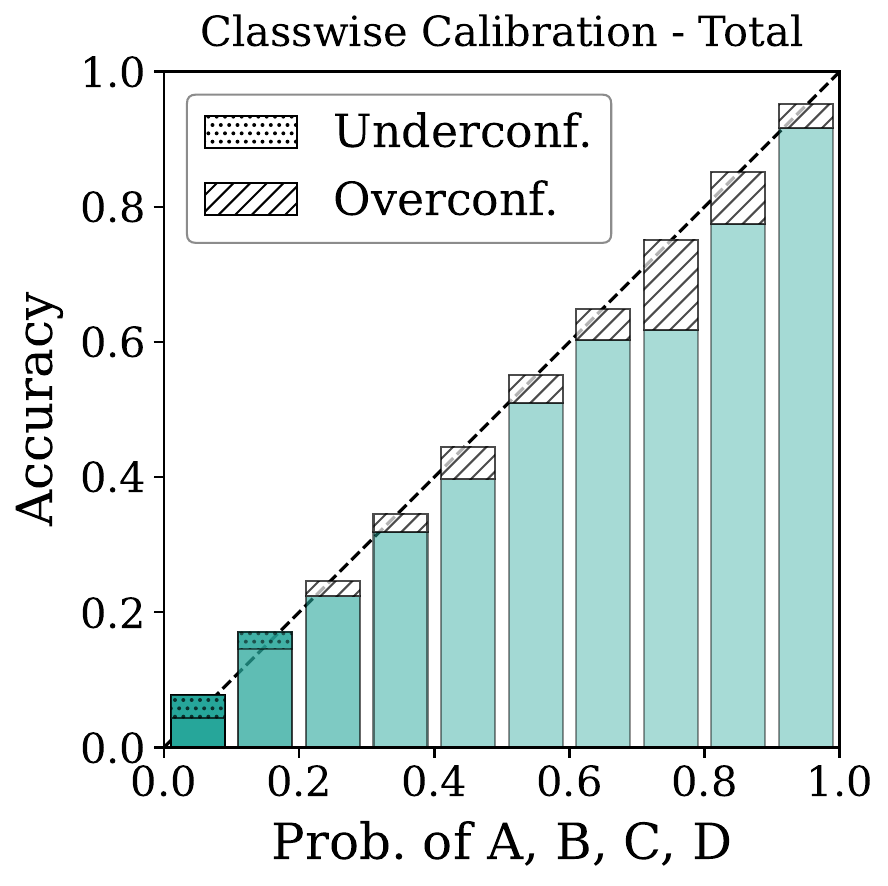}\end{subfigure}
    \begin{subfigure}[b]{0.21\textwidth}\centering\includegraphics[width=\linewidth]{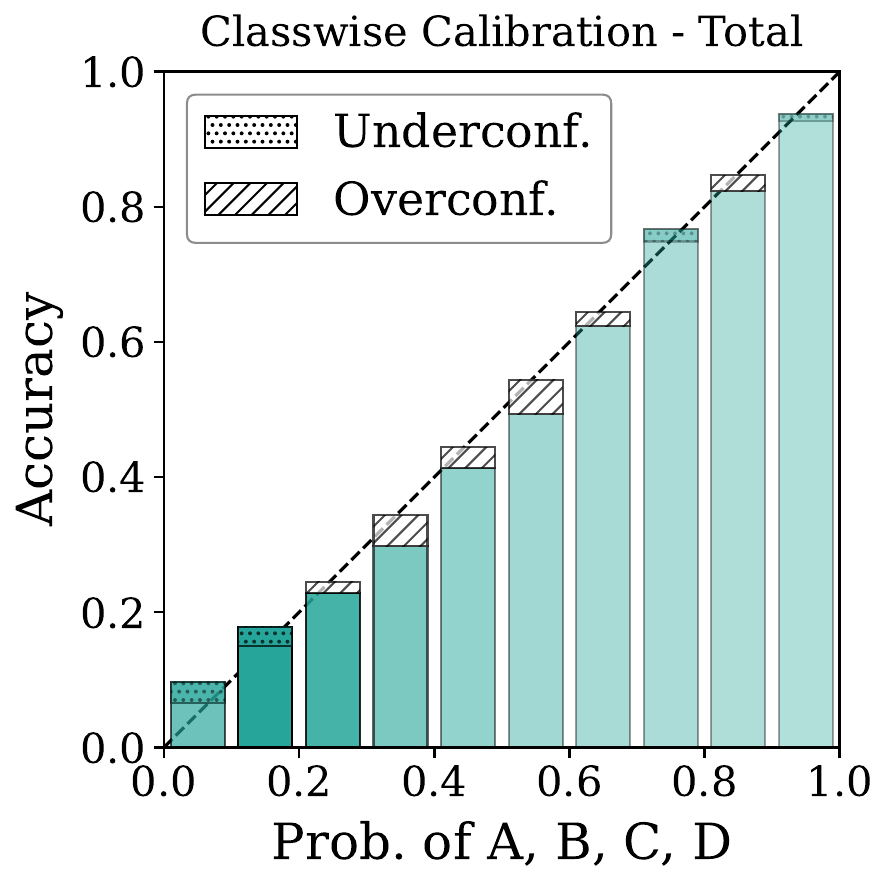}\end{subfigure}
    \begin{subfigure}[b]{0.21\textwidth}\centering\includegraphics[width=\linewidth]{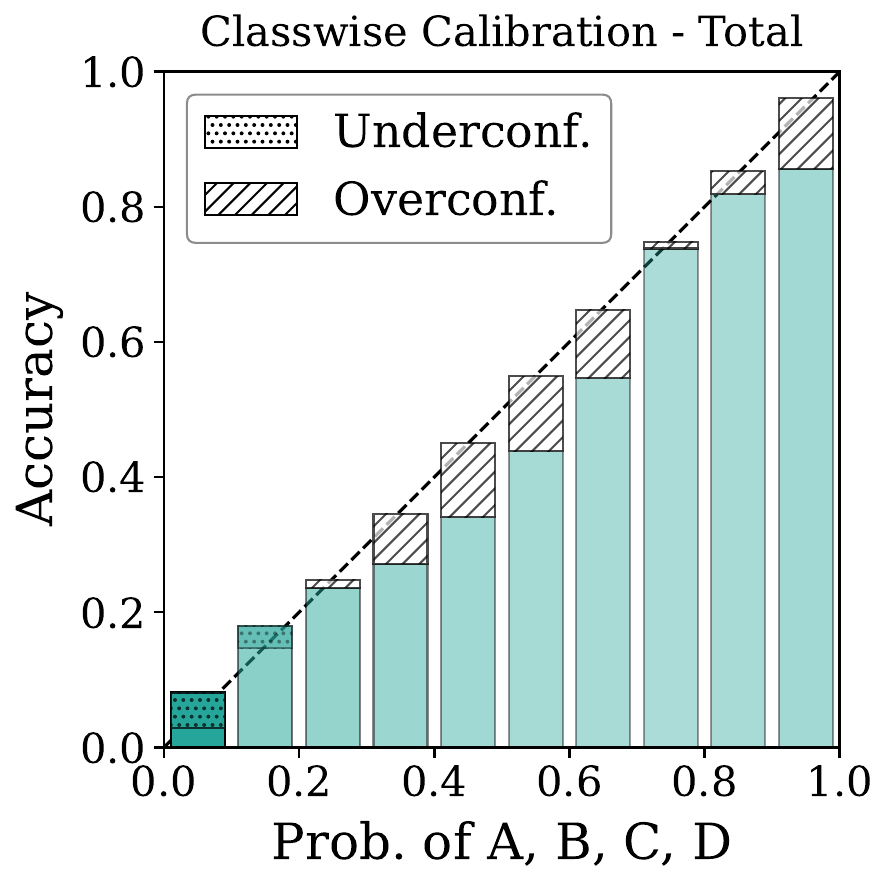}\end{subfigure}
    \begin{subfigure}[b]{0.21\textwidth}\centering\includegraphics[width=\linewidth]{figures/hypergradient_best_ece_cwece_total.pdf}\end{subfigure}
    \\[3pt]
    \caption{Classwise (Total) reliability diagrams in the \textbf{OOD MCQA setting}. Each block is a method (bold header); within a block the four panels are Llama-3.1, Vicuna-7B, OLMo-2-7B, and Mistral-7B (left to right, labeled in the top block). The dashed diagonal is perfect calibration; bars below it indicate overconfidence.}
    \label{fig:ood}
\end{figure}

\begin{figure}[htbp]
    \centering
    \textbf{DPO/RLHF}\\[1pt]
    \begin{subfigure}[b]{0.21\textwidth}\centering\includegraphics[width=\linewidth]{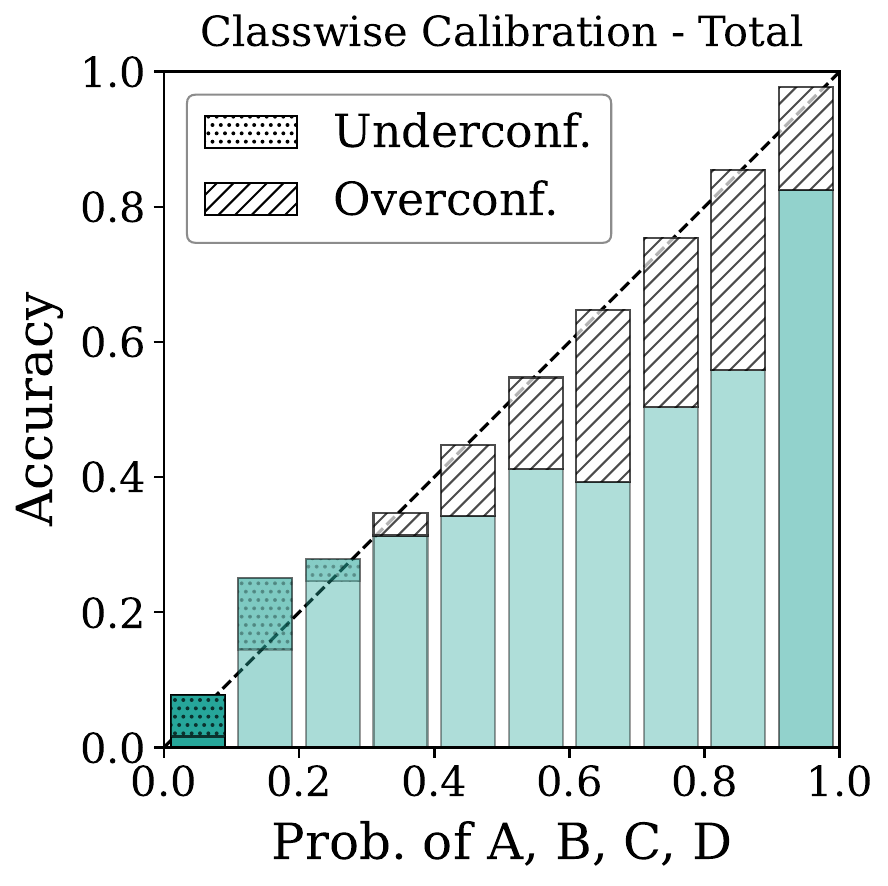}\caption{Llama-3.1}\end{subfigure}
    \begin{subfigure}[b]{0.21\textwidth}\centering\includegraphics[width=\linewidth]{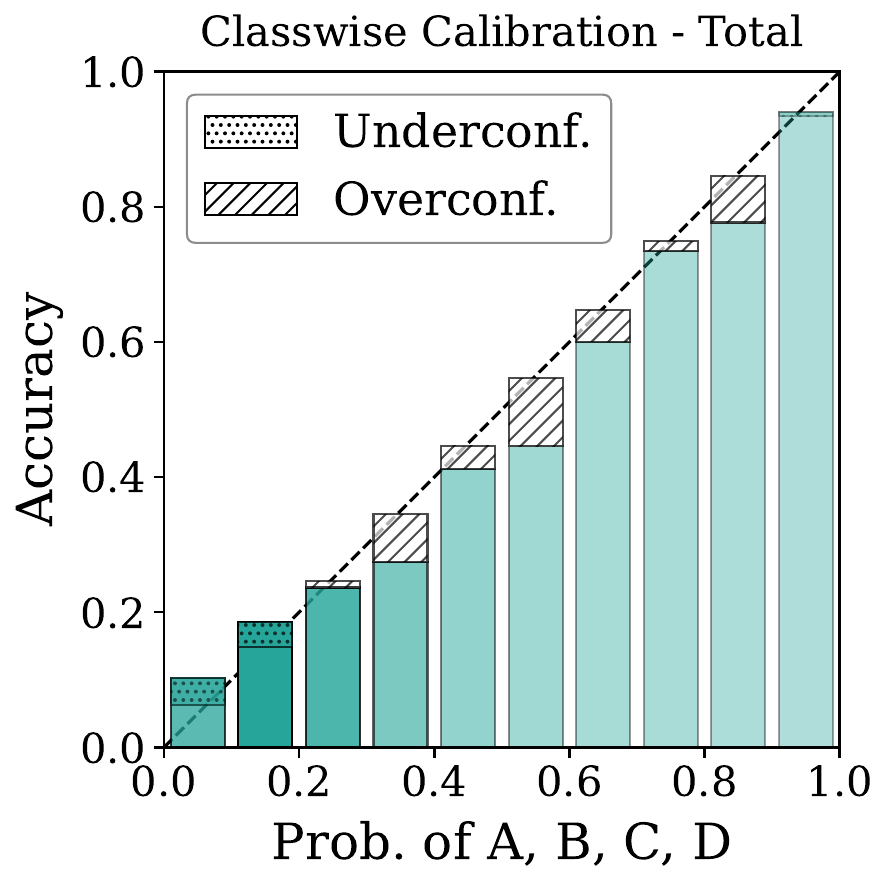}\caption{Vicuna-7B}\end{subfigure}
    \begin{subfigure}[b]{0.21\textwidth}\centering\includegraphics[width=\linewidth]{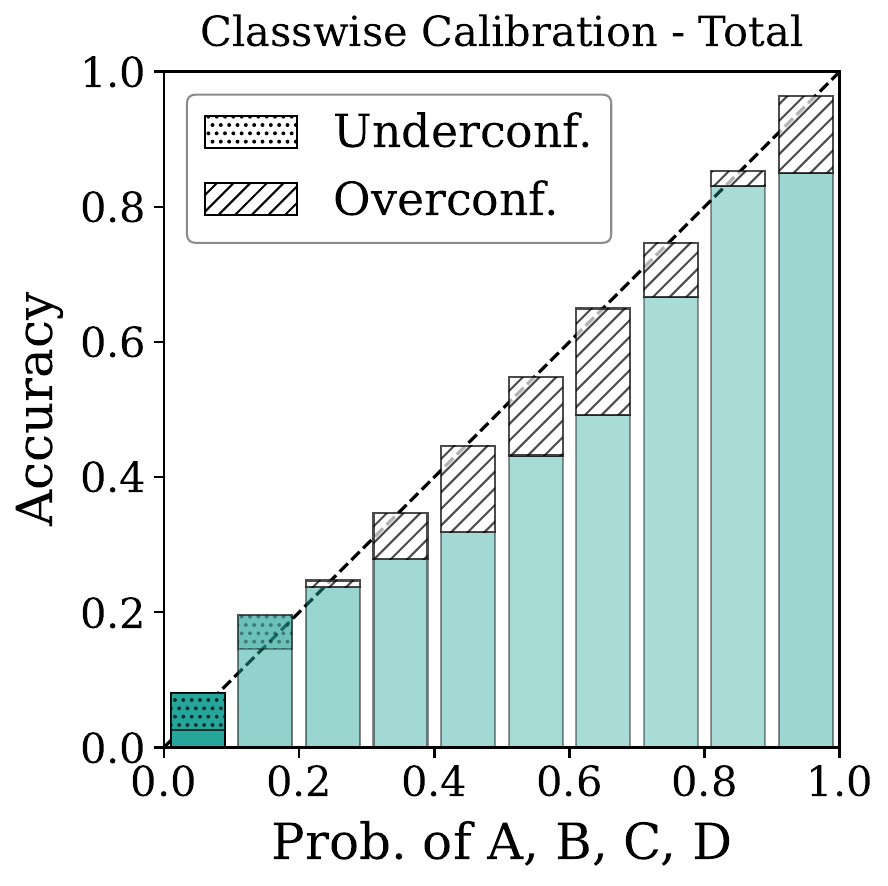}\caption{OLMo-2-7B}\end{subfigure}
    \begin{subfigure}[b]{0.21\textwidth}\centering\includegraphics[width=\linewidth]{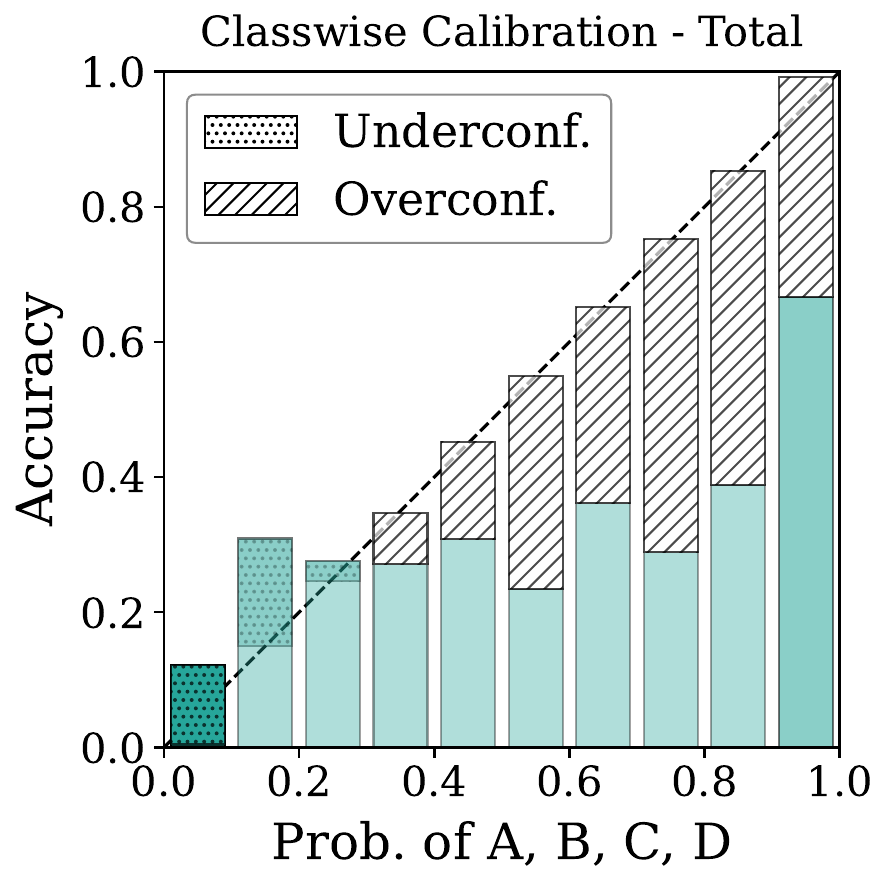}\caption{Mistral-7B}\end{subfigure}
    \\[3pt]
    \textbf{Temp.\ Scale.}\\[1pt]
    \begin{subfigure}[b]{0.21\textwidth}\centering\includegraphics[width=\linewidth]{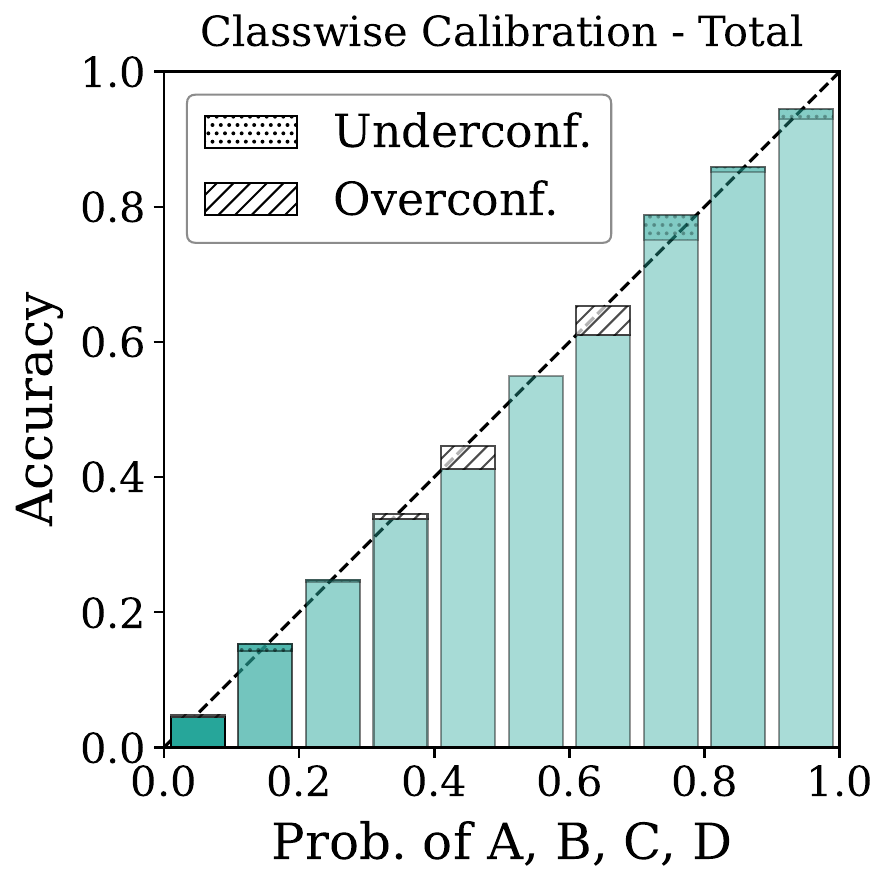}\end{subfigure}
    \begin{subfigure}[b]{0.21\textwidth}\centering\includegraphics[width=\linewidth]{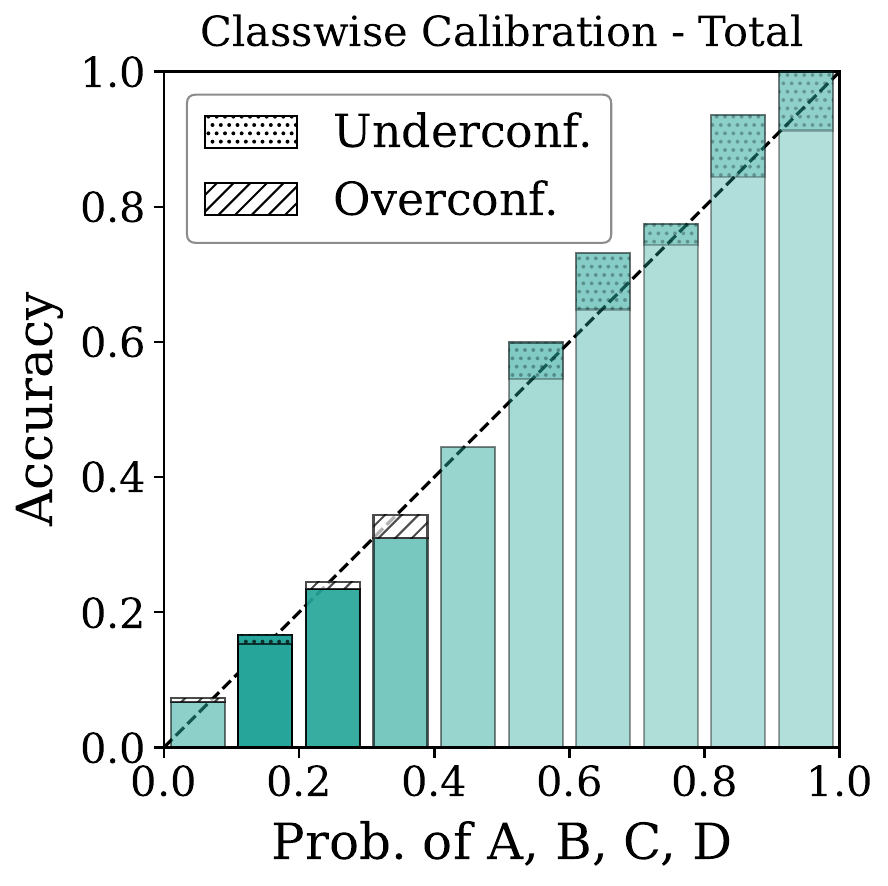}\end{subfigure}
    \begin{subfigure}[b]{0.21\textwidth}\centering\includegraphics[width=\linewidth]{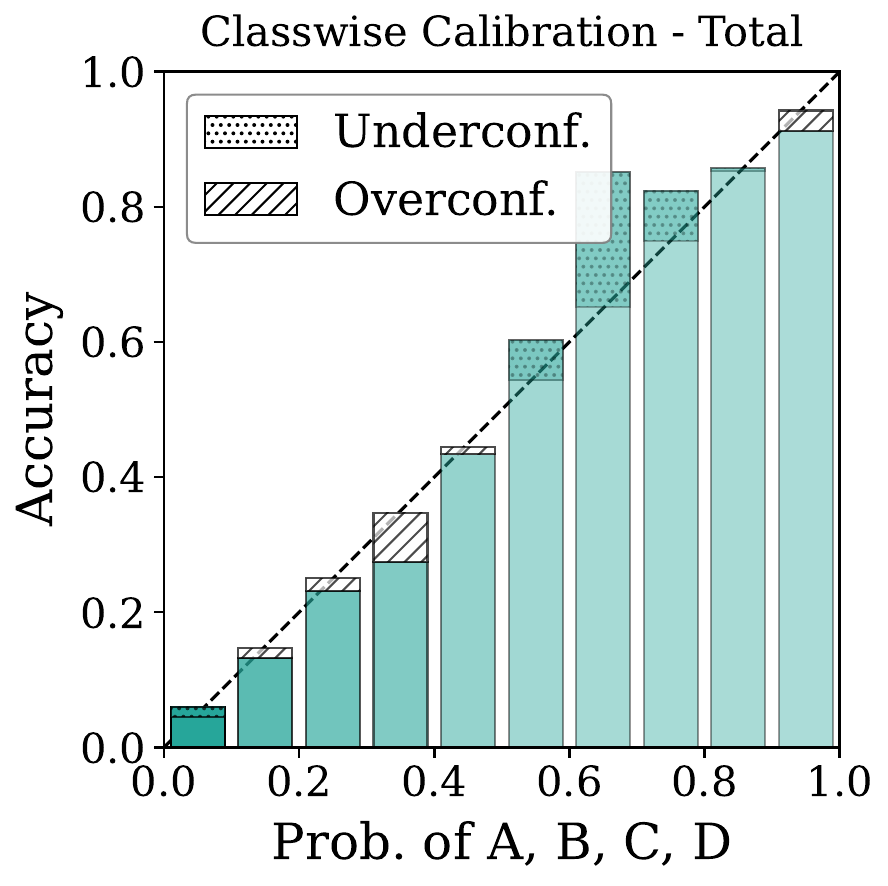}\end{subfigure}
    \begin{subfigure}[b]{0.21\textwidth}\centering\includegraphics[width=\linewidth]{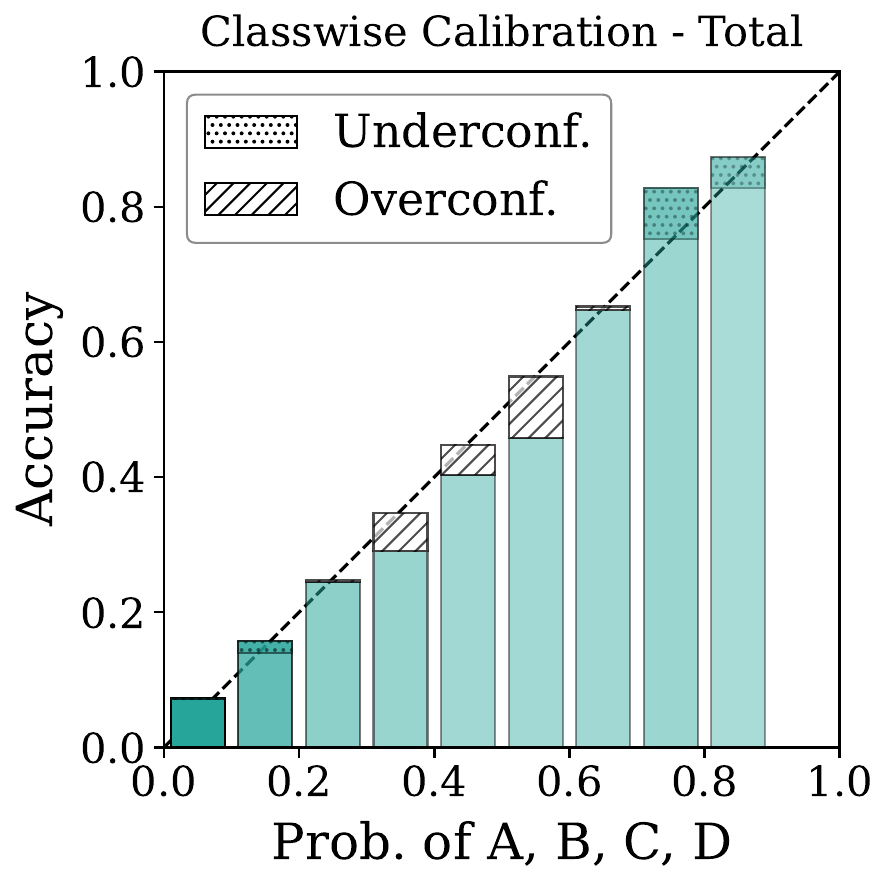}\end{subfigure}
    \\[3pt]
    \textbf{Label Smooth.}\\[1pt]
    \begin{subfigure}[b]{0.21\textwidth}\centering\includegraphics[width=\linewidth]{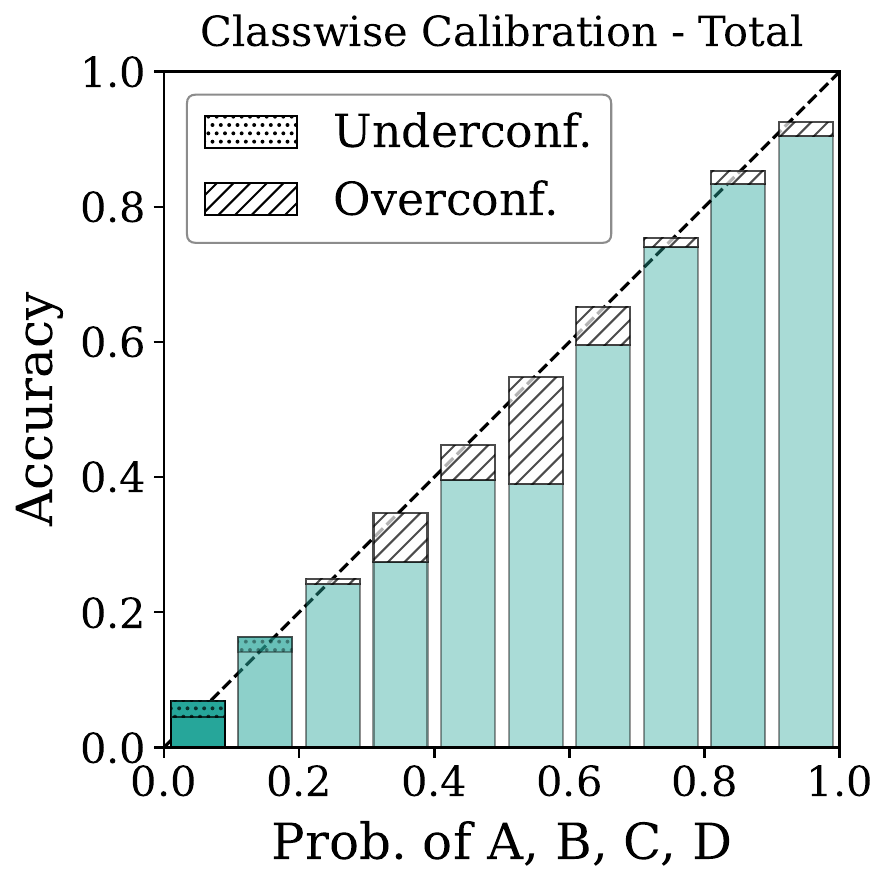}\end{subfigure}
    \begin{subfigure}[b]{0.21\textwidth}\centering\includegraphics[width=\linewidth]{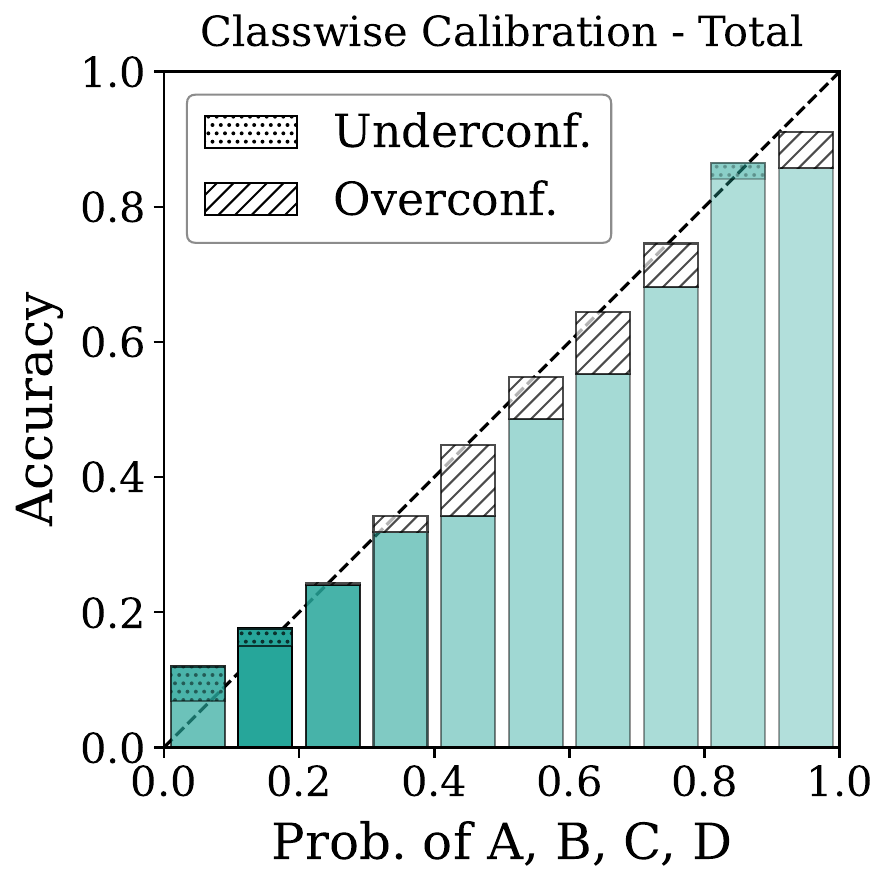}\end{subfigure}
    \begin{subfigure}[b]{0.21\textwidth}\centering\includegraphics[width=\linewidth]{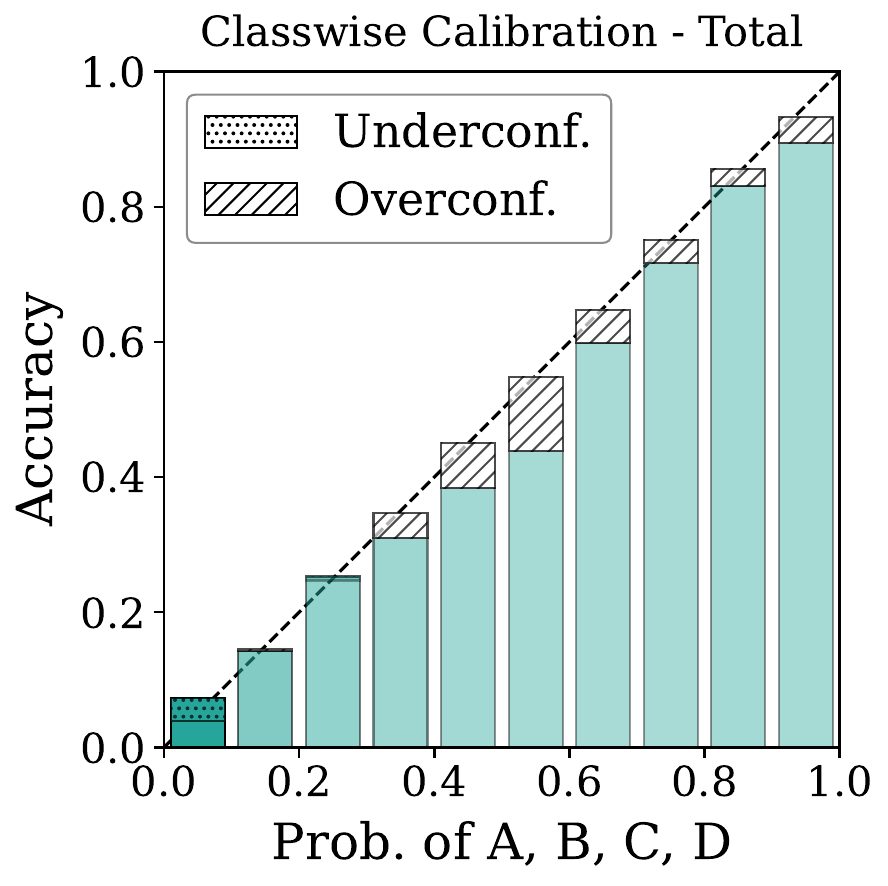}\end{subfigure}
    \begin{subfigure}[b]{0.21\textwidth}\centering\includegraphics[width=\linewidth]{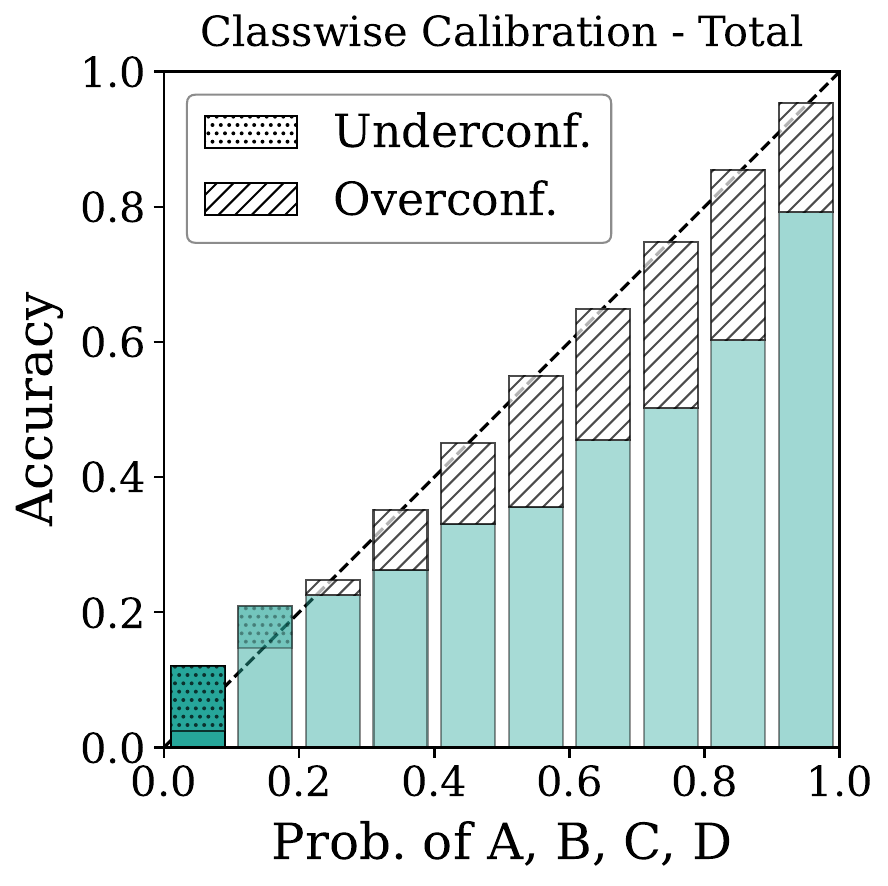}\end{subfigure}
    \\[3pt]
    \textbf{Regularization}\\[1pt]
    \begin{subfigure}[b]{0.21\textwidth}\centering\includegraphics[width=\linewidth]{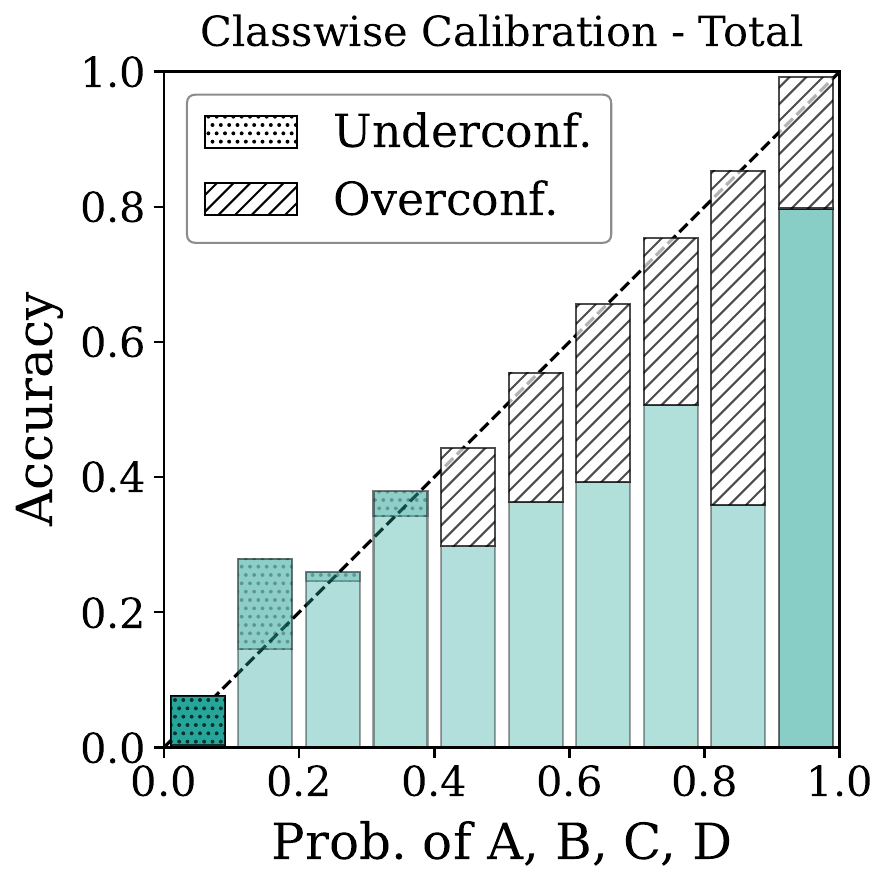}\end{subfigure}
    \begin{subfigure}[b]{0.21\textwidth}\centering\includegraphics[width=\linewidth]{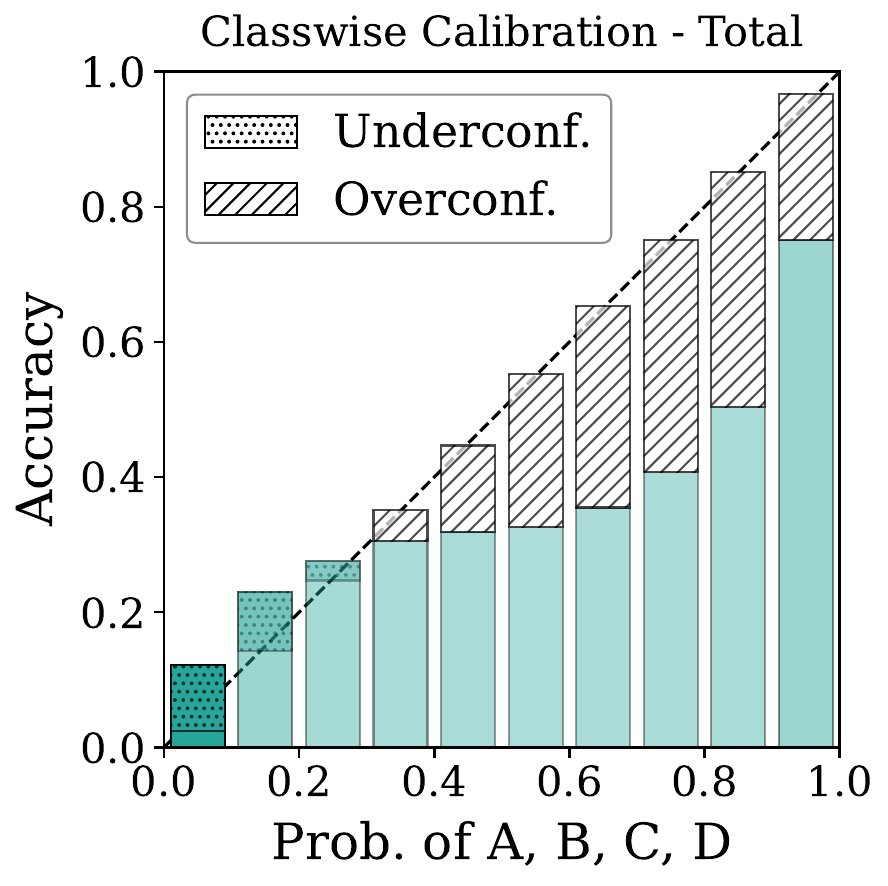}\end{subfigure}
    \begin{subfigure}[b]{0.21\textwidth}\centering\includegraphics[width=\linewidth]{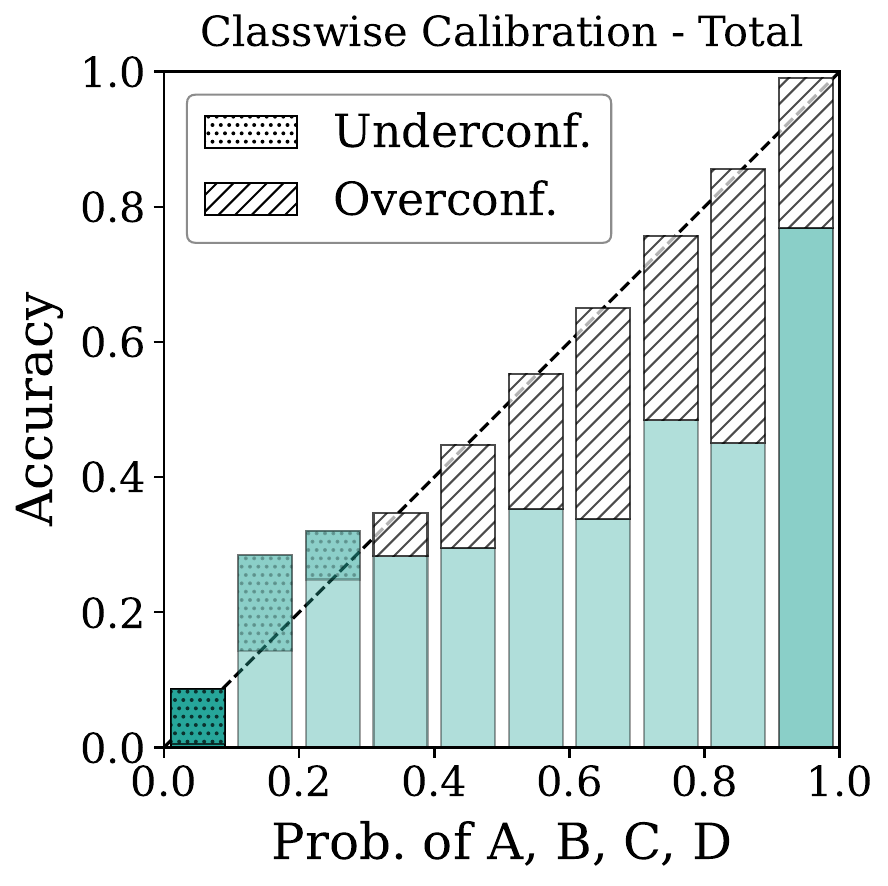}\end{subfigure}
    \begin{subfigure}[b]{0.21\textwidth}\centering\includegraphics[width=\linewidth]{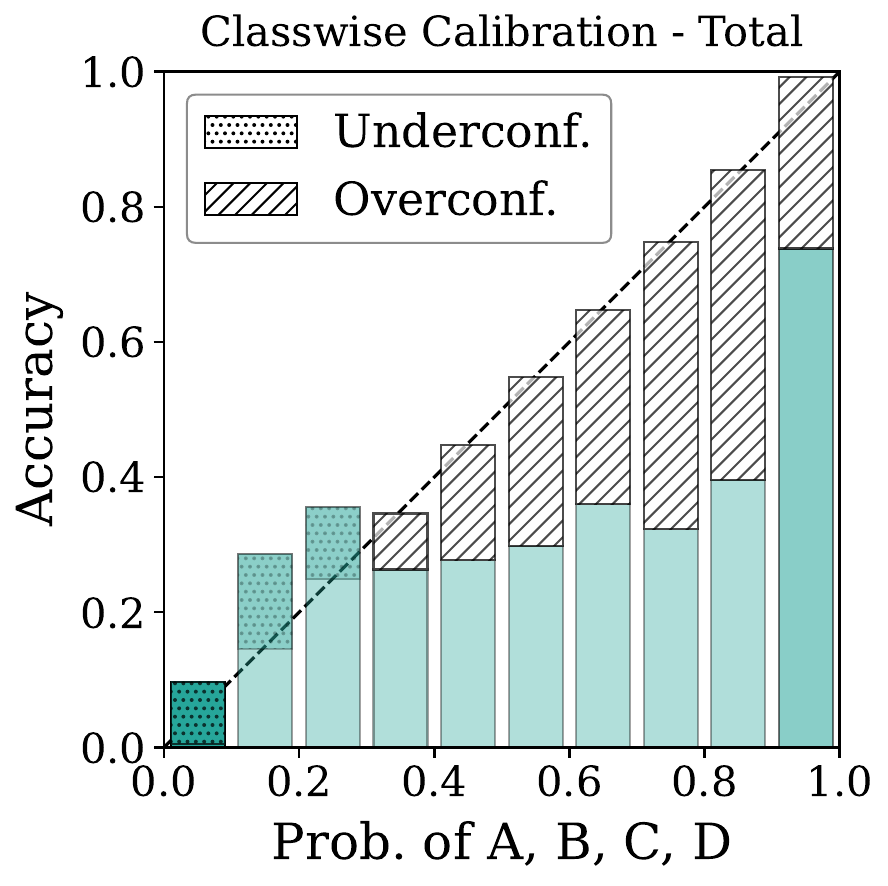}\end{subfigure}
    \\[3pt]
    \caption{Classwise (Total) reliability diagrams in the \textbf{ID MCQA setting}, part 1: DPO/RLHF, Temperature Scaling, Label Smoothing, and Regularization. Within each block the four panels are Llama-3.1, Vicuna-7B, OLMo-2-7B, Mistral-7B (left to right). Remaining methods in Figure~\ref{fig:id_b}.}
    \label{fig:id_a}
\end{figure}

\begin{figure}[htbp]
    \centering
    \textbf{Iterate}\\[1pt]
    \begin{subfigure}[b]{0.21\textwidth}\centering\includegraphics[width=\linewidth]{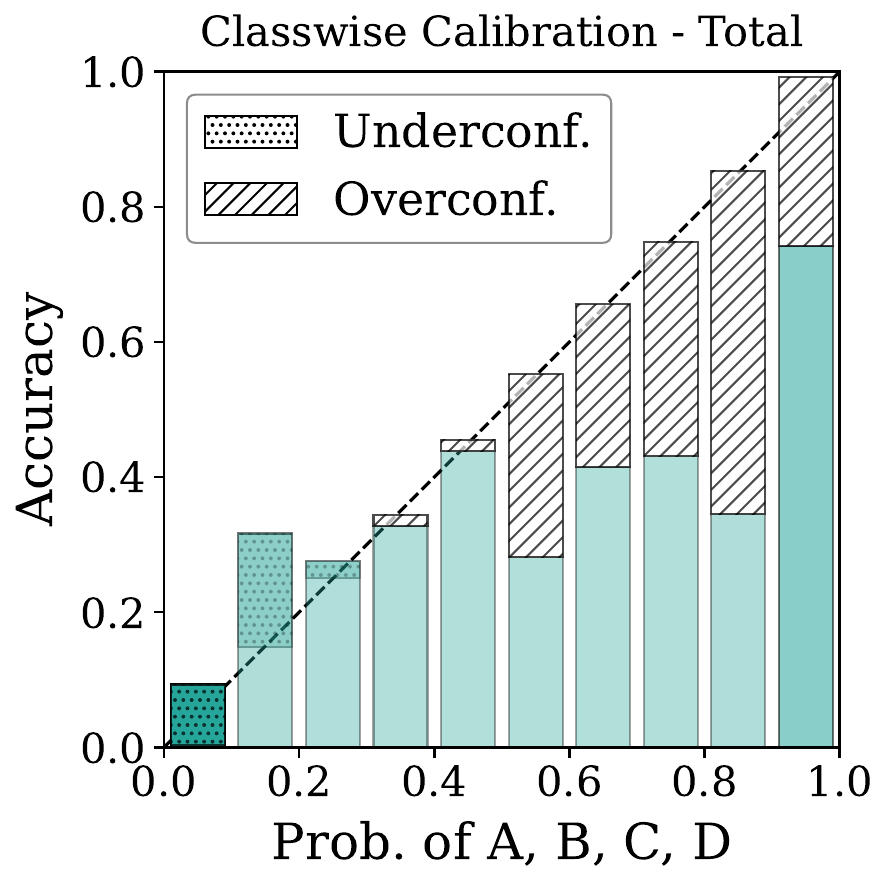}\caption{Llama-3.1}\end{subfigure}
    \begin{subfigure}[b]{0.21\textwidth}\centering\includegraphics[width=\linewidth]{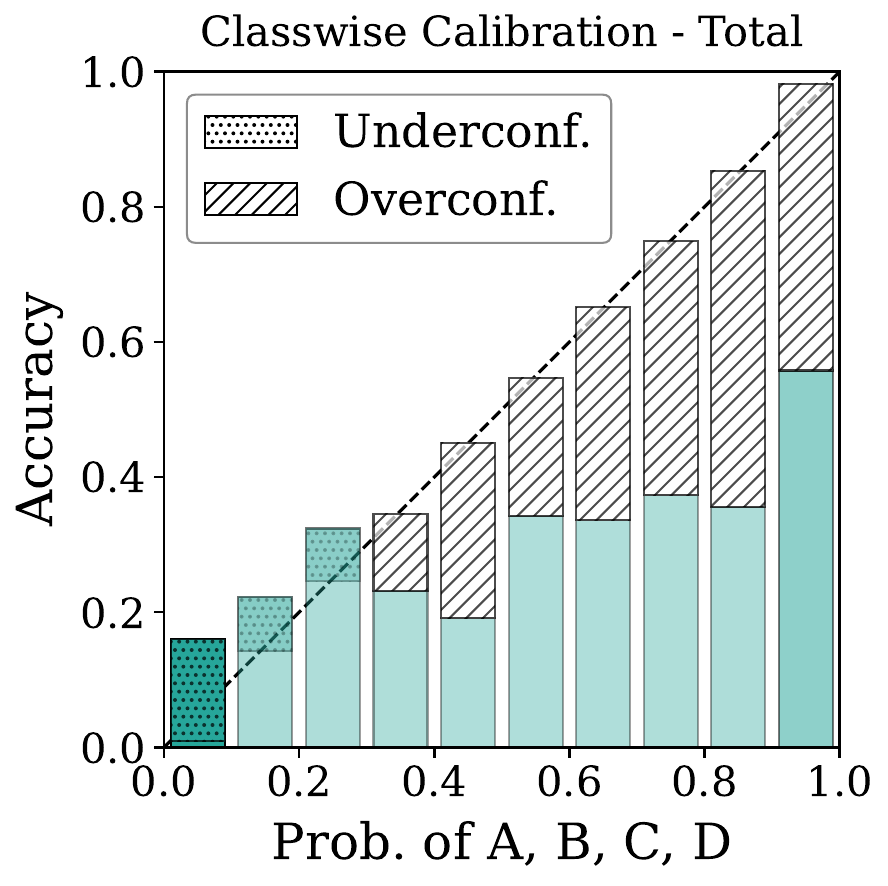}\caption{Vicuna-7B}\end{subfigure}
    \begin{subfigure}[b]{0.21\textwidth}\centering\includegraphics[width=\linewidth]{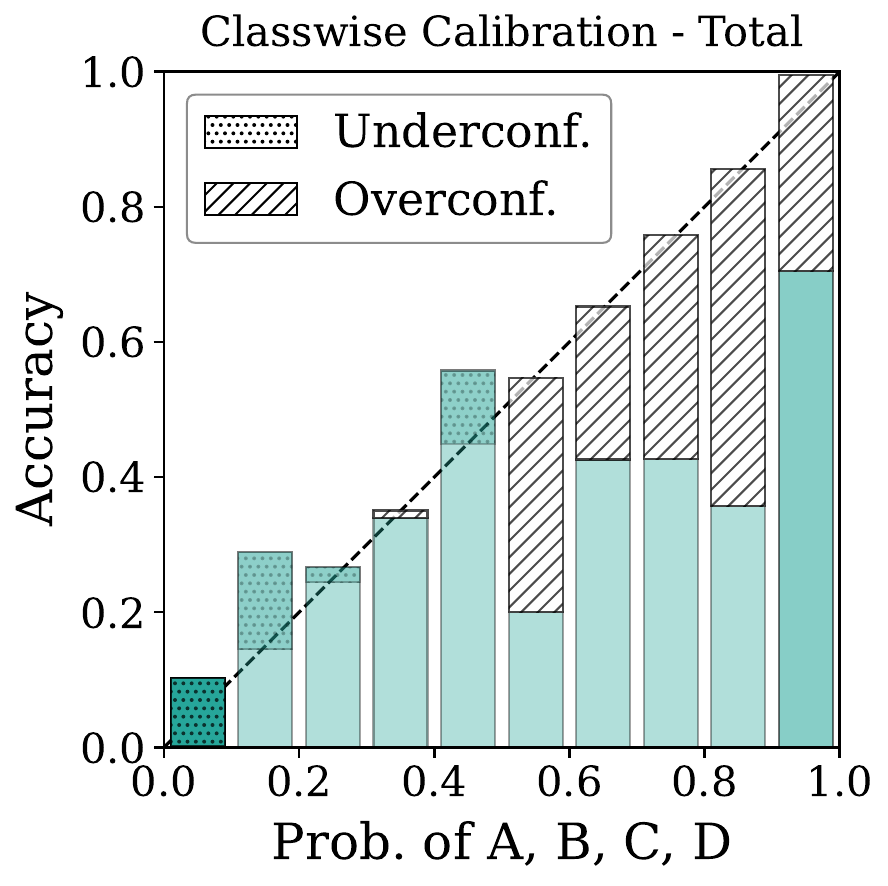}\caption{OLMo-2-7B}\end{subfigure}
    \begin{subfigure}[b]{0.21\textwidth}\centering\includegraphics[width=\linewidth]{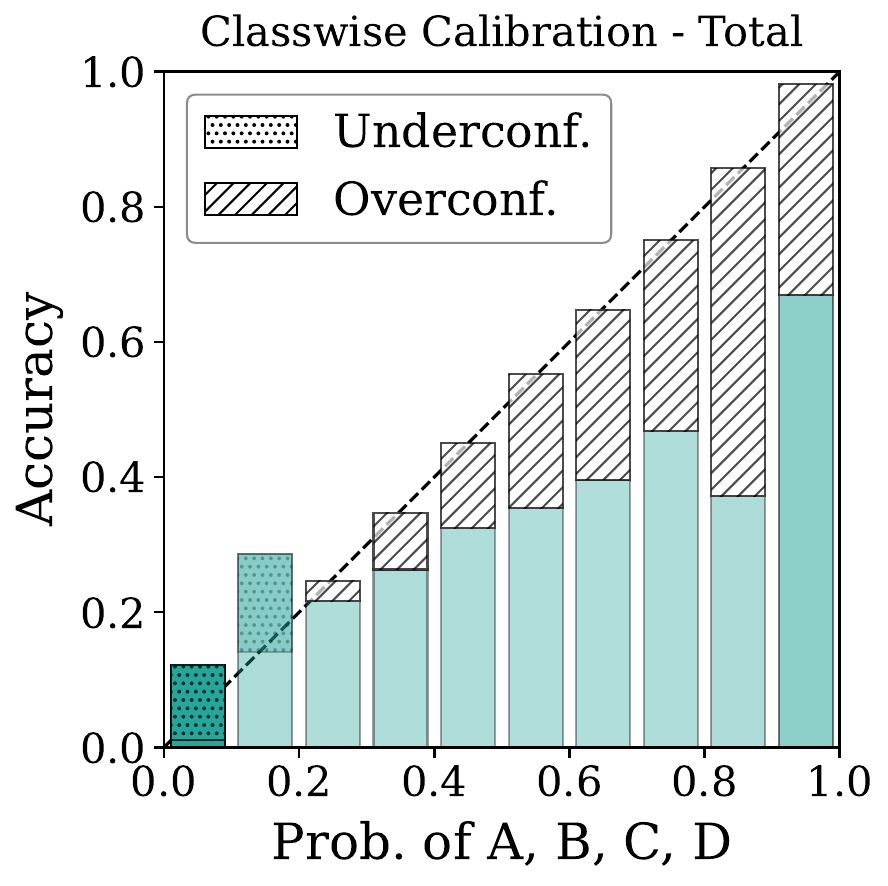}\caption{Mistral-7B}\end{subfigure}
    \\[3pt]
    \textbf{CFT}\\[1pt]
    \begin{subfigure}[b]{0.21\textwidth}\centering\includegraphics[width=\linewidth]{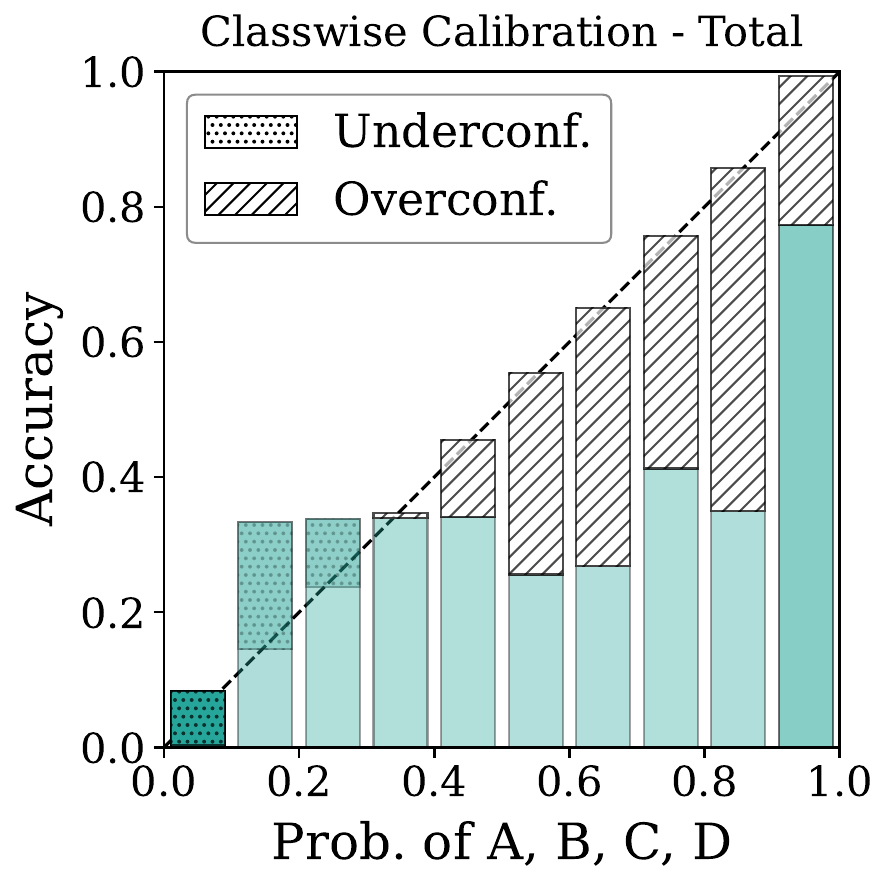}\end{subfigure}
    \begin{subfigure}[b]{0.21\textwidth}\centering\includegraphics[width=\linewidth]{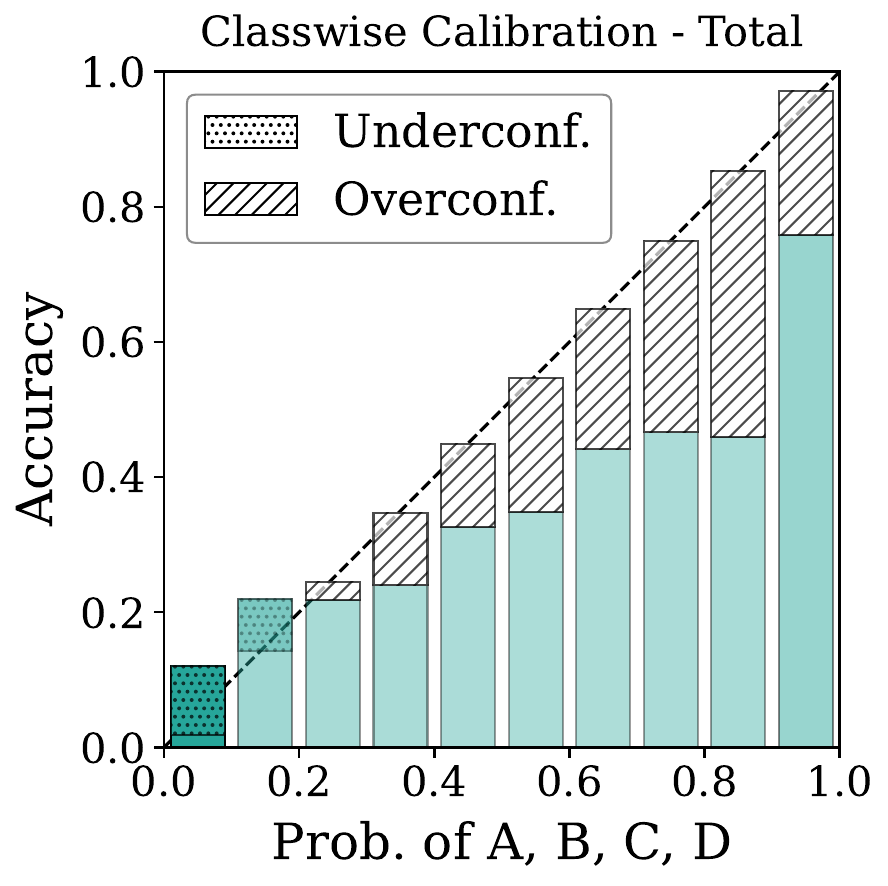}\end{subfigure}
    \begin{subfigure}[b]{0.21\textwidth}\centering\includegraphics[width=\linewidth]{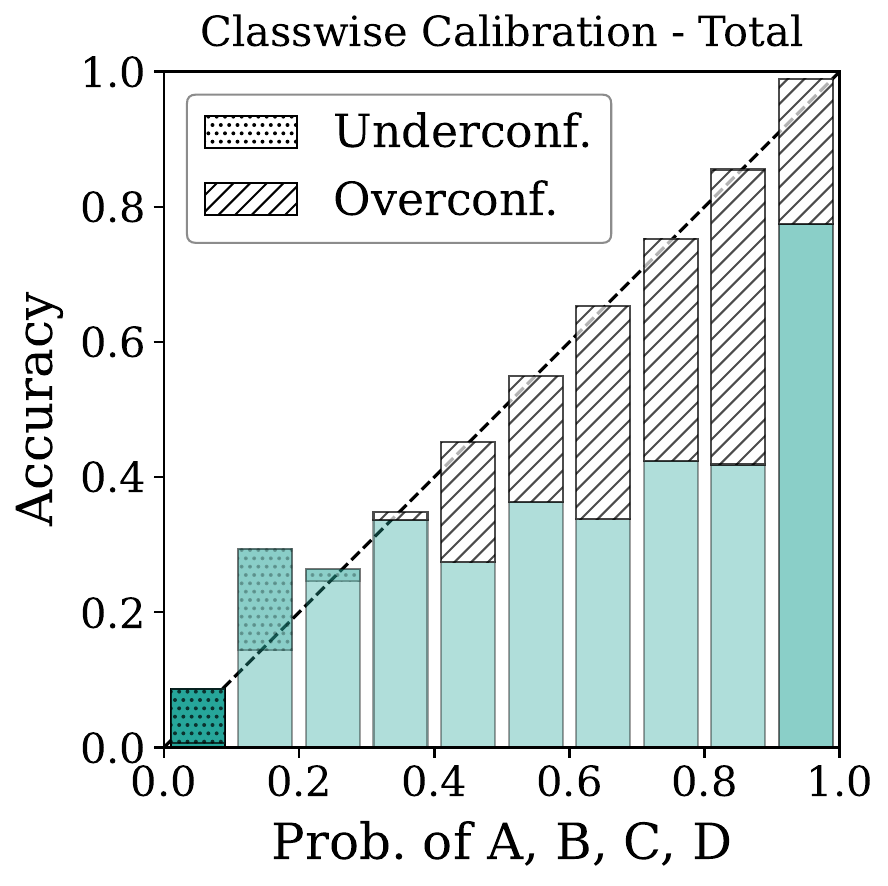}\end{subfigure}
    \begin{subfigure}[b]{0.21\textwidth}\centering\includegraphics[width=\linewidth]{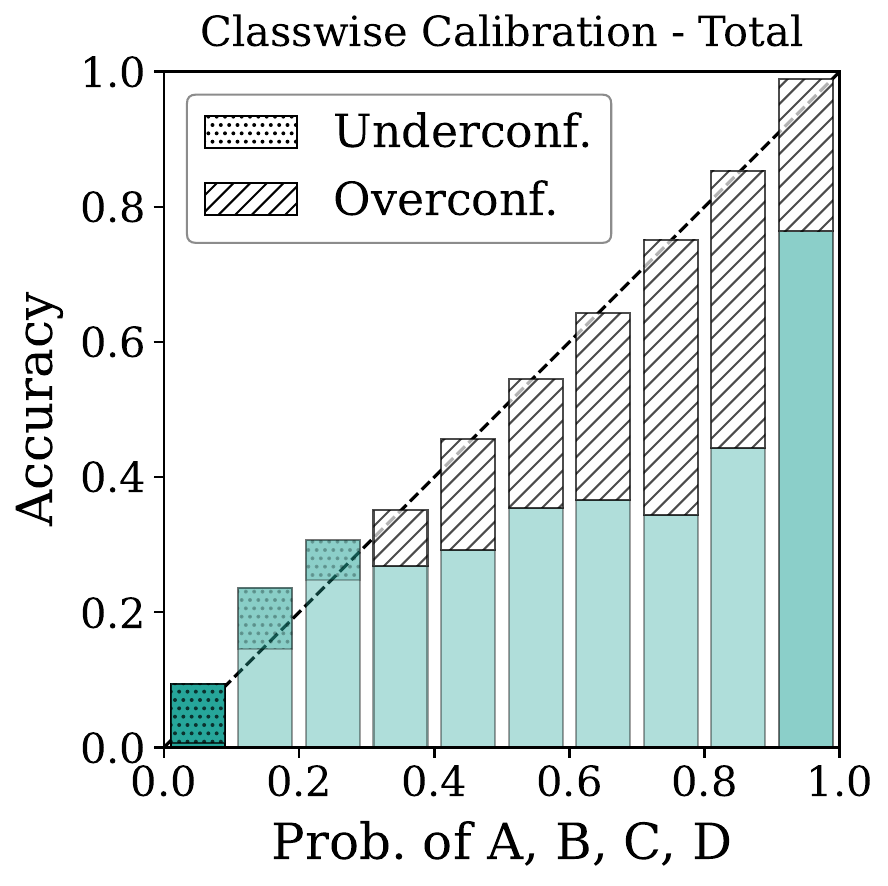}\end{subfigure}
    \\[3pt]
    \textbf{CALM}\\[1pt]
    \begin{subfigure}[b]{0.21\textwidth}\centering\includegraphics[width=\linewidth]{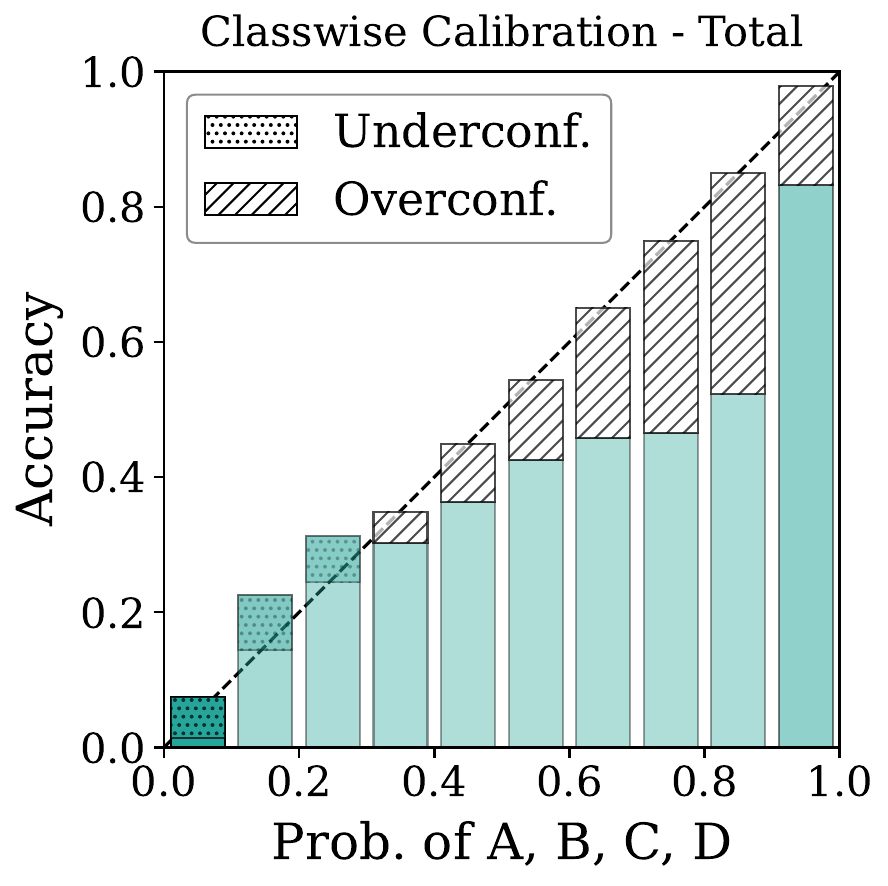}\end{subfigure}
    \begin{subfigure}[b]{0.21\textwidth}\centering\includegraphics[width=\linewidth]{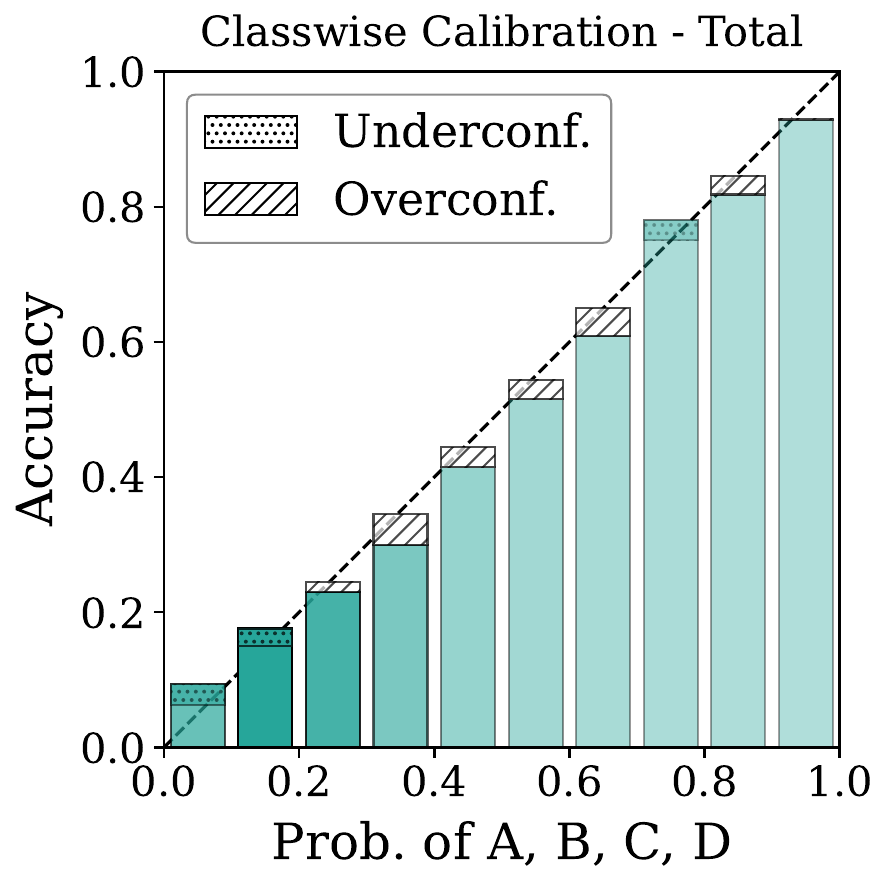}\end{subfigure}
    \begin{subfigure}[b]{0.21\textwidth}\centering\includegraphics[width=\linewidth]{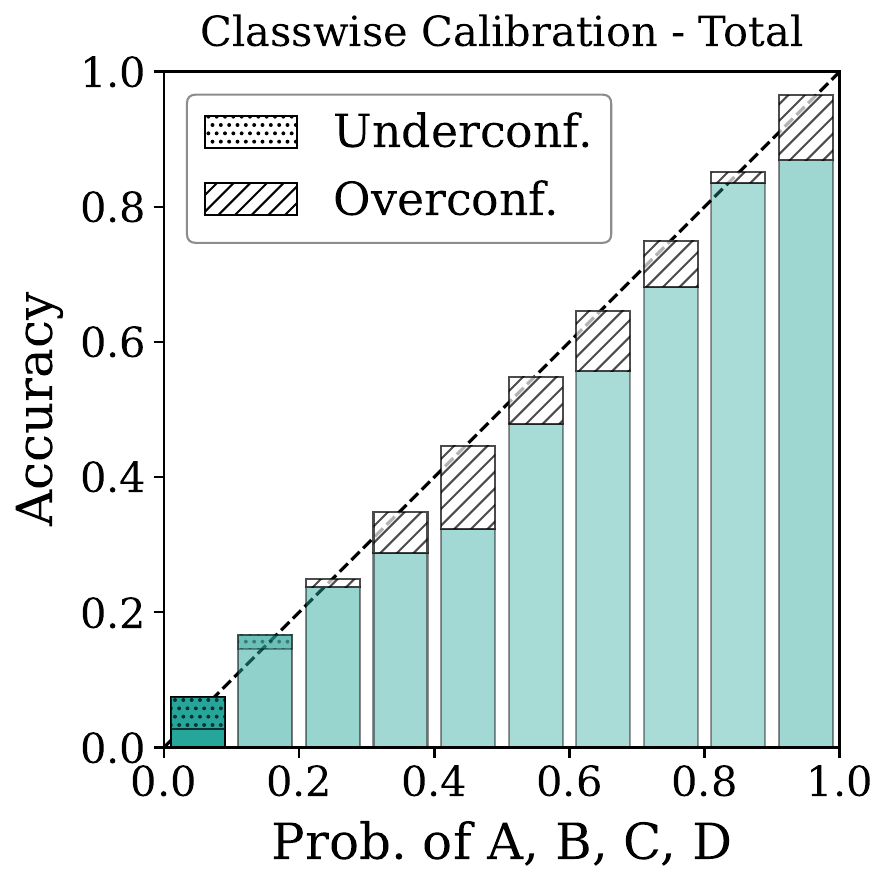}\end{subfigure}
    \begin{subfigure}[b]{0.21\textwidth}\centering\includegraphics[width=\linewidth]{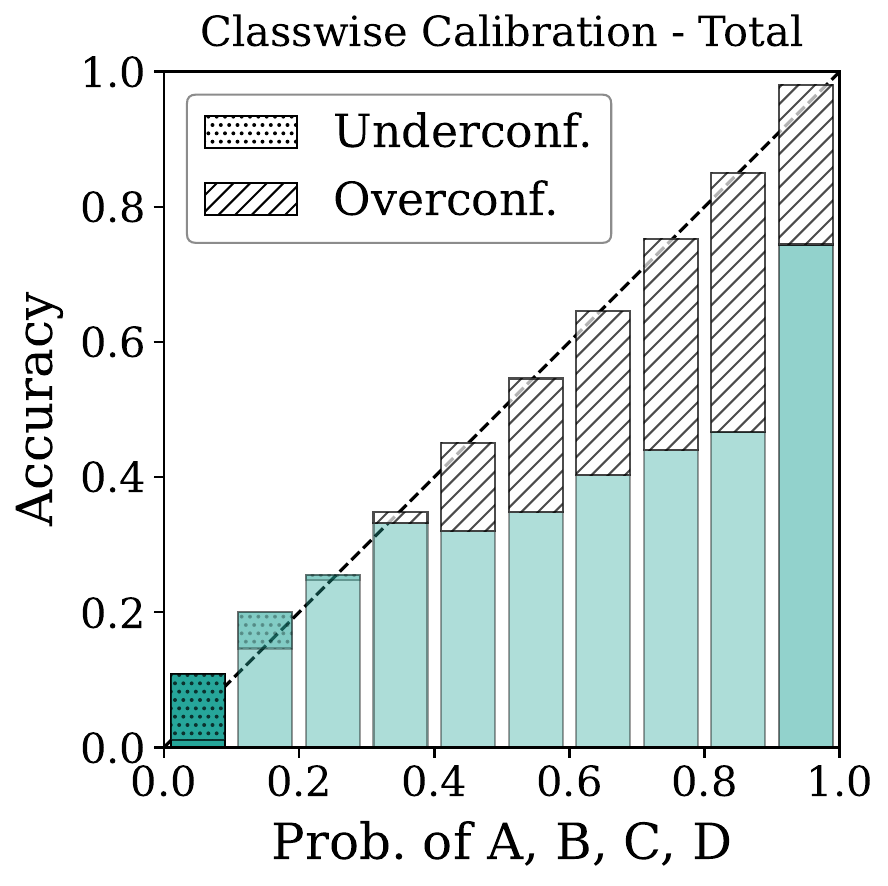}\end{subfigure}
    \\[3pt]
    \caption{Classwise (Total) reliability diagrams in the \textbf{ID MCQA setting}, part 2: Iterate, CFT, and CALM. Panels within each block are ordered as in Figure~\ref{fig:id_a}.}
    \label{fig:id_b}
\end{figure}

This section presents the complete set of classwise (Total) reliability diagrams for all four models and all methods. Figure~\ref{fig:ood} covers the OOD MCQA setting; Figures~\ref{fig:id_a} and~\ref{fig:id_b} cover the ID MCQA setting (split across methods for readability). Across settings, the DPO/RLHF baselines are clearly overconfident; Temperature Scaling and CFT reduce miscalibration only partially and inconsistently (for Llama-3.1, TS even increases it); Regularization and Iterate help to varying degrees but can distort the curve or collapse accuracy (e.g., Iterate on Vicuna-7B); and CALM produces the most consistent alignment with the diagonal across models and settings.

\end{document}